\pdfoutput=1

\documentclass[11pt]{article}

\usepackage[final]{acl}

\usepackage{times}
\usepackage{latexsym}
 \usepackage{amsmath}  
\usepackage{amssymb}  
\usepackage{amsthm}
\usepackage{mathtools} 

\usepackage{booktabs}       
\usepackage{tabularx}       
\usepackage{threeparttable} 
\usepackage[T1]{fontenc}

\usepackage[utf8]{inputenc}
 
\usepackage{microtype}

\usepackage{inconsolata}

\usepackage{hyperref}
\usepackage{graphicx}
 \usepackage{colortbl}
\usepackage{mathrsfs}
\usepackage{mathtools}
\usepackage{algorithm,algpseudocode}
\usepackage{cleveref}
\usepackage{tcolorbox}
\usepackage{dialogue}
\usepackage{subfigure}
\usepackage{adjustbox}     
\usepackage{listings}      
\usepackage{xcolor}        
\usepackage{caption}       
 \usepackage{subcaption} 
\usepackage{float}         
\usepackage{enumitem}
\newtheoremstyle{definition}
{3pt} 
{3pt} 
{} 
{} 
{\bfseries} 
{.} 
{.5em} 
{} 

\DeclareMathOperator*{\argmax}{arg\,max}

\theoremstyle{definition}
\newtheorem{theorem}{Theorem}
\newtheorem{lemma}[theorem]{Lemma}

\newtheorem{corollary}[theorem]{Corollary}

\definecolor{darkmagenta}{rgb}{0.55, 0, 0.55}  
\newcommand{\an}[1]{\textcolor{darkmagenta}{#1}}

\newcommand{\fricname}{Frictional Agent Alignment Framework}
\newcommand{\fricabbr}{\texttt{FAAF}}

\newcommand{\actionspace}{\mathcal{Y}}
\newcommand{\frictionspace}{\mathcal{F}}

\title{\fricname: Slow Down and Don't Break Things}

\author{Abhijnan Nath, Carine Graff, Andrei Bachinin, \and Nikhil Krishnaswamy \\
    Situated Grounding and Natural Language (SIGNAL) Lab \\ Department of Computer Science, Colorado State University \\ Fort Collins, CO, USA \\ \texttt{\{abhijnan.nath,nkrishna\}@colostate.edu}}

\begin{document}
\maketitle
\begin{abstract}
AI support of collaborative interactions entails mediating potential misalignment between interlocutor beliefs. Common preference alignment methods like DPO excel in static settings, but struggle in dynamic collaborative tasks where the explicit signals of interlocutor beliefs are sparse and skewed. We propose the \fricname~(\fricabbr), to generate precise, context-aware "friction" that prompts for deliberation and re-examination of existing evidence. \fricabbr’s two-player objective decouples from data skew: a frictive-state policy identifies belief misalignments, while an intervention policy crafts collaborator-preferred responses. We derive an analytical solution to this objective, enabling training a single policy via a simple supervised loss. Experiments on three benchmarks show \fricabbr~outperforms competitors in producing concise, interpretable friction and in OOD generalization. By aligning LLMs to act as adaptive "thought partners"---not passive responders---\fricabbr~advances scalable, dynamic human-AI collaboration. Our code and data can be found at \url{https://github.com/csu-signal/FAAF_ACL}.

\end{abstract}

\section{Introduction}
\label{sec:intro}
\vspace*{-2mm}


\begin{figure}
    \centering
    \includegraphics[width=.95\linewidth]{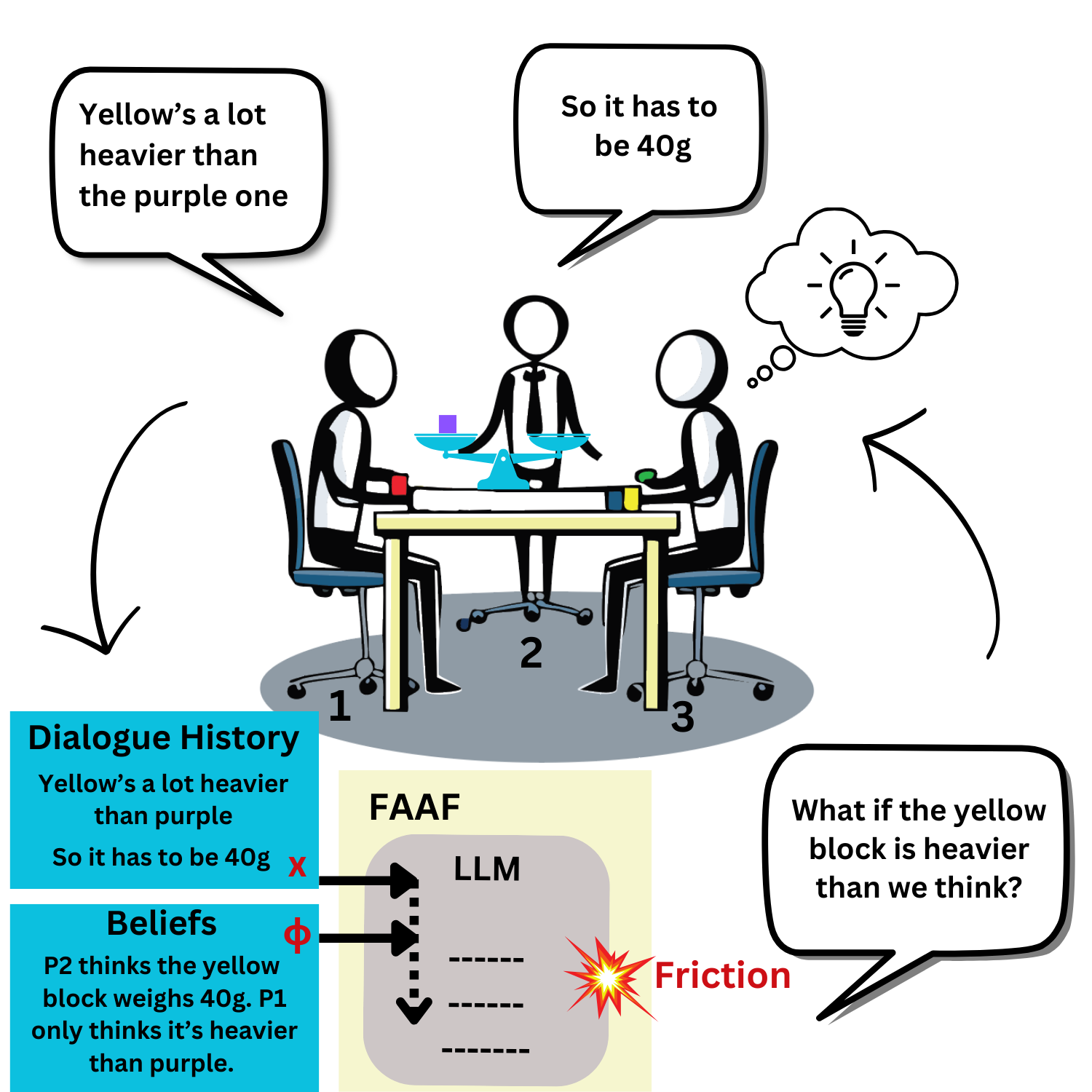}
\vspace*{-2mm}
    \caption{\fricabbr~conditions responses on both the dialogue context $x$ and representation of the "frictive" (belief) state $\phi$, to generate outputs that prompt for reflection, deliberation, and verification of evidence.}
    \label{fig:faaf-cartoon}
\vspace*{-2mm}
\end{figure}

When collaborating to solve problems, humans continually interrogate each other’s intentions and assumptions \cite{stalnaker2002common,asher2003common,klein2005common}. With the rapid integration of generative AI, exemplified by large language models (LLMs), into personal, educational, business, and even governmental workflows, AI systems will increasingly be called upon to act as collaborators with humans; to adequately fill this role, AIs must be able to recapitulate the reflection and deliberation that makes human-human collaboration successful, but also causes temporary slowdowns in dialogue while interlocutors construct a {\it common ground} on which to collectively reason---we will call this phenomenon {\bf friction}.

"Friction" in this sense is something that LLMs struggle with. To prompt an interlocutor to reflect upon their assumptions requires that one have an approximate understanding of what those assumptions are and entail~\cite{lewis2024reflective}. This is predicated upon a {\it theory of mind} (ToM; \citet{premack1978does}), which is likewise a challenge for LLMs~\cite{sap2022neural,ullman2023large}.

To address this, we present the {\bf \fricname~(\fricabbr)}, a novel approach to aligning LLMs to be adept {\it collaborators} in dialogue-driven tasks. Unlike common preference alignment approaches which focus predominantly on reward differences between textual surface forms to generate the best possible completions as a sequence of actions, \fricabbr~takes a state-driven approach based on the notion of a {\it frictive state}---a dynamic natural language representation that integrates task context and the beliefs of participants as they change over time (Fig.~\ref{fig:faaf-cartoon}). We use this state-wise representation to train "friction agent" models aligned to prompt collaborators toward reflection and deliberation in shared tasks, to help them resolve conflicting beliefs and assumptions that result in frictive states. Our results on two challenging collaborative task datasets and variants show that \fricabbr's belief state conditioning consistently produces output that is more relevant, impactful on the dialogue, and thought-provoking than competing methods. Our key contributions are:
\begin{itemize}
\vspace*{-1mm}
    \item a novel LLM alignment framework focused on generating outputs to support critical reasoning and assessments in a collaborative task environment;
\vspace*{-1mm}
    \item an in-depth mathematical and theoretical foundation for the above approach grounded in collaborative task dynamics;
\vspace*{-1mm}
    \item evaluations on three challenging collaborative task settings that show the advantages of \fricabbr~over competing alignment methods and demonstrate robustness to out-of-distribution (OOD) data.
\vspace*{-1mm}
\end{itemize}

\vspace*{-2mm}
\section{Related Work}
\label{sec:related}
\vspace*{-2mm}

RLHF-inspired preference alignment in LLMs has become a cornerstone of developing generative AI systems that cater to user preferences~\cite{stiennon2020learning}. Both "offline" approaches like Direct Preference Optimization (DPO;~\citet{rafailov2024direct}), Identity Preference Optimization (IPO;~\citet{azar2024general}) and other supervised methods~\citep{meng2024simposimplepreferenceoptimization, Hong2024ORPOMP, fisch2024robustpreferenceoptimizationreward, pal2024smaugfixingfailuremodes, nath2024simultaneous} and "online" methods~\citep{schulman2017proximal,pang2024iterative} focus predominantly on preference samples often sourced from datasets like Reddit TL;DR~\citep{volske2017tl} or Ultrafeedback~\citep{Cui2024UltraFeedbackBL} for algorithm development. 

These methods excel in generating summaries or completions that reflect human preferences including on single-turn human-AI interaction datasets like SGD~\cite{rastogi2020towards} or MultiWOZ~\cite{zang-etal-2020-multiwoz,ye-etal-2022-multiwoz}, but are often ill-equipped to handle the complexities of real-world multiparty interactions, where communication occurs across diverse modalities~\cite{krishnaswamy2018evaluation}, including sparse and ambiguous spoken dialogues between multiple collaborators~\cite{karadzhov2023delidata, khebour-etal-2024-common}.

A key challenge in these multiparty shared task settings is the scarcity of annotated data~\cite{bradford2023automatic}, particularly where interventions emerge contextually but sparsely~\cite{karadzhov2023delidata, khebour-etal-2024-common}. While preference data generated with AI feedback is a viable option~\citep{li2023controllabledialoguesimulationincontext, yuan2024self}, DPO-trained models depend crucially on the sampling or data-generating distribution due to its Bradley-Terry (BT) model of "implicit rewards," limiting their applications to dialogue-driven settings where preferences may be intransitive~\cite{tversky1969intransitivity} or change over time. This data-dependence holds even for more sophisticated methods that optimize on human utility~\cite{Ethayarajh2024KTOMA}, discard the BT assumption~\citep{azar2024general}, or use iterative online approaches~\cite{Rosset2024DirectNO, pang2024iterative, zheng2024toward}. Game-theoretic approaches to reduce this dependence focus on optimizing a "general preference model"~\citep{munos2023nash, calandriello2024human} that does \textit{not} suffer from this data-bias. But these have limited practical application due to their compute-intensive nature, often requiring the storage and computations with intermediate-stage policies during training~\cite{choi2024robust}. In contrast, \texttt{FAAF} avoids this data-dependence by explicitly conditioning policies on belief-misalignment in a specific dual alignment formulation which we derive in a simple "one-step" supervised manner without requiring computations of complicated mixture policies during training. \fricabbr~represents an instance of "frictive policy optimization" (FPO) as argued for by \citet{pustejovsky2025frictive}---specifically an instance of {\it Friction-Based Preference Pairing} (FPP).

\vspace*{-2mm}
\section{Definitions}
\label{sec:defs}
\vspace*{-2mm}

Let us first define key terms we rely on.

\vspace*{-2mm}
\paragraph{Frictive state}
Entailed by \citet{clark1996using}'s {\it common ground}, or the set of beliefs shared by interlocutors, a {\it frictive state} arises during a collaborative task when different interlocutors have contradictory beliefs about a task-relevant proposition (i.e., one believes $p$ and another sees evidence against $p$). This can be realized as a formal model of agent beliefs in an evidence-based dynamic epistemic logic \citep{van2014evidence,pacuit2017neighborhood}, or a natural language description thereof, as we use.
Different evidence leads to different predictions of future trajectories \cite{craik1943nature}. Thus frictive states, though sparse in dialogues, can critically delay or preclude success in a collaboration due to unresolved misunderstandings. The occurrence of a frictive state may not guarantee task failure, as the relevant propositions may be trivial to actual task completion. Therefore, in a {\it functionally frictive state}, the lack of common ground impedes progress on the task, or presents a significant risk of failure unless it is resolved.

\vspace*{-2mm}
\paragraph{Friction intervention}
Friction can indicate an impasse (the frictive state), but can also be used to resolve it, through a {\it friction intervention} that inserts into the dialogue indirect prompting to the participants to reevaluate their beliefs and incorrect assumptions or positions in light of available evidence \cite{oinas2009persuasive}, rather than accepting possibly erroneous presuppositions inherent in the dialogue. Importantly, a frictive intervention may be non-contradictory to the individual beliefs on display (i.e., neither asserting $p$ nor $\neg p$), but slows down the dialogue for reflection and deliberation, such as the {\it probing utterances} in \citet{karadzhov2023delidata} and \citet{nath2024any}. In the context of LLMs and \fricabbr, the {\it friction agent} constitutes a language model aligned toward the capacity to make frictive interventions.\footnote{We use \( \pi_f \) to denote the friction agent which generates high-quality interventions, but refer to it as the "optimal policy" for consistency with RLHF literature.} An ideal friction agent does \textit{not} intervene arbitrarily, which would cause distraction in collaborative tasks, but is conditioned to resolve the lack of common ground between human collaborators.

\vspace*{-2mm}
\section{Task Formulation and Background}
\label{sec:task-formulation}
\vspace*{-2mm}

Let $f$ be a frictive intervention (utterance) that is not required to contradict any particular belief encapsulated in a frictive state $\phi$, and let the human preference probability \( \mathcal{P}(f \succ \phi) \) be the probability that an expert annotator would prefer  \( f \) over maintaining \( \phi \), given prior dialogue history, $x$. An RLHF-based approach to LLM alignment toward an optimal policy \( \pi^*_f \) would assume a partition function $Z^*(\phi,x)$ that normalizes the probabilities of all possible responses (see \Cref{appendices:frictive_rlhf} for more details). While the optimal policy formulation is closed form, the dependence on $Z^*$ makes it practically intractable to estimate it for LLMs since $Z^*$ is a summation over the set of all possible sequences of  tokens in the tokenizer, often requiring methods like importance sampling~\citep{korbak2022reinforcement} or ensembling models~\citep{go2023aligning} for an unbiased estimate. This problem remains even if the set of friction interventions $\frictionspace$ were a restricted subset of the space of all possible actions $\actionspace$. To overcome this, prior RLHF and Preference-based RL~\cite{wirth2017survey} literature suggests supervised learning algorithms for obtaining an optimal policy induced \textit{under the expectation} over a preference dataset. These offline methods, such as DPO~\cite{rafailov2024direct}, IPO~\cite{azar2024general}, or Kahneman-Tversky Optimization (KTO;~\citet{Ethayarajh2024KTOMA}), either rely on the BT model of preferences \cite{BradleyTerry1952} where the optimal policy can be induced from a static preference dataset using implicitly-defined pointwise rewards, or assume that alignment is conducted with access to a non-biased data-generation or "sampling" distribution $\mu$ from which $\pi^*_f$ can be learned using pairwise preferences without adopting a strictly BT assumption~\cite{azar2024general}.\footnote{By "sampling," we mean those actions that make it to the preference annotation phase after being sampled with the data-generator $\mu$.} These approaches would give us the following formulation for $\pi^*_f$:

\vspace*{-6mm}

\begin{equation}
\small
   \pi_f^* = \frac{\pi_{\text{ref}} \exp \left(\beta^{-1} \mathbb{E}_{\substack{f \sim \mu(\cdot \mid x) \\ \phi \sim \mu(\cdot \mid x)}} \Psi(\mathcal{P}(f \succ \phi \mid x)) \right)}{Z^*(\phi, x)},
   \label{eq:optimal_policy_dpo_kto_ipo}
\end{equation}
where \( \Psi(p) \) is the identity mapping for IPO, and \( \log \left( \frac{\mathcal{P}}{1 - \mathcal{P}} \right) \) (inverse sigmoid) for DPO and KTO.

While the practicality of these supervised algorithms is a clear advantage, their dependence on preference data selected via sampling is a limitation in reconstructing the human preference probability $\mathcal{P}$. This is particularly true for collaborative dialogue tasks where common ground changes over time, meaning that the occurrence of frictive states is dynamic, and where participants may not intervene due to variables obscure to a language model, such as not realizing the existence of a frictive state or judging the frictive state to be {\it non-functional} (Sec.~\ref{sec:defs}). Operationally, even if the true underlying preferences ($\mathcal{P}$) of collaborators are transitive and consistent, constructing a preference dataset for use with existing offline training methods is not straightforward as dialogues may be skewed or sparse~\cite{khebour-etal-2024-common}. When using generative AI to create denser training data, even high-capacity LLMs like GPT-4 are prone to various forms of biases such as toward length~\citep{lambert2024rewardbench} or certain linguistic registers. Therefore, the core motivation of \texttt{\fricabbr} is as follows---\textit{how do we train a high-quality friction agent that can leverage the inherent scalability of offline alignment methods and reconstruct the true underlying preference distribution while still being robust to the data skew that may arise when sampling a preference dataset, whether using generative AI or from real-life collaborative dialogues?}

\vspace*{-1mm}
\subsection{\fricabbr~Objective}
\vspace*{-1mm}

We define a novel two-player adversarial optimization objective $J^*_{\text{\fricabbr}}$ (Eq.~\ref{eq:two_stage_main_objective_main_paper}). Specifically, given a reference model \( \pi_{\text{ref}} \) and a regularization parameter \( \beta \in \mathbb{R}_+ \), our goal is to learn two interdependent "collaborative" policies: (i) a \textit{frictive state policy} \( \pi_\phi^* \) that generates the most semantically rich frictive states \( \phi \), capturing tensions or uncertainties (in the form of first-order beliefs of dialog participants) in dialogue, and (ii) a \textit{friction intervention policy} \( \pi_f^* \) that generates constructive interventions \( f \), conditioned on the frictive state, to improve discourse clarity and converge onto a common ground between participants. Mathematically,

\vspace*{-4mm}
\begin{align}
    J^*_{\text{\fricabbr}} &= \min_{\pi_\phi} \max_{\pi_f} 
    \mathbb{E}_{
    \substack{x \sim \rho \\ \phi \sim \pi_\phi(\cdot \mid x) \\ f \sim \pi_f(\cdot \mid \phi, x)}
    } \Bigg[
    \mathcal{P}(f \succ \phi \mid x)  \notag \\
    &\quad - \beta D_{\text{KL}}(\pi_f \parallel \pi_{\text{ref}} \mid \phi, x)  \notag \\
    &\quad + \beta D_{\text{KL}}(\pi_\phi \parallel \pi_{\text{ref}} \mid x) 
    \Bigg].
    \label{eq:two_stage_main_objective_main_paper}
\end{align}

Notice how the optimal intervention policy $\pi^*_f$, by definition of the inner $\max$ operator, generates interventions that are, on average, most preferred by collaborators, while the first KL-divergence term, defined as $D_{\text{KL}}(\cdot \mid \phi, x)$, stabilizes learning in $\pi^*_f$ by keeping it closer to a reference model. Unlike the standard RLHF objective as would be required for a BT model loss, the \fricabbr~loss contains no sigmoid term. Compared to a standard RLHF objective, the additional KL term $D_{\text{KL}}(\pi_\phi \parallel \pi_{\text{ref}} \mid x)$ forces the frictive state policy \( \pi_\phi^* \) to be adversarially robust, in that it must ensure that sampled frictive states $\phi$ $\sim$ \( \pi_\phi^* \) cannot be exploited by $\pi^*_f$ to generate subpar interventions that remain too close to the reference model. Thus, \fricabbr~serves as an agent policy that adapts to dialogues over time: the frictive state policy searches for the most immediate tension points or exposes the lack of common ground between task participants, while the intervention policy generates outputs that remain grounded in the particulars of the relevant frictive state (e.g., regarding the correct task items or propositions), and naturally and intuitively prompts for reflection and deliberation on these points. \textbf{The key takeaway is that optimal friction interventions should not be arbitrary interventions in the dialogue, but should surface the presuppositions that gave rise to the most logically necessary frictive state, making interventions precise and interpretable.}

\vspace*{-1mm}
\subsection{Dataset Annotation and Generation}
\label{ssec:datasets}
\vspace*{-1mm}

In Sec.~\ref{sec:related}, we discuss why common preference optimization datases such as Ultrafeedback, Reddit TL;DR, SGD, or MultiWOZ are not appropriate for \fricabbr's collaborative task use case. Therefore, we consider two collaborative task datasets to evaluate \fricabbr---DeliData~\citep{karadzhov2023delidata} and the Weights Task Dataset (WTD;~\citet{khebour2024text}). These datasets also exemplify the data sparsity problem with deliberation and friction in collaboration (Sec.~\ref{sec:task-formulation}). 

\textbf{DeliData} contains dialogues from 500 groups of 5 attempting the Wason Card task~\cite{wason1968reasoning}, which involves reasoning about if a card with a specific characteristic (e.g., even number on one side) must have a different characteristic (e.g., a vowel on the other). \citet{karadzhov2023delidata} annotated DeliData with "probing" interventions, or naturally-occurring friction that prompts for reasoning and deliberation without introducing new information. However, these amount to an average of only 3.46 probing interventions per group, out of 17,110 total utterances. 

\textbf{WTD} is an audiovisual dataset of 10 triads collaborating to deduce the weights of differently-colored blocks and infer the pattern describing them, and is similarly sparse. We annotated WTD for naturally-occurring friction given a definition following \citet{oinas2009persuasive}.\footnote{Frictive interventions in this setting act as indirect persuasion \cite{oinas2009persuasive} where participants are passively prompted to reevaluate their beliefs and assumptions or propositions, in light of incoming goal-specific evidence. See \Cref{appendices:friction_def} for the complete definition.} Two annotators annotated half the groups each while a third annotated all 10. They then collectively adjudicated each annotation following the definition. Cohen's $\kappa$ between initial and final annotations was 0.632, indicating substantial agreement. An average of 4 naturally-occurring friction interventions per group were found in the WTD. 

The individual dialogues in each dataset are quite long, numbering in the thousands of utterances (WTD, for instance, comprises almost 3 hours of audio-visual data~\cite{khebour2024text}. This contrasts with other preference alignment data, in which, although there are many more individual samples, each tends to be shorter due to the nature of typical preference alignment tasks, such as article summarization. At the utterance level, the collaborative task datasets we use number in the thousands to tens of thousands of utterances, roughly equivalent in size to a dataset like Ultrafeedback~\cite{Cui2024UltraFeedbackBL}, and indicating the distinct nature of collaborative task data.

\vspace*{-1mm}
\paragraph{Training Dataset Construction}

This extreme sparsity does not capture anything close to the possible frictive states available in the combinatorics of the problem space, and so motivated the need for data augmentation to construct sufficiently diverse preference datasets for training and evaluation. We used GPT-4o as a high-capacity LLM for our sampling distribution $\mu$. We used a {\it self-rewarding} approach \cite{yuan2024self} to simultaneously generate candidate interventions (and their rationales) and assign them rewards, which naturally induced an implicit preference ranking. We provided GPT-4o with sequences of $h$ utterances from each dialogue in the two datasets, and prompted it to label frictive states and generate friction interventions following colloquial renderings of the definitions in Sec.~\ref{sec:defs}.\footnote{$h$ was set to 15 for DeliData and 10 for WTD. The prompts showing how frictive states are rendered into plain text are given in Figs.~\ref{fig:deli_friction_generation_prompt} and \ref{fig:wtd_friction_generation_prompt} in \Cref{appendices:generation}.} Finally, we conducted contrastive pairing of "winning" and "losing" interventions $f_w$ and $f_l$ with the corresponding dialogue history $x$ to construct the final preference datasets for each task, comprising tuples of $x$, frictive state $\phi$, $f_w$, and $f_l$.

For each dataset, we conducted additional task-appropriate augmentation. For DeliData, we constructed alternative tuples where the specific cards mentioned in the original data were replaced with other cards of the same classes that preserved the relevant rule (e.g., replacing even numbers with other even numbers, consonants with other consonants, etc.). This resulted in 68,618 preference samples for training, with average $\mu$-assigned reward for preferred samples of 8.03 and for dispreferred samples of 3.96 (out of 10). We held out 50 randomly-sampled dialogues for testing.

Since the original WTD contains only 10 dialogues, holding one or two out for evaluation would adversely impact the data distribution. Therefore we used \citet{shani2024multiturnreinforcementlearningpreference}'s method to generate novel simulated collaborative conversations about the Weights Task, providing a task descriptions and ground-truth values for the weights. GPT-4o was prompted to role-play personality-facet
combinations from the Big 5 personality types \cite{goldberg2013alternative}, and for each labeled frictive state $\phi$ we generated and scored 6 friction interventions. This resulted in two distinct versions of the WTD preference dataset. The {\bf Simulated WTD} friction dataset consisted of 56,698 training preference samples, with mean scores of 8.48 (preferred interventions) and 6.01 (dispreferred). 54 dialogues were held out for testing. The {\bf Original WTD} friction dataset (see above) contained 4,299 preference samples (preferred mean score 8.36, dispreferred 6.35). These were \textit{all} retained for an OOD evaluation of \fricabbr~trained on the Simulated WTD data. See \Cref{appendices:generation} for specifics of data generation and \Cref{app:wtd_distributional_analysis} for a distributional analysis of the Original and the Simulated WTD data.

\vspace*{-1mm}
\paragraph{Human Validation}
We conducted a human evaluation to assess the quality of the GPT-generated friction intervention on a random representative subset of 50 pairwise samples each from both the DeliData and WTD generated test datasets.\footnote{WTD samples include both Original and Simulated interventions.} For each sample, 2 annotators were asked to choose which of the two candidate interventions was more appropriate for provoking participants' reflection to help them advance in their task without being given the solution. Average Cohen's $\kappa$ on WTD samples was 0.58 and on DeliData samples was 0.92, indicating substantial to near complete agreement on which was the better intervention, and indicates that the preferred/dispreferred friction distinction sourced from GPT-4o as $\mu$ aligns with human judgments. See \Cref{appendices:human_val} for more.






\vspace*{-1mm}
\subsection{Deriving the Empirical \fricabbr~Loss}
\label{ssec:faaf_practical_algo_datasets}
\vspace*{-1mm}

While the data is constructed using a standard pairwise preference format, the \fricabbr~{\it optimization} conditions upon the dialogue context $x$ {\it and} textual rendering of the frictive state $\phi$. To derive an empirical offline (supervised) preference learning loss from the two-player objective (Eq.~\ref{eq:two_stage_main_objective_main_paper}), we use a divide-and-conquer approach. Deriving the inner maximization loop of Eq.~\ref{eq:two_stage_main_objective_main_paper} results in an analytical expression of the optimal frictive intervention policy, $\pi_f^*$ (see \Cref{appendices:optimal_friction}, Eq.~\ref{eq:optimal_frictive_agent_policy}). However, we observe that $\pi_f^*$ in its analytical form (Eqs.~\ref{eq:optimal_policy_dpo_kto_ipo} and \ref{eq:optimal_frictive_agent_policy}) is not fully expressive since it does \textit{not} contain the optimal frictive-state policy $\pi_\phi^*$ term. Therefore, we derive $\pi_\phi^*$ using a Lagrangian formulation (see \Cref{sec:optimal_frictive_state_derivation} for details)
that expresses the preference for any intervention $f_1$ over $f_2$ analytically in terms of \textbf{both} the optimal friction intervention policy ($\pi_f^*(\cdot \mid \phi, x)$) and the optimal frictive-state policy ($ \pi_\phi^*(\cdot|x)$). 
This allows us to use a straightforward supervised ($\ell_2$) objective—similar in spirit to IPO~\cite{azar2024general}—that empirically regresses the predicted preference expression derived from $\pi_f^*(\cdot \mid \phi, x)$ and $\pi_\phi^*(\cdot|x)$ to the observed relative preferences $p(f_1 \succ f_2 \mid x)$ (relative to $\phi$), assuming access to a large enough preference-annotated dataset of friction interventions. Notably, this objective is optimized by a \textit{single} parametrized policy that leverages the inherent expressivity of LLMs and induces a unique global minimum in the space of policies (see \Cref{theorem:faaf_uniqueness} in \Cref{appendices:optimal_friction}). \Cref{alg:friction_agent_training} shows the full training algorithm.\footnote{For compactness reasons here we represent all policies $\pi$ as parameterized by weights $\theta$. Similarly to approaches such as \citet{choi2024robust}, because we formulate two distinct policies with the preference equation, we can empirically enforce it using $\ell_2$ loss and learn it with a single expressive policy parameterized by $\theta$.}
 
\begin{algorithm}
\small
\caption{Frictional Agent Alignment Framework}
 \label{alg:friction_agent_training}
\begin{algorithmic}[1]
\Require Training data $\mathcal{D}_\mu$ containing tuples $(x, \phi, f_w, f_l)$, where $x$: prompt, $\phi$: frictive state, $f_w$: preferred response, $f_l$: non-preferred response.

\State Define likelihood ratios:
\\ $\Delta R = \log\left(\frac{\pi_\theta(f_w|\phi,x)}{\pi_{\text{ref}}(f_w|\phi,x)}\right) - \log\left(\frac{\pi_\theta(f_l|\phi,x)}{\pi_{\text{ref}}(f_l|\phi,x)}\right)$
\\ $\Delta R' = \log\left(\frac{\pi_\theta(f_w|x)}{\pi_{\text{ref}}(f_w|x)}\right) - \log\left(\frac{\pi_\theta(f_l|x)}{\pi_{\text{ref}}(f_l|x)}\right)$

\State Loss function:
$\mathcal{L} = \mathbb{E}_{\mathcal{D}_\mu} [(1 - \beta(\Delta R + \Delta R'))^2]$

\State Gradient update:
$\nabla_\theta \mathcal{L} = \mathbb{E}_{\mathcal{D}_\mu} [-2\beta\delta \nabla_\theta\log(\Delta R \cdot \Delta R')]$, where $\delta = 1 - \beta(\log\Delta R + \log\Delta R')$

\State Update policy parameters $\theta$ using gradient descent
\end{algorithmic}
\end{algorithm}

\vspace*{-2mm}
\section{Experimental Setup}
\label{sec:exp}
\vspace*{-2mm}

\paragraph{Training Setup and Baselines}

We use \texttt{Meta-Llama-3-8B-Instruct}~\citep{llama3modelcard}\footnote{\url{https://huggingface.co/meta-llama/Meta-Llama-3-8B-Instruct}} for \textit{all} experiments including baselines. All aligned models received exposure to the frictive state annotations during training to ensure fair comparisons on the friction intervention task.  For an in-depth evaluation of \fricabbr's capabilities, we include comparisons to the Supervised-Finetuned (SFT) model as well as the base instruct model generations in our experiments. For "offline" contrastive approaches to compare to, we choose DPO~\citep{rafailov2024direct} and IPO~\citep{azar2024general} and for "online" approaches, we include Proximal Policy Optimization (PPO;~\citet{schulman2017proximal}) baseline. For SFT, we employ rejection sampling~\citep{xu2023some} to maximize the likelihood of interventions that receive high rewards under $\mu$. For SFT, DPO, and IPO, the respective losses are computed only on the output tokens and frictive states $\phi$, excluding dialogue context tokens. This training approach ensures that the models learn to generate effective interventions while maintaining contextual understanding. For PPO, we train an OPT 1.3B~\citep{zhang2022opt} reward model on each dataset using a standard Bradley-Terry loss~\citep{stiennon2020learning} over preference pairs. For \textit{ablations}, we consider variants of \fricabbr~that ablate the different likelihood ratios---\fricabbr$_{\Delta R}$ keeps \textit{only} the $\phi$-conditioned implicit rewards in the \fricabbr~objective (line 2 in \Cref{alg:friction_agent_training}), and \fricabbr$_{\Delta R'}$ removes $\phi$-conditioning (keeping only line 3). See \Cref{app:hyperparameters} for more details on training and hyperparameters. At inference time, all models only receive the dialogue history up until the point at which an intervention is generated, and receive no look-ahead.

\vspace*{-1mm}
 \paragraph{Evaluation Strategies}

As LLM generation is open-ended, we employ an LLM-as-a-judge (using GPT-4o) "win-rate" evaluation method where a high-capacity model is prompted to select its preference, given two completions, and conducts a multidimensional evaluation of its preferences.

First, we sampled friction interventions from all competing models on 500 randomly sampled prompts from the \textbf{DeliData}, \textbf{Simulated WTD} and \textbf{Original WTD} test sets. Next, we conducted two evaluations using said completions, one with a \textbf{preference-model}~\citep{munos2023nash} and another with a \textbf{reward-model}~\citep{Hong2024ORPOMP}. Since GPT-4o also served as the data generation distribution $\mu$, preference-model evaluation compares the two presented choices and nothing else in the data, mitigating lingering bias toward $\mu$ \citep{munos2023nash}.

Within preference-based evaluation settings, we adopt the framework proposed by \citet{Cui2024UltraFeedbackBL} to retrieve utility scores across seven friction dimensions, building on insights from \citet{chen2024exploringbehavioralmodelpositive}. Specifically, we assess \textit{relevance} and \textit{alignment with rationale and golden samples}\footnote{A subset of these golden friction interventions was used for human evaluations (see~\Cref{appendices:human_val}).} to determine how well a friction intervention aligns with surface-level semantics. Meanwhile, \textit{actionability}, \textit{specificity}, \textit{thought-provoking}, and \textit{impact} measure its expected long-term influence on behavior, reasoning, and decision-making. The LLM-judge assigns Likert-type scores across these dimensions, providing a fine-grained evaluation of task-specific preference desiderata. These scores are collected in a pairwise fashion where $\pi_\theta$-generated interventions $f_i$ from a baseline are compared with $\pi_\text{ref}$-generated counterparts, $f_j$. We positionally swap these interventions in the evaluation prompt (Fig.~\ref{fig:friction_eval_prompt}) for each API call and average the scores for each of the seven dimensions over two runs to mitigate positional bias~\citep{lambert2024rewardbench} in computing the final win rates. Specifically, for any pair of interventions $(f_i, f_j)$, let $s(x, f_*)$ denote the score estimate\footnote{In this evaluation, "overall" (first column in Table~\ref{tab:full_results_gpt_evals_table}) is computed based on the judge's choice of winner {\it after} rating all other dimensions. As such, $s(x, f_*)$ represents scores over these fine-grained friction preference desiderata, and "overall" does not necessarily represent an average or aggregate of the other dimensions but rather a binary judgment based on them.} for intervention $f_*$ given context $x$. The win-rate percentage for a run is computed as 
$100 \times \frac{1}{N} \sum_{m=1}^{N} \mathbf{1}\{s(x^{(m)}, f^{(m)}_i) > s(x^{(m)}, f^{(m)}_j)\}$, where $N$ is the total number of samples, and $x^{(m)}$ represents the context of the $m^\text{th}$ sample. See Table~\ref{tab:full_results_gpt_evals_table}.

The above evaluation tests for preference alignment advantage of the aligned model over $\pi_\text{ref}$. For a more robust evaluation, we compare \fricabbr's generations "head-to-head" against all baselines. Here, instead of the preference model, we utilize the trained OPT 1.3B Reward Model (RM) as described in our PPO training setup. These pointwise estimates of rewards provide a more accurate assessment of the advantage provided by \fricabbr~proposed approach when directly pitted against other alignment baselines. Specifically,  we compare \fricabbr$_{\Delta R}$, \fricabbr$_{\Delta R'}$  as well as our full objective baseline (\fricabbr$_{\Delta(R+R')}$) against all chosen baselines. We compute the reward accuracy (or win-rates) similarly and report our results in Table~\ref{tab:opt_rm_performance_comparison_percentage}.

\vspace*{-2mm}
\section{Results and Analysis}
\label{sec:results}
\vspace*{-2mm}

\begin{table*}[h!] 
\centering
\small
\begin{tabular}{lcccccccc}
 
\toprule
\textbf{Policy} & \textbf{Overall} & \textbf{Ac} & \textbf{Ga} & \textbf{Im} & \textbf{Rf} & \textbf{Re} & \textbf{Sp} & \textbf{Th} \\
\cmidrule(lr){2-9}
\multicolumn{9}{c}{\textsc{DeliData}} \\
\cmidrule(lr){1-9}

PPO & 68.9$_{\pm 1.5}$ & 59.9$_{\pm 1.5}$ & 65.4$_{\pm 1.5}$ & 68.6$_{\pm 1.5}$ & 64.9$_{\pm 1.5}$ & 65.1$_{\pm 1.5}$ & 71.1$_{\pm 1.4}$ & 64.0$_{\pm 1.5}$ \\
\rowcolor{gray!10}IPO & 70.1$_{\pm 1.4}$ & 61.2$_{\pm 1.5}$  & 65.7$_{\pm 1.5}$ & 69.3$_{\pm 1.5}$ & 65.3$_{\pm 1.5}$ & 65.5$_{\pm 1.5}$ & 72.1$_{\pm 1.4}$ & 64.1$_{\pm 1.5}$ \\
DPO & 70.8$_{\pm 1.4}$ & 61.0$_{\pm 1.5}$  & 66.8$_{\pm 1.5}$ & 69.6$_{\pm 1.5}$ & 66.1$_{\pm 1.5}$ & 67.5$_{\pm 1.5}$ & 72.2$_{\pm 1.4}$ & 66.2$_{\pm 1.5}$ \\
\rowcolor{blue!15}\fricabbr & \textbf{75.7}$_{\pm 1.4}$ & \textbf{65.6}$_{\pm 1.5}$  & \textbf{69.5}$_{\pm 1.5}$ & \textbf{75.0}$_{\pm 1.4}$ & \textbf{72.0}$_{\pm 1.4}$ & \textbf{71.1}$_{\pm 1.4}$ & \textbf{75.3}$_{\pm 1.4}$ & \textbf{70.4}$_{\pm 1.4}$ \\
\midrule

\multicolumn{9}{c}{\textsc{WTD Original}} \\ 
\cmidrule(lr){1-9}
PPO & 76.0$_{\pm 4.3}$ & 74.0$_{\pm 4.4}$ & 75.0$_{\pm 4.3}$ & 75.0$_{\pm 4.3}$ & 67.0$_{\pm 4.7}$ & 70.0$_{\pm 4.6}$ & 73.0$_{\pm 4.4}$ & 74.0$_{\pm 4.4}$ \\
\rowcolor{gray!10}IPO & 82.0$_{\pm 3.8}$ & 87.0$_{\pm 3.4}$ & 75.0$_{\pm 4.3}$ & 84.0$_{\pm 3.7}$ & 75.0$_{\pm 4.3}$ & 80.0$_{\pm 4.0}$ & 88.0$_{\pm 3.2}$ & 78.0$_{\pm 4.1}$ \\
DPO & 89.0$_{\pm 3.1}$ & \textbf{92.0}$_{\pm 2.7}$ & 82.0$_{\pm 3.8}$ & 89.0$_{\pm 3.1}$ & 84.0$_{\pm 3.7}$ & 87.0$_{\pm 3.4}$ & \textbf{89.0}$_{\pm 3.1}$ & 79.0$_{\pm 4.1}$ \\
\rowcolor{blue!15}\fricabbr & \textbf{90.9}$_{\pm 2.9}$ & 81.8$_{\pm 3.9}$ & \textbf{84.8}$_{\pm 3.6}$ & \textbf{90.9}$_{\pm 2.9}$ & \textbf{86.9}$_{\pm 3.4}$ & \textbf{89.9}$_{\pm 3.0}$ & 88.9$_{\pm 3.1}$ & \textbf{90.9}$_{\pm 2.9}$ \\
\midrule

\multicolumn{9}{c}{\textsc{WTD Simulated}} \\
\cmidrule(lr){1-9}
PPO & 73.6$_{\pm 1.5}$ & 69.7$_{\pm 1.5}$ & 64.9$_{\pm 1.6}$ & 74.2$_{\pm 1.5}$ & 67.6$_{\pm 1.6}$ & 71.9$_{\pm 1.5}$ & 78.1$_{\pm 1.4}$ & 78.3$_{\pm 1.4}$ \\
\rowcolor{gray!10}IPO & 83.0$_{\pm 1.3}$ & 74.8$_{\pm 1.4}$ & 78.4$_{\pm 1.4}$ & 82.9$_{\pm 1.3}$ & 76.9$_{\pm 1.4}$ & 81.4$_{\pm 1.3}$ & 82.5$_{\pm 1.3}$ & 83.2$_{\pm 1.2}$ \\
DPO & 82.9$_{\pm 1.3}$ & 80.4$_{\pm 1.3}$ & 75.8$_{\pm 1.4}$ & 81.3$_{\pm 1.3}$ & 72.9$_{\pm 1.5}$ & 76.3$_{\pm 1.4}$ & 80.2$_{\pm 1.3}$ & 79.2$_{\pm 1.4}$ \\
\rowcolor{blue!15}\fricabbr & \textbf{91.5}$_{\pm 0.9}$ & \textbf{87.5}$_{\pm 1.1}$ & \textbf{87.1}$_{\pm 1.1}$ & \textbf{90.1}$_{\pm 1.0}$ & \textbf{82.0}$_{\pm 1.3}$ & \textbf{85.1}$_{\pm 1.2}$ & \textbf{90.3}$_{\pm 1.0}$ & \textbf{90.1}$_{\pm 1.0}$ \\
\bottomrule
\end{tabular}
\vspace*{-2mm}
\caption{Win-rates (\%) against the SFT model ($\pi_\text{ref}$) for all alignment methods on sampled interventions (temperature of 0.7, top-$p$ of 0.9) from 500 randomly-sampled prompts from DeliData and WTD evaluation sets, according to GPT-4o. Metrics: \textbf{Ac} (\textit{Actionability}), \textbf{Ga} (\textit{Gold-alignment}), \textbf{Im} (\textit{Impact}), \textbf{Rf} (\textit{Rationale-fit}), \textbf{Re} (\textit{Relevance}), \textbf{Sp} (\textit{Specificity}), and \textbf{Th} (\textit{Thought-provoking}). The LLM-as-a-judge evaluation follows~\citet{Cui2024UltraFeedbackBL}. Average win rates are reported over two runs, with positional swapping to mitigate position bias.}
\label{tab:full_results_gpt_evals_table}
\vspace*{-2mm}
\end{table*}

\begin{table*}[h!]

\centering
\small
\resizebox{\textwidth}{!}{
\begin{tabular}{llccccc}
\toprule
       \textbf{Dataset} & \textbf{Policy} & Win-rate \textbf{vs. Base} & Win-rate \textbf{vs. SFT} & Win-rate \textbf{vs. DPO} & Win-rate \textbf{vs. IPO} & Win-rate \textbf{vs. PPO} \\

       \midrule
       {DeliData} 
       
       & \fricabbr$_{\Delta R'}$    & 82.2$_{\pm 1.7}$ & 78.8$_{\pm 1.8}$ & 74.0$_{\pm 1.9}$ & 53.6$_{\pm 2.2}$ & \textbf{79.2}$_{\pm 1.8}$ \\
       
       & \fricabbr$_{\Delta R}$    & 85.8$_{\pm 1.5}$ & 81.4$_{\pm 1.7}$ & 73.2$_{\pm 1.9}$ & 54.2$_{\pm 2.2}$ & 73.4$_{\pm 1.9}$ \\
       
      & \fricabbr$_{\Delta(R+R')}$ & \textbf{86.2}$_{\pm 1.5}$ & \textbf{84.0}$_{\pm 1.6}$ & \textbf{75.6}$_{\pm 1.9}$ & \textbf{79.6}$_{\pm 1.8}$ & 76.0$_{\pm 1.9}$ \\
       
       \midrule
       {WTD Orig.} 
       & \fricabbr$_{\Delta R'}$   & 78.0$_{\pm 5.8}$ & \textbf{78.0}$_{\pm 5.8}$ & \textbf{76.0}$_{\pm 6.0}$ & 58.0$_{\pm 6.9}$ & 58.0$_{\pm 6.9}$ \\
       
       & \fricabbr$_{\Delta R}$   & 68.0$_{\pm 6.5}$ & 74.0$_{\pm 6.2}$ & 72.0$_{\pm 6.3}$ & 62.0$_{\pm 6.8}$ & 70.0$_{\pm 6.4}$ \\
       
     & \fricabbr$_{\Delta(R+R')}$ & \textbf{84.0}$_{\pm 5.1}$ & 76.0$_{\pm 6.0}$ & 74.0$_{\pm 6.2}$ & \textbf{74.0}$_{\pm 6.2}$ & \textbf{82.0}$_{\pm 5.4}$ \\
            \midrule

     {WTD Sim.} 
       & \fricabbr$_{\Delta R'}$   & 79.1$_{\pm 1.9}$ & 80.2$_{\pm 1.8}$ & 70.4$_{\pm 2.1}$ & 68.6$_{\pm 2.1}$ & 60.8$_{\pm 2.3}$ \\
       
       & \fricabbr$_{\Delta R}$   & 85.7$_{\pm 1.6}$ & 80.8$_{\pm 1.8}$ & 70.8$_{\pm 2.1}$ & 72.2$_{\pm 2.1}$ & 74.8$_{\pm 2.0}$ \\
       
    & \fricabbr$_{\Delta(R+R')}$ & \textbf{88.0}$_{\pm 1.5}$ & \textbf{83.7}$_{\pm 1.7}$ & \textbf{72.8}$_{\pm 2.0}$ & \textbf{73.7}$_{\pm 2.0}$ & \textbf{75.1}$_{\pm 2.0}$ \\
       
       \bottomrule
\end{tabular}}
\vspace*{-2mm}
\caption{Win rates of of \fricabbr~variants---\fricabbr$_{\Delta R'}$ (not $\phi$-conditioned), \fricabbr$_{\Delta R}$ ($\phi$-conditioned), and \fricabbr$_{\Delta(R+R')}$ (full objective)---against competing methods in pairwise comparisons (temperature of 0.7, top-$p$ of 0.9). All alignment baselines are SFT-initialized and \texttt{Meta-Llama-3-8B-Instruct} is used as Base.}

\label{tab:opt_rm_performance_comparison_percentage}
\vspace*{-4mm}
\end{table*}

Table~\ref{tab:full_results_gpt_evals_table} shows that in the eyes of the the LLM-judge, \fricabbr~models have a consistently greater advantage over the SFT model $\pi_\text{ref}$ than other baselines across the 7 preference dimensions and overall, noting that when competitor methods beat \fricabbr, the advantage is usually within the margin of error. For instance, in "overall" preference on the DeliData test samples, \fricabbr~achieves a 75.7\% win-rate over $\pi_\text{ref}$, surpassing PPO (68.9\%), DPO (70.8\%), and IPO (70.1\%). On the WTD datasets, win rates for all models are higher, reflecting $\pi_\text{ref}$'s weakness with the underspecified nature of WTD dialogues; alignment on this data has a greater net effect on win-rates than the generally less ambiguous DeliData. On WTD \fricabbr~is a clear all-around winner, at 90.9\% (vs. DPO's 89.0\% and 82.0\%) and 91.5\% (vs. DPO's 82.9\% and IPO's 83.0\%) on the Original and Simulated WTD datasets, respectively.

We find that \fricabbr's win-rates on dimensions such as \textit{actionability} and \textit{gold-alignment} are somewhat lower compared to other dimensions---possibly reflecting that multiple kinds of interventions may be appropriate in context. However, across dimensions like \textit{thought-provoking} and \textit{rationale-fit} we find that \fricabbr~improves 5-6\%, or even up to 12\% over equivalent PPO, IPO, and DPO win-rates.
PPO's win-rates consistently lag across all datasets (this is particularly pronounced on the WTD data), indicating the challenge that the dimensions of friction pose for a standard approach. DPO is typically \fricabbr's closest competitor against $\pi_\text{ref}$, with the narrowest average gap in win-rates.


\vspace*{-1mm}
\paragraph{Robustness to OOD Generalization} 

\fricabbr~maintains superior performance on the \textbf{Original WTD} dataset (Overall: +1.9\% over DPO, +8.9\% over IPO, and +14.9\% over PPO). No model was explicitly aligned to this data, and so this result shows \fricabbr's robustness to OOD settings compared to other approaches. This is particularly noteworthy given that the Original WTD dataset comprises word-for-word transcriptions of actual human dialogues---with disfluencies, sentence fragments, etc.---which differs markedly from the grammatical, structured text typically found in LLM training data or the preference pair samples in the \textbf{Simulated WTD} data. That \fricabbr~generalizes well to organic human data provides a strong basis of confidence that a \fricabbr-aligned agent, jointly-conditioned on the dialogue transcript and frictive state rendering $\phi$, could effectively intervene in and mediate real collaborations, where dialogues are often sparse, informal, and structurally distinct from LLM-generated text~\cite{martins2020sparsetextgeneration}.
DPO, although also optimized against $\phi$ as part of the context (Sec.~\ref{sec:exp}), suffers from the Longest-common-subsequence problem~\citep{pal2024smaugfixingfailuremodes}\footnote{The LCS issue in DPO, where gradient signals from tokens shared by winning and losing responses are ignored, is well-studied~\citep{pal2024smaugfixingfailuremodes, zhang2024chain, rafailov2024r}.} due to the Bradley-Terry preference model assumption where dependence on the context via DPO's log-partition term is effectively canceled in gradient estimates. In contrast, \fricabbr's combined $\Delta R$ and $\Delta R'$ regularization (\Cref{alg:friction_agent_training}) avoids missing such signals in its learning, thereby allowing it to capture more nuanced human preferences. 

While the multidimensional analysis on dimensions such as {\it impact} and {\it actionability} does not test actual human responses to the sampled intervention, it does maintain evaluation under consistent conditions without creating counterfactual branching due to responses generated under the influence of interventions, which would create divergent evaluation conditions. See \Cref{app:eval-rationale} for more discussion.



\vspace*{-1mm}
\paragraph{Does $\phi$-conditioning help \fricabbr~learn more accurate preferences?}
Table~\ref{tab:opt_rm_performance_comparison_percentage} shows results from the trained OPT-1.3B RM's evaluation of the full \fricabbr$_{\Delta(R+R')}$ objective and its ablated variants---$\phi$-conditioned \fricabbr$_{\Delta R}$ and unconditioned \fricabbr$_{\Delta R'}$---"head-to-head" against all baselines, including the base \texttt{Meta-Llama-3-8B-Instruct} model. Across the three datasets, \fricabbr~win-rates computed with pointwise reward estimates on sampled interventions exceed 80\%, on average, against the base and SFT models, consistent with prior work~\citep{Hong2024ORPOMP}. We also find that while explicit conditioning on $\phi$ provides clear advantages (e.g., +6.6\% vs. Base on Simulated WTD, +14\% vs. PPO), and even the unconditioned version consistently wins over baselines, neither term alone achieves the robust performance of \fricabbr$_{\Delta(R+R')}$.

Both IPO and \fricabbr~use a squared $\ell_2$ loss. IPO's performance against \fricabbr's ablations suggests that this structural similarity makes it more competitive with \fricabbr~(\fricabbr~ablations beat IPO 53.6\% and 54.2\% on DeliData and 58.0\% and 62.0\% on Original WTD, compared to anywhere from a 68--85\% win rate against the Base model). In general, these ablations demonstrate that neither variant alone is sufficient. \fricabbr$_{\Delta(R+R')}$ (the full objective) shows consistently stronger performance against IPO (79.6\% on DeliData, 73.7\% on WTD Sim., 74.0\% on WTD Orig.) while maintaining high win rates across other baselines ($\sim$81\% vs Base/SFT, $\sim$74\% vs. DPO). These results, in light of the trends observed previously in OOD evaluation, suggest that while \fricabbr$_{\Delta R}$ learns rich $\phi$-conditioned preferences, the additional regularization term $\Delta R'$ enables better reward space exploration and generalized preference learning. The combination is crucial for robust performance.



 

\paragraph{Hyperparameter Ablations}
We report ablations on $\beta$ in Fig.~\ref{fig:wtd_training_beta_metrics} (\Cref{app:hyperparameters}). Greater values of $\beta$ lead toward greater implicit reward for winning interventions, stability, convergence, and ability to distinguish preferences. A lower $\beta$ (5 or lower) tends to increase implicit reward at first but the policy degrades quickly, and ends up assigning low likelihood for the actual winning samples.

\vspace*{-2mm}
\section{Future Work} 
\label{sec:fw}
\vspace*{-2mm}

Real humans are infamous for flummoxing the most theoretically-rigorous AI systems and so the performance of \fricabbr~(or any other alignment method) in a real multiparty collaborative setting remains an open question. \fricabbr~provides a theoretically-grounded and empirically-validated basis of confidence for success. We have focused on the alignment technique in this paper (and thus framed this paper as a preference alignment paper), and demonstrated feasibility on challenging collaborative task datasets, but human user studies, e.g., using \citet{vanderhoeven2025trace}'s platform for real-time common ground and multimodal task tracking, remain the topic of future work. Formal or hybrid approaches such as the Common Ground Tracking that originally motivated the Weights Task~\cite{khebour-etal-2024-common} or the associated propositional extraction approach~\cite{venkatesha2024propositional,venkatesha2025propositional} could be used to validate the inferred frictive state description before it is used for generation.

Excessive introduction of friction could bring dialogue and collaboration to a halt, and a poorly-aligned agent could insert misleading or off-topic friction (some examples of this occur in DPO, PPO, and SFT responses in Tables~\ref{tab:model_comparison_wtd_original} and \ref{tab:model_comparison_deli1} in \Cref{app:prompts-samples}) and derail task progress. The pragmatics of when and how to intervene in an interaction remains a challenging open problem. Some failsafes may include including a symbolic, interpretable planning or constraint satisfaction approach in the interaction loop that would only allow friction to be inserted if it determines the task to be at risk of failure, or even limiting interventions to at most every $N$ utterances for an appropriate value of $N$.

\vspace*{-2mm}
\section{Conclusion} 
\label{sec:conc}
\vspace*{-2mm}

\fricabbr~introduces a novel perspective on LLM alignment, focusing on the problem of generating outputs that elicit reasoning and reexamination of assumptions and evidence in a collaborative context.  This critical capacity can help avert collaboration failure due to groups or individuals proceeding hastily according to their own preconceptions~\cite{koschmann2016communicative}, such that a fragile common ground collapses. We proposed a novel two-player objective with an analytical form that can be optimized using a single policy (Sec.~\ref{sec:task-formulation}). Through evaluations on three datasets representing two different collaborative tasks, and with detailed ablations (Sec.~\ref{sec:results}), we showed that \fricabbr~bests other common preference alignment methods in performance against a reference model, and that \fricabbr's simultaneous conditioning on both the frictive state $\phi$ and surface context $x$ is critical to its success.

In the process, we also put forth operational definitions of "friction" in human-AI collaboration (Sec.~\ref{sec:defs}). Friction creates opportunity for negotiation of intents toward a common goal, and space for accountability and collaborative reasoning. These moments may result in a net slower interaction, but are critical to eventual task success. The study of friction has broad applicability to fields like discourse studies, team science, and education \cite{sonneland2019friction,collins2024modulating,sutton2024friction}, and is something we believe the NLP community would do well to invest effort in. Counter to AI being sold as a speed and efficiency multiplier, our formulation of alignment to specialize in friction shifts LLMs from mere responders to being "thought partners," and sets a new standard for dynamic, dialogue-centric environments.

\vspace*{-2mm}
\section*{Limitations}
\vspace*{-2mm}

\fricabbr~addresses only the question of aligning language models to generate friction conditioned upon a task state where the terms of the task (though not the solution) are known, rather than toward a general response generation problem such as instruction following or summarization. Our goal is to train an LLM aligned toward the generation of interventions that prompt reflection and deliberation, and {\it not} a general dialogue agent/chatbot. In our results we have shown that common alignment methods of the kind used in dialogue or chatbot alignment are inferior to \fricabbr~in ability to generate these kinds of utterances. This does not necessarily mean that \fricabbr~is superior to other methods in aligning for human preference in other tasks, and as discussed in Sec.~\ref{ssec:datasets}, it is not clear that this would be a meaningful comparison because of the domain difference.

Although we motivate \fricabbr~based in part on theory of mind (Sec.~\ref{sec:intro}), we do not claim that it necessarily imbues an LLM with ToM and acknowledge that \fricabbr~aligned models could still inherit potential non-topical biases (say, from pretrains) in generating interventions as well as risks of overly confident or misaligned suggestions that could derail group dynamics. Instead, we use an "agentic" framework that trains a model to perform interventions for a desired effect \cite{russell2016artificial,krishnaswamy2022voxworld}. This is not to be confused with senses of LLM-agents such as "tool using" agents \cite{liu2024llava}. Within this framework, we render the frictive state $\phi$ in plain English text to make it amenable to LLM input, but as briefly mentioned in Sec.~\ref{sec:defs}, frictive states have a formal defintion based on evidence-based dynamic epistemic logic: a mental model $\mathcal{M} = \langle A,W,E,V \rangle$ consists of agents $A$, worlds $W$, evidence relation $E$ defining accessibility between worlds, and valuation function $V$. This allows the agent to assess alternatives and predict future developments from past events \cite{craik1943nature}. Thus, other formal structures to encode the frictive state could be explored (e.g., cf. \citet{obiso2025dynamic}) but were out of scope for this paper.

Finally, in terms of computational limitations, while we constructed \fricabbr~in a way that addresses data skewness and evaluated in a manner that sought to mitigate biases in the data generation distribution $\mu$, we cannot guarantee for certain that our results are bias-free. And, \fricabbr~still requires a reference model to be kept in memory, which leads to some additional compute requirements.

\vspace*{-2mm}
\section*{Acknowledgments}
\vspace*{-2mm}

This material is based in part upon work supported by Other Transaction award HR00112490377 from the U.S. Defense Advanced Research Projects Agency (DARPA) Friction for Accountability in Conversational Transactions (FACT) program, and by Other Transaction award 1AY2AX000062 from the U.S. Advanced Research Projects Agency for Health (ARPA-H) Platform Accelerating Rural Access to Distributed Integrated Medical Care (PARADIGM) program. The views and conclusions contained in this document are those of the authors and should not be interpreted as representing the official policies, either expressed or implied, of the U.S. Government. Thanks to Samuel Abeeb Abdullahi and Trevor Chartier for their annotation efforts. Thanks also to James Pustejovsky, Bruce Draper, Nathaniel Blanchard, and Sarath Sreedharan for the formative discussions on the foundational problems that led to this work. Portions of this work were performed on the Colorado State University Data Science Research Institute high-performance computer {\it Riviera}.

\bibliography{custom, nips_deception, nk_cites}
\bibliographystyle{acl_natbib}

\appendix


\section{Functional Definition and Samples of Naturally-Occurring Friction}
\label{appendices:friction_def}

The functional operative definition of friction in collaborative contexts that we used is given below. This definition was used when annotating the WTD for naturally-occurring frictive utterances, and used to construct the prompt for friction intervention generation, following work by \citet{oinas2009persuasive} and \citet{karadzhov2023delidata}.

\begin{tcolorbox}[width=\linewidth,title={\sc Functional Definition of Friction in Collaborative Tasks}]

Frictive interventions in this setting acts as indirect persuasion \cite{oinas2009persuasive} where participants are passively prompted to reevaluate their belief-states and incorrect assumptions or propositions, in light of incoming goal-specific evidence. We define productive or positive friction as interventions that act as indirect persuasion: agentic interventions that prompt participants to reevaluate their beliefs and assumptions about the task state, primarily but not solely, in light of incoming evidence (say, occurrences in the physical environment or a correct "declaration" previously occurring in the dialogue that any participant missed) that negates their preconceived notions about the state of the task. We call this indirect persuasion since we do not want our friction agent to directly offer hints about the task and thereby biasing task performance or negatively affecting the deliberation process that is proven to beneficial for successful task completion in reasoning-based, collaborative tasks \cite{karadzhov2023delidata}.

\end{tcolorbox}

Table~\ref{tab:friction-table} shows a sample friction annotation and training sample from the Weights Task Dataset, consisting of the dialogue history $x$, GPT-4o-identified frictive state $\phi$, rationale, and preferred and dispreferred friction interventions $f_w$ and $f_l$. Because the WTD is a multimodal dataset, the transcriptions we use are enriched using {\it dense paraphrasing}~\cite{tu2024dense}, a textual enrichment technique that uses the multimodal channels to decontextualize referents and in this case transform contextually-dependent phrasings such as demonstratives to explicit denotations of the content.  For example, under dense paraphrasing, "seems like {\it these} might be about the same" while the speaker in the video is pointing to the red and blue blocks becomes "seems like \textit{red block, blue block} might be about the same."  The dense paraphrased utterances are included as part of the publicly-available WTD~\cite{khebour2024text,khebour-etal-2024-common}.

\begin{table*}[ht]
\fontsize{8}{8}\selectfont
\centering
\begin{threeparttable}
\resizebox{\textwidth}{!}{
\begin{tabular}{p{0.15\textwidth}|p{0.85\textwidth}}
\toprule
\textbf{Field} & \textbf{Content} \\ \midrule
\textbf{Dialogue History ($x$)} & 
\begin{minipage}[t]{0.85\textwidth}
P1: i guess if red block red one's ten grams \\ 
P2: we got red ten \\
P1: seems like red block, blue block might be about the same \\
P3: i would agree yeah so blue block one's ten \\
P3: Alright let's see if we can find a twenty \\
P3: Too heavy so \\
P2: Way too heavy \\
P2: this is a sensitive scale \\
P2: Looks like about twenty \\
P1: that's looking pretty even \\
P3: Alright let's see if we can find a thirty \\
P1: so yellow block one is noticeably heavier than \\
P2: probably yellow block big sucker \\
P1: the purple ish one \\
P1: making sure that purple block didn't have the weight at the bottom \\
P2: it's just stuff written at the bottom that's a so red block, green block's a ten and a twenty right now right that's looking \\
P2: Well \\
P2: red block, blue block, green block, yellow block, purple block're increments of ten i would say that's probably \\
P1: Yeah I think \\
P2: cause purple block's also a twenty let's double check that purple block's not also a twenty \\
P1: yeah it looks a little \\
P2: cause it um just purple block one there \\
P3: is blue block one a twenty \\
P2: ok so purple block's more than twenty but it almost seems like the thirty takes it past but \\
P2: it's so sensitive \\
P2: if red block, blue block, green block, yellow block, purple block're only in increments of ten purple block has to be
\end{minipage} \\
\midrule


\textbf{Frictive state ($\phi$)} & P2 initially identified the red and blue blocks as both 10 grams and has speculated about the green at 20 grams, but is uncertain about the actual weights of the yellow and purple blocks. \\ \midrule
\textbf{Rationale} & P2 suggests that the red, green, and yellow blocks are all in increments of ten. Encourage a double-check. \\ \midrule
\textbf{Preferred Friction ($f_w$)} & Since the purple block seems heavier and we’re unsure about its exact weight, should we double-check the increments of ten assumption? Maybe the purple block doesn’t fit this pattern. \\ \midrule
\textbf{Dispreferred Friction ($f_l$)} & You know, the purple block being heavier might actually mean the blocks aren’t increasing consistently at all. What if the increments are random, like 10, 15, or 25 grams, and we’re forcing a pattern that isn’t there? \\ \bottomrule

\end{tabular}}
\end{threeparttable}
\vspace{-2ex}
\caption{\label{tab:friction-table}
A transcribed, sparse collaborative dialogue from the Weights Task Dataset~\cite{khebour2024text} with frictive states and friction interventions. Preferred and dispreferred friction interventions are shown at the bottom. Positive friction interventions prompt participants to reevaluate their assumptions with frictive states (evolution of beliefs and rationales) providing indirect hints and directions. Here, P2's uncertainty about the green block and assumption of weight increment by 10g is addressed by the positive friction. In contrast, the dispreferred intervention introduces randomness, instigating the group to abandon structured reasoning.}
\vspace{-1em}
\end{table*}

\section{Frictive-state conditioning and RLHF}
\label{appendices:frictive_rlhf}

In its simplest formulation within Chain-of-Thought (CoT) settings~\citep{wei2023chainofthought}, the friction agent is modeled as a policy distribution \( \pi_f \) that sequentially generates frictive states, sampling \( \phi_i \sim \pi_f(\cdot \mid x, \phi_1, \dots, \phi_{i-1}) \), and ultimately producing the final friction intervention \( f \sim \pi_f(\cdot \mid x, \phi_1, \dots, \phi_n) \). Here, \( x \) represents the dialogue history, \( f \) denotes the intervention, and \( \phi \) consists of sequentially sampled frictive state tokens, analogous to "thoughts" in standard CoT-based reasoning frameworks~\cite{yao2023tree}. Unlike standard CoT-based alignment, which relies on self-rewarding strategies, we frame friction agent alignment within preference-based RL (PbRL;~\citet{wirth2017survey}). Prior work~\cite{zhang2024chain} shows CoT frameworks benefit significantly from contrastive signals in preference learning.

In this setting, we define the human preference probability \( \mathcal{P}(f \succ \phi) \) as the probability that an expert annotator would prefer  \( f \) over maintaining the frictive state \( \phi \), given prior dialogue history, $x$. The key insight is that to retrieve the optimal policy \( \pi_f^* \), we can leverage established methods from RLHF and PbRL by formulating the problem as a KL-divergence constrained minimum relative entropy optimization~\citep{ziebart2008maximum}, a well-known approach with a closed-form solution~\cite{peng2019advantageweightedregressionsimplescalable}. 

\begin{align}
\small
J^*_{\text{RLHF}}(\pi_f) = \notag \\
&\max_{\pi_f}\mathbb{E}_{f\sim\pi_f}
\left[\mathcal{P}(f\succ\phi \mid x)\right] - \notag \\
&\beta D_{\text{KL}}(\pi_f \parallel \pi_{\text{ref}}).
\label{eq:RLHF_phi}
\end{align}

This formulation (Eq.~\ref{eq:RLHF_phi})---where $J^*_{\text{RLHF}}$  enforces a KL-based "soft"-constraint on the parametric form of \( \pi_f^* \) wrt the reference policy $\pi_\text{ref}$---provides crucial tradeoffs between training stability and balancing exploration vs exploitation. Specifically, $J^*_{\text{RLHF}}$ ensures that \( \pi_f^* \) retrieves the best possible preference probabilities for its generated interventions $f$, as assigned by \( \mathcal{P}(f \succ \phi) \), whether over the distribution of preferences encoded in an offline dataset~\citep{rafailov2024direct} or from online sampling $\sim$ \( \pi_f \) during training~\cite{schulman2017proximal} while being distributionally close to an "already-good" imitator, the Supervised-Finetuned (SFT) reference model~\cite{hussein2017imitation}. \textbf{Notice that unlike standard RLHF, we formulate  $J^*_{\text{RLHF}}$ such that $\pi_f^*$ takes the form $\pi^*_f (\cdot \mid \phi, x)$ and is explicitly conditioned on the frictive state $\phi$, apart from $x$.} This is intentional since we hypothesize that an ideal friction agent does \textit{not} intervene arbitrarily, causing distraction in collaborative tasks and is conditioned to resolve the lack of common ground thereof between human collaborators, \textit{by definition}---as observed in $\phi$. While prior work~\citep{choi2024robust, zhang2024chain} explores preference alignment in LLMs in such CoT-conditioned scenarios, we provide a more principled approach to proving the existence and the uniqueness of $\pi_f^*$ that $J^*_{\text{RLHF}}$ seeks to retrieve. Mathematically, 
\begin{equation}
   \pi_f^* = \frac{\pi_{\text{ref}}\exp(\beta^{-1}\mathcal{P}(f\succ\phi|x))}{Z^*(\phi,x)},
   \label{eq:optimal_frictive_agent_policy}
\end{equation}
where $Z^* = \sum_{f'}\pi_{\text{ref}}\exp(\beta^{-1}\mathcal{P}(f'\succ\phi|x))$ is the partition function which is fixed and does not depend on $f$ and can be safely ignored in the optimization of $J^*_{\text{RLHF}}$~\citep{rafailov2024direct}. See \Cref{appendices:optimal_friction} and \Cref{eq:optimal_frictive_agent_policy} for the full-proof and optimal policy form respectively.

\subsection{Existence and uniqueness of the optimal friction intervention policy}
\label{appendices:optimal_friction}
In order to derive an empirical offline (supervised) preference learning loss from the complicated two-staged \fricabbr-alignment objective defined in \Cref{eq:two_stage_main_objective_main_paper}, we use a divide and conquer approach---\textit{our core insight here is to express the preference of interventions conditioned on the frictive states in terms of two mutually supportive "twin" policies.} As such, we first derive the inner maximization loop of Eq.~\ref{eq:two_stage_main_objective_main_paper} to get an analytical expression of the optimal frictive intervention policy, $\pi_f^*$ as shown in the proof for Eq.~\ref{eq:optimal_frictive_agent_policy}. However, we observe that $\pi_f^*$ in its analytical form is not fully expressive since it does \textit{not} contain the optimal frictive-state policy $\pi_\phi^*$ term. Therefore, we propose a novel method to derive $\pi_\phi^*$ using a Lagrangian formulation. We show the detailed derivation for this part in \Cref{sec:optimal_frictive_state_derivation} including supporting results from \Cref{lemma:threshold_theorem_sequential_choice} and \Cref{lemma:policy_ratio_equivalence}.

This above result is one of our \textit{main contributions} since it lets us express the preference for any intervention $f_1$ over $f_2$ analytically in terms of \textbf{both} the optimal friction intervention policy ($\pi_f^*(\cdot \mid \phi, x)$) and the optimal frictive-state policy ($ \pi_\phi^*(\cdot|x)$). Finally, this core result is used to propose a straightforward supervised ($\ell_2$) objective—similar in spirit to IPO~\cite{azar2024general}—that empirically regresses the predicted preference expression derived from $\pi_f^*(\cdot \mid \phi, x)$ and $\pi_\phi^*(\cdot|x)$ to the observed relative preferences $p(f_1 \succ f_2 \mid x)$ (relative to $\phi$), assuming access to a large-enough preference-annotated dataset of frictive interventions. Notably, this objective is optimized by a \textit{single} parametrized policy that leverages the inherent expressivity of LLMs with billions of parameters. 

In particular, this \texttt{FAAF} objective formulation avoids some of the policy degeneracy issues that popular supervised "offline" alignment algorithms like Direct Preference Optimization (DPO)~\citep{rafailov2024direct} face due to its unbounded rewards. Additionally, unlike~\citet{fisch2024robustpreferenceoptimizationreward}, our regression objective works directly on preference labels and does not require an external reward model in avoiding such degeneracies. Finally, we also prove that \texttt{FAAF}-trained policies are unique solutions in the policy space in \Cref{theorem:faaf_uniqueness}.

For completeness, we first prove the existence of the optimal friction/frictive intervention policy that solves the inner maximization of our two-part minimax objective. The structural solution to this objective is well-studied in the RL/control-theory literature including popular frameworks in preference alignment in LLMs~\cite{ziebart2008maximum, peng2019advantageweightedregressionsimplescalable,rafailov2024direct, azar2024general} as well as Chain-of-Thought (CoT)-based preference alignment frameworks~\cite{choi2024robust}. We show how it specifically applies to our unique parametrization. Our proof follows similar logic as \citet{azar2024general}. Let us recall two-part minimax objective (Eq.~\ref{eq:two_stage_main_objective_main_paper}) for clarity here:

\begin{align}
\small
    J^*_{\text{\fricabbr}} &= \min_{\pi_\phi} \max_{\pi_f} 
    \mathbb{E}_{
    \substack{x \sim \rho \\ \phi \sim \pi_\phi(\cdot \mid x) \\ f \sim \pi_f(\cdot \mid \phi, x)}
    } \Bigg[
    \mathcal{P}(f \succ \phi \mid x)  \notag \\
    &\quad - \beta D_{\text{KL}}(\pi_f \parallel \pi_{\text{ref}} \mid \phi, x)  \notag \\
    &\quad + \beta D_{\text{KL}}(\pi_\phi \parallel \pi_{\text{ref}} \mid x) 
    \Bigg].
\label{eq:two_stage_main_objective}
\end{align}

For fixed $\pi_\phi$, the inner maximization reduces to our regularized objective:
\begin{align}
\mathcal{L}_\beta(\pi_f) &= \mathbb{E}_{f\sim\pi_f}[p(f\succ\phi|x)] - \notag \\
&\quad\beta D_{\text{KL}}(\pi_f \parallel \pi_{\text{ref}}|\phi,x), \notag \\
&= \sum_{f}\pi_f(f|\phi,x)p(f\succ\phi|x) - \notag \\ 
&\quad\beta D_{\text{KL}}(\pi_f \parallel \pi_{\text{ref}}|\phi,x),
\end{align}
where $f \in \mathcal{F}$ is from a finite friction token alphabet $\mathcal{F}$, $p(f\succ\phi|x)$ maps elements of $\mathcal{F}$ to the utility of generating a frictive intervention $f$ defined as the preference of $f$ over the frictive-state $\phi$, given context $x$, $\beta\in\mathbb{R}^*_+$ is a strictly positive real number, and $\pi_f, \pi_{\text{ref}}$ are conditional probability distributions. In particular, notice that the conditional probability distribution $\pi_f(f|\phi,x)$ can be identified as a positive real function satisfying:

\begin{equation}
   \sum_{f}\pi_f(f|\phi,x) = 1.
\end{equation}

Now, if we define the optimal friction intervention policy $\pi_f^*$ as:
\begin{equation}
   \pi_f^*(f|\phi,x) = \frac{\pi_{\text{ref}}(f|\phi,x)\exp(\beta^{-1}p(f\succ\phi|x))}{Z^*(\phi,x)},
   \label{eq:optimal_frictive_agent_policy}
\end{equation}
recalling Eq.~\ref{eq:optimal_policy_dpo_kto_ipo}, where $Z^*(\phi,x) = \sum_{f'}\pi_{\text{ref}}(f'|\phi,x)\exp(\beta^{-1}p(f'\succ\phi|x))$,
then, under the previous definitions, we have:
\begin{equation}
   \pi_f^* = \argmax_{\pi_f}\mathcal{L}_\beta(\pi_f)
\end{equation}

\vfill

\begin{figure*}[htbp]
\begin{tcolorbox}[width=\textwidth]
\begin{proof}
\begin{align*}
\frac{\mathcal{L}_\beta(\pi_f)}{\beta} &= \sum_{f \in \mathcal{F}}\pi_f(f|\phi,x)\frac{p(f\succ\phi|x)}{\beta} - D_{\text{KL}}(\pi_f \parallel \pi_{\text{ref}}|\phi,x),
\\
&= \sum_{f \in \mathcal{F}}\pi_f(f|\phi,x)\frac{p(f\succ\phi|x)}{\beta} - \sum_{f \in \mathcal{F}}\pi_f(f|\phi,x)\log\big(\frac{\pi_f(f|\phi,x)}{\pi_{\text{ref}}(f|\phi,x)}\big),
\\
&= \sum_{f \in \mathcal{F}}\pi_f(f|\phi,x)\big(\frac{p(f\succ\phi|x)}{\beta} - \log\big(\frac{\pi_f(f|\phi,x)}{\pi_{\text{ref}}(f|\phi,x)}\big)\big),
\\
&= \sum_{f \in \mathcal{F}}\pi_f(f|\phi,x)\big(\log(\exp(\beta^{-1}p(f\succ\phi|x))) - \log\big(\frac{\pi_f(f|\phi,x)}{\pi_{\text{ref}}(f|\phi,x)}\big)\big),
\\
&= \sum_{f \in \mathcal{F}}\pi_f(f|\phi,x)\log\big(\exp(\beta^{-1}p(f\succ\phi|x))\frac{\pi_{\text{ref}}(f|\phi,x)}{\pi_f(f|\phi,x)}\big),
\\
&= \sum_{f \in \mathcal{F}}\pi_f(f|\phi,x)\log\big(\frac{\pi_{\text{ref}}(f|\phi,x)\exp(\beta^{-1}p(f\succ\phi|x))}{\pi_f(f|\phi,x)}\big),
\\
&= \sum_{f \in \mathcal{F}}\pi_f(f|\phi,x)\log\big(\frac{\pi_{\text{ref}}(f|\phi,x)\exp(\beta^{-1}p(f\succ\phi|x))}{\pi_f(f|\phi,x)}\frac{Z^*(\phi,x)}{Z^*(\phi,x)}\big),
\\
&= \sum_{f \in \mathcal{F}}\pi_f(f|\phi,x)\log\big(\frac{\pi_{\text{ref}}(f|\phi,x)\exp(\beta^{-1}p(f\succ\phi|x))}{Z^*(\phi,x)}\frac{Z^*(\phi,x)}{\pi_f(f|\phi,x)}\big),
\\
&= \sum_{f \in \mathcal{F}}\pi_f(f|\phi,x)\log\big(\frac{\pi_f^*(f|\phi,x)}{\pi_f(f|\phi,x)}\big) + \sum_{f \in \mathcal{F}}\pi_f(f|\phi,x)\log Z^*(\phi,x),
\\
&= \sum_{f \in \mathcal{F}}\pi_f(f|\phi,x)\log\big(\frac{\pi_f^*(f|\phi,x)}{\pi_f(f|\phi,x)}\big) + \sum_{f \in \mathcal{F}}\pi_f(f|\phi,x)\log Z^*(\phi,x),
\\
&= -D_{\text{KL}}(\pi_f \parallel \pi_f^*) + \log Z^*(\phi,x), \quad \text{(using normalization } \sum_{f \in \mathcal{F}}\pi_f(f|\phi,x) = 1\text{)}
\end{align*}

By definition of the KL divergence, we know that $\pi_f^* = \argmax_{\pi_f}\big[-D_{\text{KL}}(\pi_f \parallel \pi_f^*)\big]$ and as:
\begin{equation*}
-D_{\text{KL}}(\pi_f \parallel \pi_f^*) = \frac{\mathcal{L}_\beta(\pi_f)}{\beta} - \log Z^*(\phi,x)
\end{equation*}
where $\log Z^*(\phi,x)$ is the partition function~\cite{peng2019advantageweightedregressionsimplescalable,rafailov2024direct} and has no dependency on $\pi_f$ and $\beta\in\mathbb{R}^*_+$ is a strictly positive real number. Therefore, the argmax of $-D_{\text{KL}}(\pi_f \parallel \pi_f^*)$ coincides with that of $\mathcal{L}_\beta(\pi_f)$, concluding the proof.
\end{proof}
\end{tcolorbox}
\end{figure*}

\clearpage

\begin{lemma}[Value of Inner Maximization]
\label{lemma:value_of_inner_max}
When substituting the optimal friction intervention policy $\pi_f^*$, as derived in Eq. \ref{eq:optimal_frictive_agent_policy}, into Eq.~\ref{eq:two_stage_main_objective}, the objective in Eq.~\ref{eq:two_stage_main_objective} reduces to:
\begin{align}
    J^*_\fricabbr = & \min_{\pi_\phi} \mathbb{E}_{x\sim\rho, \phi\sim\pi_\phi(\cdot|x)}[\beta\log(Z^*(\phi,x)) + \notag \\
    & \beta D_{KL}(\pi_\phi||\pi_{ref}|x)]
\end{align}
\end{lemma}

\begin{tcolorbox}
\begin{proof}
Substituting $\pi_f^*$ into the KL divergence term:
\begin{align}
    &D_{KL}(\pi_f^*||\pi_{ref}|\phi,x) \notag \\
    &= \mathbb{E}_{f\sim\pi_f^*}\bigg[\log(\pi_f^*(f|\phi,x)) - \notag \\
    &\quad\log(\pi_\text{ref}(f|\phi,x))\bigg] \notag \\
    &= \mathbb{E}_{f\sim\pi_f^*}\bigg[\frac{p(f\succ\phi|x)}{\beta} - \notag \\
    &\quad\log(Z^*(\phi,x))\bigg]
\end{align}

The original objective becomes:
\begin{align}
    &p(f \succ \phi|x) - \beta\mathbb{E}_{f\sim\pi_f^*}\bigg[\frac{p(f\succ\phi|x)}{\beta} - \notag \\
    &\quad\log(Z^*(\phi,x))\bigg] \notag \\
    &= \beta\log(Z^*(\phi,x))
\end{align}

The result follows by substituting this value back into the full objective.
\end{proof}
\end{tcolorbox}

\section{Derivation of Optimal Frictive State Policy}
\label{sec:optimal_frictive_state_derivation}
We begin with the reduced objective function after solving the inner maximization as shown in \Cref{lemma:value_of_inner_max}.

\begin{align}
    J^*_\fricabbr = &\min_{\pi_\phi} \mathbb{E}_{x\sim\rho, \phi\sim\pi_\phi(\cdot|x)}\big[\beta\log(Z^*(\phi,x)) \notag \\
    &\quad + \beta D_{KL}(\pi_\phi||\pi_\text{ref}|x)\big]
    \label{eq:j_star_after_inner_max}
\end{align}

The Kullback-Leibler divergence term expands as follows:
\begin{equation}
    D_{KL}(\pi_\phi||\pi_\text{ref}|x) = \mathbb{E}_{\phi\sim\pi_\phi}\left[\log\frac{\pi_\phi(\phi|x)}{\pi_\text{ref}(\phi|x)}\right]
\end{equation}

Substituting this back into our objective:
\begin{align}
    J^*_\fricabbr = &\min_{\pi_\phi} \mathbb{E}_{\phi\sim\pi_\phi}\Big[\beta\log(Z^*(\phi,x)) 
    + \notag \\
    &\quad\beta\log(\pi_\phi(\phi|x)) - \beta\log(\pi_\text{ref}(\phi|x))\Big]
\end{align}

Since \( \pi_\phi \) must be a valid probability distribution satisfying \( \sum_{\phi} \pi_\phi(\phi|x) = 1 \), we introduce a Lagrange multiplier \( \lambda \) and define the corresponding Lagrangian function to derive the optimality conditions:




\begin{align}
    L(\pi_\phi) &= \mathbb{E}_{\phi\sim\pi_\phi} \Big[\beta\log(Z^*(\phi,x)) + \notag \\
    &\quad\beta\log(\pi_\phi(\phi|x)) - \beta\log(\pi_\text{ref}(\phi|x)) \Big] + \notag \\
    &\quad \lambda \left( 1 - \sum_{\phi} \pi_\phi(\phi|x) \right) \notag \\
    &= \sum_{\phi} \pi_\phi(\phi|x) \bigg[\beta\log(Z^*(\phi,x)) + \notag \\
    &\quad\beta\log(\pi_\phi(\phi|x)) - \beta\log(\pi_{ref}(\phi|x)) \bigg] \notag \\
    &\quad + \lambda \left( 1 - \sum_{\phi} \pi_\phi(\phi|x) \right).
\end{align}

Now, to find the optimal policy $\pi_\phi^*(\phi|x)$, we take the derivative of the Lagrangian with respect to $\pi_\phi(\phi|x)$ and equate it to zero:

\begin{align}
    \frac{\delta L}{\delta \pi_\phi(\phi|x)} &= \beta \log(Z^*(\phi,x)) + \notag \\
    &\quad\beta \frac{\delta}{\delta \pi_\phi} \bigg[\pi_\phi(\phi|x) \log(\pi_\phi(\phi|x))\bigg] \notag \\
    &\quad- \beta\log(\pi_{ref}(\phi|x)) - \lambda = 0.
\end{align}



From the standard functional derivative of entropy $\displaystyle \frac{\delta}{\delta \pi_\phi} \bigg[\pi_\phi(\phi|x) \log(\pi_\phi(\phi|x))\bigg] = 1 + \log(\pi_\phi(\phi|x))$, we obtain: 

\begin{align}
    &\beta \log(Z^*(\phi,x)) + \beta (1 + \log(\pi_\phi(\phi|x))) - \notag \\
    &\quad\beta \log(\pi_{ref}(\phi|x)) + \lambda = 0.
\end{align}

Rearranging the terms:

\begin{align}
    \log(\pi_\phi(\phi|x)) &= \log(\pi_{ref}(\phi|x)) - \notag \\
    &\quad\log(Z^*(\phi,x)) - \frac{\lambda}{\beta} - 1.
\end{align}







Taking the exponential on both sides:
\begin{equation}
    \pi_\phi(\phi|x) = e^{-1 - \frac{\lambda}{\beta}} \pi_{\text{ref}}(\phi|x) e^{-\log Z^*(\phi,x)}.
\end{equation}

To ensure \( \pi_\phi(\phi|x) \) is a valid probability distribution, we define the normalization constant:
\begin{equation}
    Z(x) = \sum_{\phi} \pi_{\text{ref}}(\phi|x) e^{-\beta\log Z^*(\phi,x)}.
\end{equation}

Thus, the optimal frictive-state policy is:
\begin{equation}
    \pi_\phi^*(\phi|x) = \frac{\pi_{\text{ref}}(\phi|x) e^{-\beta\log Z^*(\phi,x)}}{Z(x)}.
    \label{eq:optimal_frictive_state_policy}
\end{equation}

Notice that without losing any generality, we can parametrize the above optimal frictive-state policy with any outcome $f$ consistent with the structure in Eq.~\ref{eq:optimal_frictive_state_policy} as follows:

\begin{equation}
    \pi_\phi^*(f|x) = \frac{\pi_\text{ref}(f|x)}{Z(x)} e^{-\beta\log(Z^*(f,x))}.
     \label{eq:optimal_frictive_agent_policy_unconditioned}
\end{equation}

Note that although this formulation of the optimal  frictive-state policy ($ \pi_\phi^*(\phi|x)$) is an analytical solution to $ J^*$ from Eq.~\ref{eq:j_star_after_inner_max}, we still need to represent $ \pi_\phi^*(\phi|x)$ in terms of the optimal friction intervention policy, $\pi_f^*(\cdot \mid \phi, x)$ proposed in Eq.~\ref{eq:optimal_frictive_agent_policy} and the preference probabilities $p(f \succ \phi|x)$, the preference probability of the friction $f$ over the frictive-state $\phi$, given context $x$. This is crucial to derive the empirical \fricabbr~optimization objective that can be used for standard offline learning. Therefore, to represent the $p(f \succ \phi|x)$  in terms of $\pi_f^*(\cdot\mid \phi, x)$, we take the logarithm of Eq.~\ref{eq:optimal_frictive_agent_policy} on both sides and some algebra, we obtain:


\begin{align*}
&\log(\pi_f^*(f|\phi,x)) = \frac{p(f\succ\phi|x)}{\beta} + \notag \\
&\quad\log(\pi_\text{ref}(f|\phi,x)) - \log(Z^*(\phi,x)).
\end{align*}

Multiplying both sides by $\beta$ and rearranging terms, we obtain: 

\begin{align}
&p(f \succ \phi|x) = \beta[\log(\pi_f^*(f|\phi,x)) - \notag \\
&\quad\log(\pi_\text{ref}(f|\phi,x)) + \log(Z^*(\phi,x))].
\label{eq:preference_prob_friction_over_frictive_state}
\end{align}

Similar to \citet{munos2023nash}, \citet{azar2024general}, and \citet{choi2024robust}, we can apply the identity that $p(\phi \succ \phi|x) = \frac{1}{2}$ and substitute $f = \phi$ into the previous equation and derive:

\begin{align}
&\frac{1}{2} = \beta[\log(\pi_f^*(\phi|\phi,x)) - \notag \\
&\quad\log(\pi_\text{ref}(\phi|\phi,x)) + \log(Z^*(\phi,x))].
\end{align}

Solving for $\log(Z^*(\phi,x))$ gives:

\begin{align}
&\log(Z^*(\phi,x)) = \frac{1}{2\beta} - \notag \\
&\quad[\log(\pi_f^*(\phi|\phi,x)) - \log(\pi_\text{ref}(\phi|\phi,x))].
\end{align}

Substituting this back into Eq.~\ref{eq:preference_prob_friction_over_frictive_state} results in:
\begin{align}
p(f \succ \phi|x) 
&= \beta[\log(\pi_f^*(f|\phi,x)) - \notag \\
&\quad\log(\pi_\text{ref}(f|\phi,x)) \notag \\
&\quad + \frac{1}{2\beta} - (\log(\pi_f^*(\phi|\phi,x)) - \notag \\
&\quad\log(\pi_\text{ref}(\phi|\phi,x)))] \notag \\
&= \beta \log \left( \frac{\pi_f^*(f|\phi,x)}{\pi_\text{ref}(f|\phi,x)} \right) + \frac{1}{2} \notag \\
&\quad- \beta \log \left( \frac{\pi_f^*(\phi|\phi,x)}{\pi_\text{ref}(\phi|\phi,x)} \right).
\label{eq:identity_optimal_conditioned_ratios}
\end{align}

The $\log \left( \frac{\pi_f^*(\phi|\phi,x)}{\pi_\text{ref}(\phi|\phi,x)}\right)$ term in the above step is a self-referential term signifying the friction intervention policy's ($\pi_f^*(\cdot \mid \phi, x)$) estimate of the frictive state given $\phi$. However, this term does \textit{not} provide much information on the regularized preference in terms of the frictive state policy. Recall that our outer minimization objective operates over $ \pi_\phi(\cdot|x)$. Fortunately, we can use our results from \Cref{lemma:threshold_theorem_sequential_choice} and \Cref{lemma:policy_ratio_equivalence} to express Eq.~\ref{eq:identity_optimal_conditioned_ratios} in terms of  the optimal frictive state policy $\pi_\phi^*(\cdot|x)$. Therefore, from \Cref{lemma:policy_ratio_equivalence} we can express $\pi_f^*$ and $\pi_{\text{ref}}$ as follows: 

For the optimal policy $\pi_f^*$:
\begin{align}
\log(\pi_f^*(\phi|\phi,x)) &= \log(\pi_\phi^*(\phi|x)) - \notag \\
&\quad\log(\pi_\phi^*(f|x))
\end{align}

For the reference policy $\pi_{\text{ref}}$:
\begin{align}
\log(\pi_{\text{ref}}(\phi|\phi,x)) &= \log(\pi_{\text{ref}}(\phi|x)) - \notag \\
&\quad\log(\pi_{\text{ref}}(f|x))
\end{align}

Now, substituting these expressions into \Cref{eq:identity_optimal_conditioned_ratios}, we get:
\begin{align}
p(f \succ \phi|x) 
&= \beta \Big[\log(\pi_f^*(f|\phi,x)) - \notag \\
&\quad\log(\pi_{\text{ref}}(f|\phi,x)) + \notag \\
&\quad\frac{1}{2\beta} - \log(\pi_\phi^*(\phi|x)) + \notag \\
&\quad\log(\pi_\phi^*(f|x)) - \log(\pi_{\text{ref}}(\phi|x)) + \notag \\
&\quad\log(\pi_{\text{ref}}(f|x)) \Big] \notag \\
&= \beta \Big[\log(\pi_f^*(f|\phi,x)) - \notag \\
&\quad\log(\pi_{\text{ref}}(f|\phi,x)) + \notag \\
&\quad\frac{1}{2\beta} - \Big(\log(\pi_\phi^*(\phi|x)) - \notag \\
&\quad\log(\pi_{\text{ref}}(\phi|x)) - \notag \\
&\quad(\log(\pi_\phi^*(f|x)) - \log(\pi_{\text{ref}}(f|x)))\Big)\Big] \notag \\
&= \beta\Bigg[\log\left(\frac{\pi_f^*(f|\phi,x)}{\pi_{\text{ref}}(f|\phi,x)}\right) + \frac{1}{2\beta} + \notag \\
&\quad\log\left(\frac{\pi_\phi^*(f|x)}{\pi_{\text{ref}}(f|x)}\right) - \notag \\
&\quad\log\left(\frac{\pi_\phi^*(\phi|x)}{\pi_{\text{ref}}(\phi|x)}\right)\Bigg].
\end{align}

Now, replacing $f$ by $f_1$ in $p(f \succ \phi|x)$:
\begin{align}
p(f_1 \succ \phi|x) &= \beta\Bigg[\log\left(\frac{\pi_f^*(f_1|\phi,x)}{\pi_{\text{ref}}(f_1|\phi,x)}\right) + \notag \\
&\quad\frac{1}{2\beta} + \log\left(\frac{\pi_\phi^*(f_1|x)}{\pi_{\text{ref}}(f_1|x)}\right) - \notag \\
&\quad\log\left(\frac{\pi_\phi^*(\phi|x)}{\pi_{\text{ref}}(\phi|x)}\right)\Bigg]
\end{align}

Similarly, expressing $f_2$ in $p(f \succ \phi|x) $, we obtain: 
\begin{align}
p(f_2 \succ \phi|x) &= \beta\Bigg[\log\left(\frac{\pi_f^*(f_2|\phi,x)}{\pi_{\text{ref}}(f_2|\phi,x)}\right) + \notag \\
&\quad\frac{1}{2\beta} + \log\left(\frac{\pi_\phi^*(f_2|x)}{\pi_{\text{ref}}(f_2|x)}\right) - \notag \\
&\quad\log\left(\frac{\pi_\phi^*(\phi|x)}{\pi_{\text{ref}}(\phi|x)}\right)\Bigg]
\end{align}

Now, expressing $p(f_1 \succ \phi|x) - p(f_2 \succ \phi|x)$, the relative preference probability of $f_1$ over $f_2$ given $\phi$ and $x$,  we observe that $\log\left(\frac{\pi_\phi^*(\phi|x)}{\pi_{\text{ref}}(\phi|x)}\right)$ terms cancel out and we derive:

\begin{align}
&p(f_1 \succ \phi|x) - p(f_2 \succ \phi|x) = \notag \\
&\quad\beta\Bigg[\log\left(\frac{\pi_f^*(f_1|\phi,x)}{\pi_{\text{ref}}(f_1|\phi,x)}\right) + \frac{1}{2\beta} + \notag \\
&\quad\log\left(\frac{\pi_\phi^*(f_1|x)}{\pi_{\text{ref}}(f_1|x)}\right) - \log\left(\frac{\pi_\phi^*(\phi|x)}{\pi_{\text{ref}}(\phi|x)}\right)\Bigg] \notag \\
&- \beta\Bigg[\log\left(\frac{\pi_f^*(f_2|\phi,x)}{\pi_{\text{ref}}(f_2|\phi,x)}\right) + \frac{1}{2\beta} + \notag \\
&\quad\log\left(\frac{\pi_\phi^*(f_2|x)}{\pi_{\text{ref}}(f_2|x)}\right) - \log\left(\frac{\pi_\phi^*(\phi|x)}{\pi_{\text{ref}}(\phi|x)}\right)\Bigg] \notag \\
&= \beta\Bigg[\log\left(\frac{\pi_f^*(f_1|\phi,x)}{\pi_{\text{ref}}(f_1|\phi,x)}\right) - \log\left(\frac{\pi_f^*(f_2|\phi,x)}{\pi_{\text{ref}}(f_2|\phi,x)}\right) \notag \\
&\quad + \log\left(\frac{\pi_\phi^*(f_1|x)}{\pi_{\text{ref}}(f_1|x)}\right) - \log\left(\frac{\pi_\phi^*(f_2|x)}{\pi_{\text{ref}}(f_2|x)}\right)\Bigg]
\end{align}

This above result is one of our \textit{core contributions} since it lets us express the relative preference of any friction intervention $f_1$ over $f_2$ given a frictive state ($\phi$) analytically in terms of \textbf{both} the optimal friction intervention policy (($\pi_f^*(\cdot \mid \phi, x)$)) and the optimal frictive state policy ($ \pi_\phi^*(\cdot|x)$):

\begin{align}
&p(f_1 \succ \phi|x) - p(f_2 \succ \phi|x) = \notag \\
&\quad\beta\Bigg[\log\left(\frac{\pi_f^*(f_1|\phi,x)}{\pi_{\text{ref}}(f_1|\phi,x)}\right) - \log\left(\frac{\pi_f^*(f_2|\phi,x)}{\pi_{\text{ref}}(f_2|\phi,x)}\right) \notag \\
&\quad + \log\left(\frac{\pi_\phi^*(f_1|x)}{\pi_{\text{ref}}(f_1|x)}\right) - \log\left(\frac{\pi_\phi^*(f_2|x)}{\pi_{\text{ref}}(f_2|x)}\right)\Bigg]
\end{align}

Following a standard approach for empirical estimation of the LHS~\citep{azar2024general} in the above equation, one can learn \textit{both} the optimal friction intervention policy $\pi_f^*$ and the frictive-state policy  $\pi_\phi^*$ using a trainable policy $\pi_\theta$, parametrized with $\theta$. The core insight here is to exploit the expressive nature of LLMs' hidden representations with billions of parameters to learn a \textit{single} optimal policy. A reasonable choice here is to train $\pi_\theta$ through an $\ell_2$ loss~\citep{fisch2024robustpreferenceoptimizationreward} that enforces the relative preference ordering between any pair of friction interventions $(f_1, f_2)$ with implicit reward estimates from the RHS of Eq. 34. However, unlike~\citep{fisch2024robustpreferenceoptimizationreward}, our approach in enforcing this constraint does not require access to an external reward model or an "oracle" for point-wise reward estimates, assuming we have access to labeled preference feedback in samples. Additionally, the $\ell_2$ formulation avoids placing a unbounded logit or a inverse sigmoid function over the preference since this has been shown to cause non-trivial policy degeneracy issues in learning algorithms like DPO~\citep{azar2024general}. Applying this $\ell_2$ loss, we derive:

\begin{align}
&\mathcal{L}_{\pi_\theta} = \mathbb{E}_{\substack{x \sim \rho \\ \phi \sim \pi_\theta(\cdot|x) \\ f_1,f_2 \sim \pi_\theta(\cdot|\phi,x)}} \notag \\
&\quad\Bigg(p(f_1 \succ \phi|x) - p(f_2 \succ \phi|x) - \notag \\
&\quad\beta\Bigg[\log\left(\frac{\pi_\theta(f_1|\phi,x)}{\pi_{\text{ref}}(f_1|\phi,x)}\right) - \log\left(\frac{\pi_\theta(f_2|\phi,x)}{\pi_{\text{ref}}(f_2|\phi,x)}\right) \notag \\
&\quad + \log\left(\frac{\pi_\theta(f_1|x)}{\pi_{\text{ref}}(f_1|x)}\right) - \log\left(\frac{\pi_\theta(f_2|x)}{\pi_{\text{ref}}(f_2|x)}\right)\Bigg]\Bigg)^2
\end{align}

Since the friction dataset $\mathcal{D}_\mu$ sampled from $\mu$ contains preference-annotated pairs $(f_w, f_l)$ given $\phi$ and $x$, the preference probabilities can be expressed using indicator functions as $p(f_w \succ f_l|x) = \mathbb{E}[\mathbf{1}(f_w \succ f_l|x)] = 1$ and $p(f_l \succ f_w|x) = \mathbb{E}[\mathbf{1}(f_l \succ f_w|x)] = 0$. Furthermore, the difference $p(f_w \succ f_l|x) - p(f_l \succ f_w|x) = 1 - 0 = 1$ aligns with the formulation $p(f_1 \succ \phi|x) - p(f_2 \succ \phi|x)$ when $f_1 = f_w$ and $f_2 = f_l$. Therefore, we can write our final \fricabbr-alignment empirical objective function ($\hat{\mathcal{L}}$) as follows:

\begin{align}
&\hat{\mathcal{L}} (\pi_\theta) = \mathbb{E}_{(x,\phi,f_w,f_l) \sim \mathcal{D}_\mu} \notag \\
&\quad\Bigg(1 - \beta\Bigg[\log\left(\frac{\pi_\theta(f_w|\phi,x)}{\pi_{\text{ref}}(f_w|\phi,x)}\right) - \notag \\
&\quad\log\left(\frac{\pi_\theta(f_l|\phi,x)}{\pi_{\text{ref}}(f_l|\phi,x)}\right) + \notag \\
&\quad\log\left(\frac{\pi_\theta(f_w|x)}{\pi_{\text{ref}}(f_w|x)}\right) - \notag \\
&\quad\log\left(\frac{\pi_\theta(f_l|x)}{\pi_{\text{ref}}(f_l|x)}\right)\Bigg]\Bigg)^2
\label{eq:faaf_final_full_objective}
\end{align}

where $(f_w, f_l)$ represent the winning (preferred) and losing (less preferred) friction interventions respectively in each annotated pair.

\begin{align}
&\hat{\mathcal{L}} (\pi_\theta) = \mathbb{E}_{(x,\phi,f_w,f_l) \sim \mathcal{D}_\mu} \notag \\
&\quad\Bigg(1 - \beta\Bigg[\log\left(\frac{\pi_\theta(f_w|\phi,x)\pi_{\text{ref}}(f_l|\phi,x)}{\pi_\theta(f_l|\phi,x)\pi_{\text{ref}}(f_w|\phi,x)}\right) + \notag \\
&\quad\log\left(\frac{\pi_\theta(f_w|x)\pi_{\text{ref}}(f_l|x)}{\pi_\theta(f_l|x)\pi_{\text{ref}}(f_w|x)}\right)\Bigg]\Bigg)^2
\end{align}

\begin{align}
&\hat{\mathcal{L}} (\pi_\theta) = \mathbb{E}_{(x,\phi,f_w,f_l) \sim \mathcal{D}_\mu}  \notag \\
&\quad\Bigg(1 - \Bigg[
\underbrace{\beta \log\left(\frac{\pi_\theta(f_w|\phi,x)\pi_{\text{ref}}(f_l|\phi,x)}{\pi_\theta(f_l|\phi,x)\pi_{\text{ref}}(f_w|\phi,x)}\right)}_{\Delta R} + \notag \\
&\quad\underbrace{\beta \log\left(\frac{\pi_\theta(f_w|x)\pi_{\text{ref}}(f_l|x)}{\pi_\theta(f_l|x)\pi_{\text{ref}}(f_w|x)}\right)}_{\Delta R'}
\Bigg]\Bigg)^2
\label{eq:faaf_final_concise_objective}
\end{align} where $\Delta R$ and $\Delta R'$ represent implicit reward differences~\citep{rafailov2024direct, azar2024general}, the former being explicitly conditioned on the frictive state $\phi$, with no such conditioning on the latter.  

\begin{theorem}[Uniqueness of \fricabbr~Empirical Loss]
\label{theorem:faaf_uniqueness}
We prove this by contradiction. Let $\mu$ be the sampling distribution that samples friction interventions for the preference dataset, and assume $\texttt{Supp}(\mu) = \texttt{Supp}(\pi_{\text{ref}})$. Then the FAAF loss $\mathcal{L}(\pi)$ has a unique solution in policy space $\in \Pi$.
\end{theorem}
\vfill

\begin{figure*}[h!]
\begin{tcolorbox}
\begin{proof}
 
Assume by contradiction that there exist two distinct optimal policies $\pi_A, \pi_B \in \Pi$. By their definition, $\mathcal{\hat{L}}(\pi_A) =\mathcal{\hat{L}}(\pi_A) = 0$ as $\pi_A$ and $\pi_B$ are global minima. Consider $(s_\phi^A, s^A)$ and $(s_\phi^B, s^B)$ as their respective logit parameterizations where:
\begin{align*}
\pi_k(f|\phi) &= \frac{\exp(s_\phi^k(f))}{\sum_{f'}\exp(s_\phi^k(f'))} \\
\pi_k(f) &= \frac{\exp(s^k(f))}{\sum_{f'}\exp(s^k(f'))} \quad \text{for } k \in \{A,B\}
\end{align*} where $\pi_k(f|\phi)$ and $\pi_k(f)$ are the $\phi$-conditioned and $\phi$-unconditioned policies.

By the structure of our \fricabbr~loss from \Cref{eq:faaf_final_concise_objective}:
\begin{align*}
\mathcal{\hat{L}}(\pi) &= \mathbb{E}_{f,f' \sim \mu}\Big[\big(1 - \beta(\Delta s_\phi + \Delta s)\big)^2\Big] \geq 0
\end{align*}  

Notice that adding a constant $c$ to all logits of $s_\phi$ or logits of $s$ (directionally denoted as the $(c,\ldots,c) \in \mathbb{R}$) does not affect either policy probabilities due to softmax normalization. For $\mathcal{\hat{L}}(\pi)$, this is the \textit{only} direction  where the loss function might not be strictly convex. Outside of these directions, any change in the logits would increase $\mathcal{L}(\pi)$ with strict convexity as a consequence for $\alpha \in (0,1)$, implying:
 
\begin{align*}
\mathcal{\hat{L}}(\alpha\pi_1 + (1-\alpha)\pi_2) &< \alpha\mathcal{\hat{L}}(\pi_1) + (1-\alpha)\mathcal{\hat{L}}(\pi_2) \\
&= \alpha(0) + (1-\alpha)(0) = 0
\end{align*}  

where the equality follows from $\pi_1, \pi_2$ being global minima, by definition. This contradicts the non-negativity of $\mathcal{\hat{L}}$, which proves the uniqueness of the \fricabbr~objective. 
\end{proof}
\end{tcolorbox}
\end{figure*}

\paragraph{$\hat{\mathcal{L}} (\pi_\theta)$ has no dependence on log-partition terms involving $Z^*(\phi,x)$ and $Z^*(x)$} Our final \fricabbr~empirical objective loss in Eq.~\ref{eq:faaf_final_concise_objective} has no dependence on either partition function terms. This makes it convenient for practical applications. In fact, similar to DPO's derivation~\citep{rafailov2024direct}, these log-partition terms effectively cancel out in formulating the frictive state-conditioned and unconditioned implicit rewards, scaled by the KL-strength parameter $\beta$. In its essence, $\hat{\mathcal{L}} (\pi_\theta)$ regresses the DPO-based implicit rewards ($\Delta R'$ term) with an additional $\phi$-conditioned reward term ($\Delta R$ term) onto the empirically observed preference probabilities, labeled with preference labels from $\mathcal{D}_\mu$. Notice that without the $\Delta R$ term,  $\hat{\mathcal{L}} (\pi_\theta)$ reduces to a structurally similar form as IPO~\citep{azar2024general}, differing a constant scaling term $\beta$. This suggests that under this condition, both $\hat{\mathcal{L}} (\pi_\theta)$ and IPO objective likely have similar qualitative loss landscapes though convergence rates and optimal solutions would differ---while both lead $\pi_\theta$ toward a reward-consistent preference alignment. This also explains the somewhat similar performance of the IPO baseline and \fricabbr$_{\Delta R'}$ in both DeliData and WTD OPT 1.3B reward model-based win-rate evaluations, where \fricabbr$_{\Delta R'}$ achieves comparatively middling win-rates (Table~\ref{tab:opt_rm_performance_comparison_percentage}).



\begin{lemma}[Sequential Choice Decomposition in Friction Agent Optimization]
\label{lemma:threshold_theorem_sequential_choice}
Consider the minimax optimization between frictive-state policy $\pi_\phi$ and friction intervention policy $\pi_f$ where we seek to generate optimal friction interventions $f$ from frictive states $\phi$:
\begin{align}
&J^* = \min_{\pi_\phi} \max_{\pi_f} \mathbb{E}_{
\substack{x \sim \rho \\ \phi \sim \pi_\phi(\cdot \mid x) \\ f \sim \pi_f(\cdot \mid \phi, x)}
}\Big[p(f \succ \phi \mid x) - \notag \\
&\quad\beta D_{\text{KL}}(\pi_f \parallel \pi_{\text{ref}} \mid \phi, x) 
+ \beta D_{\text{KL}}(\pi_\phi \parallel \pi_{\text{ref}} \mid x)
\Big]
\end{align}

For any policy $\pi$ (either optimal friction policy $\pi_f^*$ or reference policy $\pi_{\text{ref}}$), the sequential choice probability decomposes as:
\begin{equation}
\pi(\phi|\phi,x) = \frac{\pi(\phi|x)}{\pi(f|x)}
\end{equation}
\end{lemma}

\begin{figure*}[t!]
\begin{tcolorbox}
\begin{proof}
The key insight in deriving this decomposition lies in understanding how optimal friction interventions are generated sequentially from frictive states. For the optimal friction policy $\pi_f^*$, consider its probability space $P_{\pi_f^*}$. By definition of conditional probability, we have $\pi_f^*(\phi|\phi,x) = \frac{P_{\pi_f^*}(\phi, \phi | x)}{P_{\pi_f^*}(\phi | x)}$. This term is crucial as it captures the policy's propensity to maintain a frictive state rather than generate a friction intervention. Under choice independence\footnote{Assuming a single-step bandit setting \cite{rafailov2024direct, rafailov2024r}, choice independence holds since each frictive-state intervention is independent of past episodes. Using conditional probability, we express the joint probability under any policy \( \pi \) as \( P_{\pi}(\phi, \phi \mid x) = P_{\pi}(\phi \mid \phi, x) P_{\pi}(\phi \mid x) \). By choice independence, the probability of selecting \( \phi \) at the second step does not depend on the first selection given \( x \), i.e., \( P_{\pi}(\phi \mid \phi, x) = P_{\pi}(\phi \mid x) \). Substituting this, we obtain \( P_{\pi}(\phi, \phi \mid x) = P_{\pi}(\phi \mid x) P_{\pi}(\phi \mid x) \).

} within this policy space assuming a Markovian nature of friction intervention generation, we have $P_{\pi_f^*}(\phi, \phi | x) = P_{\pi_f^*}(\phi | x) P_{\pi_f^*}(\phi | x)$. With policy-specific preference probability symmetry~\cite{munos2023nash, fisch2024robustpreferenceoptimizationreward}, the probability $P_{\pi_f^*}(\phi | x) + P_{\pi_f^*}(f | x) = 1$, reflecting the binary choice between maintaining a frictive state or generating a friction intervention, we obtain $\pi_f^*(\phi|\phi,x) = \frac{\pi_f^*(\phi|x)}{\pi_f^*(f|x)}$, where the optimality of $\pi_f^*(f|x)$ ensures $\pi_f^*(\phi|x) \leq \pi_f^*(f|x)$. A similar argument can be made in the case of $\pi_{\text{ref}}$, the reference policy, where $\pi_{\text{ref}}$'s initialization with the supervised-finetuned (SFT) model on friction interventions ensures $\pi_{\text{ref}}(f|x) \geq \pi_{\text{ref}}(\phi|x)$. This decomposition is fundamental to the minimax objective, $J^*$, as it enables expressing the KL-regularized preference probability in terms of base policy probabilities while preserving the structure necessary for optimal friction intervention generation from frictive states.
\end{proof}
\end{tcolorbox}
\end{figure*}
\vfill
\clearpage

The sequential choice decomposition provides crucial insight into determining optimal timing for friction interventions. In other words, this decomposition has an interesting implication in deciding \textit{when} is a friction intervention most desirable or cost-effective.  Specifically, our derived identity $\pi(\phi|\phi,x) = \frac{\pi(\phi|x)}{\pi(f|x)}$ establishes a natural threshold mechanism through the ratio $\tau(x) = \frac{\pi_f^*(\phi|x)}{\pi_f^*(f|x)}$. When $\tau(x) \approx 1$, the policy maintains the current frictive state $\phi$, while $\tau(x) \ll 1$ triggers a friction intervention $f$. This mechanism emerges naturally from the preference probability $p(f \succ \phi|x) = \beta[\log(\pi_f^*(f|\phi,x)) - \log(\pi_{\text{ref}}(f|\phi,x)) + \frac{1}{2\beta} - (\log(\pi_f^*(\phi|\phi,x)) - \log(\pi_{\text{ref}}(\phi|\phi,x)))]$ in our minimax objective $J^*$, where $\pi_f^*$ optimally generates interventions when the likelihood ratio indicates low confidence in the current frictive state $\phi$. However, exploring this sequential decomposition and determining optimal timing in interventions is outside the scope of this paper. As such, we leave that for future work.

\begin{lemma}[Uniqueness of Intervention Thresholds]
The threshold $\tau(x) = \frac{\pi_f^*(\phi|x)}{\pi_f^*(f|x)}$ uniquely determines optimal intervention policy $\pi_f^*$.
\end{lemma}
\vfill

\begin{figure*}[h!]
\begin{tcolorbox}
\begin{proof}
 We prove uniqueness by contradiction. Consider two potentially optimal policies $\pi_f^1$ and $\pi_f^2$ with corresponding thresholds $\tau_1(x)$ and $\tau_2(x)$. Assume $\tau_1(x) \neq \tau_2(x)$ but both policies are optimal. By optimality, their contributions to the objective $J^*$ must be equal for any observation tuple $x$, $f$ and $\phi$:

\begin{equation}
\begin{aligned}
&\beta[\log(\pi_f^1(f|\phi,x)) - \log(\pi_{\text{ref}}(f|\phi,x)) - (\log(\tau_1(x)) - \log(\pi_{\text{ref}}(\phi|\phi,x)))] \\
&= \beta[\log(\pi_f^2(f|\phi,x)) - \log(\pi_{\text{ref}}(f|\phi,x)) - (\log(\tau_2(x)) - \log(\pi_{\text{ref}}(\phi|\phi,x)))]
\end{aligned}
\end{equation}

Simplifying and rearranging terms:
\begin{equation}
\log(\pi_f^1(f|\phi,x)) - \log(\tau_1(x)) = \log(\pi_f^2(f|\phi,x)) - \log(\tau_2(x))
\end{equation}

However, by the strict convexity of KL divergence and Jensen's inequality:
\begin{equation}
D_{\text{KL}}(\pi_f^1 \parallel \pi_{\text{ref}} \mid \phi, x) + D_{\text{KL}}(\pi_f^2 \parallel \pi_{\text{ref}} \mid \phi, x) > 2D_{\text{KL}}(\frac{\pi_f^1 + \pi_f^2}{2} \parallel \pi_{\text{ref}} \mid \phi, x)
\end{equation}

This inequality implies that a mixed policy $\pi_f^{\text{avg}} = \frac{\pi_f^1 + \pi_f^2}{2}$ would achieve a lower KL divergence cost due to strict convexity
and equal expected reward (regularized preference probabilities) from the equality of optimal policies. Therefore, $\pi_f^{\text{avg}}$ would achieve strictly better objective value than both $\pi_f^1$ and $\pi_f^2$, contradicting their assumed optimality. This proves threshold uniqueness. The contradiction arises because:
\begin{equation}
J^*(\pi_f^{\text{avg}}) > \frac{1}{2}[J^*(\pi_f^1) + J^*(\pi_f^2)]
\end{equation}
which is impossible if both $\pi_f^1$ and $\pi_f^2$ were truly optimal.
\end{proof}
\end{tcolorbox}
\end{figure*}

\clearpage
\begin{corollary}[Uniqueness of Optimal Policy Under Threshold Identity]
If two optimal intervention policies $\pi_f^1$ and $\pi_f^2$ satisfy the same threshold condition $\tau_1(x) = \tau_2(x)$ for all $x$, then $\pi_f^1 = \pi_f^2$.
\end{corollary}

\begin{tcolorbox}
\begin{proof}
Assume for contradiction that two distinct optimal policies $\pi_f^1$ and $\pi_f^2$ satisfy the threshold condition $\frac{\pi_f^1(\phi|x)}{\pi_f^1(f|x)} = \frac{\pi_f^2(\phi|x)}{\pi_f^2(f|x)} = \tau(x)$. Define the mixed policy $\pi_f^{\text{avg}} = \frac{1}{2}(\pi_f^1 + \pi_f^2)$, which preserves the threshold as $\tau_{\text{avg}}(x) = \tau(x)$ due to linearity, implying $\pi_f^{\text{avg}}$ is also optimal. 

Now, applying Jensen’s inequality to the KL divergence term in the objective, we obtain $D_{\text{KL}}(\pi_f^{\text{avg}} \parallel \pi_{\text{ref}} \mid \phi, x) \leq \frac{1}{2} D_{\text{KL}}(\pi_f^1 \parallel \pi_{\text{ref}} \mid \phi, x) + \frac{1}{2} D_{\text{KL}}(\pi_f^2 \parallel \pi_{\text{ref}} \mid \phi, x)$. Strict convexity ensures a strict inequality whenever $\pi_f^1 \neq \pi_f^2$ on a set of positive measure where $\operatorname{supp}(\pi_{\text{ref}}) > 0$, implying $J^*(\pi_f^{\text{avg}}) < \frac{1}{2} [J^*(\pi_f^1) + J^*(\pi_f^2)]$. This contradicts the assumed optimality of $\pi_f^1$ and $\pi_f^2$, proving that they must be identical.
\end{proof}
\end{tcolorbox}

\begin{lemma}[Policy Ratio Equivalence]
\label{lemma:policy_ratio_equivalence}
For the optimal friction policy $\pi_f^*$ and the optimal frictive-state policy $\pi_\phi^*$, the following expectation-based ratio holds:
\begin{equation}
    \mathbb{E}_{
    \substack{x \sim \rho \\ \phi \sim \pi_\phi^*(\cdot \mid x) \\ f \sim \pi_f^*(\cdot \mid \phi, x)}
    } \left[ \frac{\pi_f^*(\phi|x)}{\pi_f^*(f|x)} \right] 
    = \mathbb{E}_{
    \substack{x \sim \rho \\ \phi \sim \pi_\phi^*(\cdot \mid x) \\ f \sim \pi_f^*(\cdot \mid \phi, x)}
    } \left[ \frac{\pi_\phi^*(\phi|x)}{\pi_\phi^*(f|x)} \right].
\end{equation}
\end{lemma}

\begin{figure*}[h!]
\begin{tcolorbox}
\begin{proof}
We show that both the policy ratios simplify to the same value under the expectation. We begin by taking the expectation over the preference probability formulation\footnote{For clarity, the expectation $\mathbb{E}$ is taken over ${x \sim \rho, \phi \sim \pi_\phi^*(\cdot \mid x) , f \sim \pi_f^*(\cdot  \mid\phi, x)}$ throughout the proof, but this is omitted in the notation when the context is clear.}:
\begin{align}
\mathbb{E} 
\left[ p(f \succ \phi \mid x) \right] 
&= \mathbb{E}
\left[ \beta \left( \log(\pi_f^*(f|\phi,x)) - \log(\pi_{\text{ref}}(f|\phi,x)) + \log Z^*(\phi,x) \right) \right].
\end{align}
 
We first represent the ratios of the optimal frictive intervention policies (LHS of this lemma) for any tuple \((x, \phi, f)\) in terms of their parametric representations from Eq.~\ref{eq:optimal_frictive_agent_policy} as follows:

\begin{align}
\frac{\pi_f^*(\phi|x)}{\pi_f^*(f|x)} &= \frac{\pi_{\text{ref}}(\phi|x)}{\pi_{\text{ref}}(f|x)} e^{\beta^{-1} (p(\phi \succ f|x) - p(f \succ \phi|x))} \quad \text{($\log Z^*(x)$ cancels out)} 
\end{align}

Take the expectation on both sides and apply\footnote{Since learning occurs in a supervised setting with preference-annotated data, the probability follows as $p(f \succ \phi \mid x) = \mathbb{E}[{1}(f \succ \phi \mid x)] = 1$, implying $p(\phi \succ f \mid x) = 0$.}  the identity $p(\phi \succ f \mid x) = 0$:

\begin{align}
\mathbb{E} \left[ \frac{\pi_f^*(\phi|x)}{\pi_f^*(f|x)} \right] 
&= \mathbb{E} 
\left[ \frac{\pi_{\text{ref}}(\phi|x)}{\pi_{\text{ref}}(f|x)} e^{-\beta^{-1} p(f \succ \phi|x)} \right] 
\quad \text{(since $p(\phi \succ f \mid x) = 0$)}.
\label{eq:ratio_frictive_intervention_policy}
\end{align}

Notice that by definition in Eq.~\ref{eq:two_stage_main_objective_main_paper}, the optimal friction intervention policy $\pi_f^*(\cdot|\phi, x)$ is KL-constrained wrt to the reference policy $\pi_{\text{ref}}(\cdot|\phi, x)$. So under the expectation, the following has to be true for $\pi_f^*(\cdot|\phi, x)$ to be optimal: 
\begin{align}
\mathbb{E} 
\left[ \pi_f^*(f|\phi, x) \right] 
\approx \mathbb{E}
\left[ \pi_{\text{ref}}(f|\phi, x) \right].
\label{eq:optimal_friction_intervention_equal_ref}
\end{align}

Substituting the preference probability formulation $p(f \succ \phi|x)$ from Eq.~\ref{eq:preference_prob_friction_over_frictive_state} in Eq.~\ref{eq:ratio_frictive_intervention_policy} and applying the KL-regularization approximation in Eq.~\ref{eq:optimal_friction_intervention_equal_ref} we derive that: 
\begin{equation}
\mathbb{E} \left[ e^{-\left( \log(\pi_f^*(f|\phi,x)) - \log(\pi_{\text{ref}}(f|\phi,x)) + \log Z^*(\phi,x) \right)} \right] 
\approx \mathbb{E} \left[ \frac{Z^*(f,x)}{Z^*(\phi,x)} \right].
\end{equation}
Using this substitution, we rewrite Eq.~\ref{eq:ratio_frictive_intervention_policy} as:
\begin{align}
\mathbb{E} \left[ \frac{\pi_f^*(\phi|x)}{\pi_f^*(f|x)} \right] 
&= \mathbb{E} 
\left[ \frac{\pi_{\text{ref}}(\phi|x)}{\pi_{\text{ref}}(f|x)} 
e^{-\left( \log(\pi_f^*(f|\phi,x)) - \log(\pi_{\text{ref}}(f|\phi,x)) + \log Z^*(\phi,x) \right)} \right] \notag \\
&= \mathbb{E} \left[ \frac{\pi_{\text{ref}}(\phi|x)}{\pi_{\text{ref}}(f|x)} \frac{Z^*(f,x)}{Z^*(\phi,x)} \right].
\end{align}

Similarly, for the optimal frictive state policy ratio we derive: 
\begin{align}
\mathbb{E}
\left[ \frac{\pi_\phi^*(\phi|x)}{\pi_\phi^*(f|x)} \right] 
&= \mathbb{E}
\left[ \frac{\frac{\pi_{\text{ref}}(\phi|x)}{Z_\phi^*(x)} e^{-\beta^{-1} \log Z^*(\phi,x)}}{\frac{\pi_{\text{ref}}(f|x)}{Z_\phi^*(x)} e^{-\beta^{-1} \log Z^*(f,x)}} \right]
= \mathbb{E}
\left[ \frac{\pi_{\text{ref}}(\phi|x)}{\pi_{\text{ref}}(f|x)} \frac{e^{-\log Z^*(\phi,x)}}{e^{-\log Z^*(f,x)}} \right] \quad \text{($Z_\phi^*(x)$ cancels)} \\
&= \mathbb{E}
\left[ \frac{\pi_{\text{ref}}(\phi|x)}{\pi_{\text{ref}}(f|x)} \frac{Z^*(f,x)}{Z^*(\phi,x)} \right].
\end{align}
Thus, $
\mathbb{E}
\left[ \frac{\pi_f^*(\phi|x)}{\pi_f^*(f|x)} \right] 
= \mathbb{E}
\left[ \frac{\pi_\phi^*(\phi|x)}{\pi_\phi^*(f|x)} \right]$.
\end{proof}
\end{tcolorbox}
\end{figure*}
\clearpage

 \section{Operationalizing $\mu$: Frictive State and Friction Intervention Generations}
 \label{appendices:generation}

In order to train and evaluate our baselines along with \fricabbr~for friction intervention generation in collaborative tasks, we carry out a series of data augmentation procedures using GPT-4o (denoted as $\mu$) in order to construct two diverse preference datasets. For details on our choice of datasets, please refer to Sec.~\ref{ssec:datasets}.

In this section, we provide procedural details of our friction intervention datasets, that were generated out of the original Weights Task and DeliData dataset. For all our data-generation experiments, we use a high-capacity LLM (GPT-4o)~\citep{openai2024gpt4ocard} as our sampling distribution $\mu$, as defined in Sec.~\ref{sec:task-formulation}. In particular, we utilize a \textbf{self-rewarding LLM approach}~\citep{yuan2024self,xu2023some, Rosset2024DirectNO} to \textit{simultaneously} generate and assign rewards to $\mu$-generated interventions, since previous work~\citep{pace2024westofnsyntheticpreferencesselfimproving, meng2024simposimplepreferenceoptimization} provides evidence that such synthetic preference-data generation still leads to higher-quality reward models and preference-aligned policies. Prior work~\cite{zheng2023judging} provides substantial evidence that this approach leads to more high-quality LLM-as-a-judge-based evaluations especially for conversational benchmarks~\cite{lambert2024rewardbench}. Additionally, reward assignments for sampled intervention naturally provides an implicit preference ranking---which we use for constructing our respective preference datasets. After these data-generation experiments, we further conduct filtering and contrastive pairing of a "winning" ($f_w$) or preferrred interventions and "losing" ($f_l$) or dispreferred interventions along with their corresponding dialogue histories ($x$) to create our final preference datasets for each augmented dataset.

\begin{table*}[h!]
    \centering
    \small
    \renewcommand{\arraystretch}{1.2} 
    \begin{tabular}{lcccccc}
        \toprule
        \textbf{} & \multicolumn{3}{c}{\textbf{Train}} & \multicolumn{3}{c}{\textbf{Test}} \\
        \cmidrule(lr){2-4} \cmidrule(lr){5-7}
        & \textbf{Min} & \textbf{Max} & \textbf{Mean ± Std} & \textbf{Min} & \textbf{Max} & \textbf{Mean ± Std} \\
        \midrule
        \textbf{Dialogue History} & 16  & 824  & 288.43 ± 132.97 & 25  & 733  & 291.88 ± 118.03 \\
        \textbf{Belief State} & 8  & 140  & 32.93 ± 15.92  & 20  & 140  & 47.99 ± 28.38 \\
        \textbf{Chosen Friction} & 6  & 60  & 24.03 ± 4.65  & 9   & 39   & 22.05 ± 5.55 \\
        \textbf{Chosen Rationale} & 8  & 78  & 22.84 ± 8.67  & 10  & 78   & 29.61 ± 13.33 \\
        \textbf{Rejected Friction} & 9  & 45  & 23.95 ± 4.10  & 10  & 41   & 22.16 ± 5.11 \\
        \textbf{Rejected Rationale} & 8  & 73  & 19.60 ± 6.89  & 10  & 59   & 26.04 ± 11.61 \\
        \bottomrule
    \end{tabular}
    \caption{Token Length Statistics for the \textbf{DeliData Friction} dataset using the \texttt{Meta-Llama-3-8B-Instruct} tokenizer.}
    \label{tab:deli_preference_token_stats}
\end{table*}
\begin{table*}[h!]
    \centering
    \small
    \renewcommand{\arraystretch}{1.2} 
    \begin{tabular}{lcccccc}
        \toprule
        \textbf{Field} & \multicolumn{3}{c}{\textbf{Train}} & \multicolumn{3}{c}{\textbf{Test}} \\
        \cmidrule(lr){2-4} \cmidrule(lr){5-7}
        & \textbf{Min} & \textbf{Max} & \textbf{Mean ± Std} & \textbf{Min} & \textbf{Max} & \textbf{Mean ± Std} \\
        \midrule
        \textbf{Dialogue History} & 4  & 1464  & 227.83 ± 189.48 & 4  & 1031  & 235.04 ± 180.36 \\
        \textbf{Belief State} & 11  & 65  & 30.55 ± 6.65  & 17  & 54  & 30.47 ± 6.29 \\
        \textbf{Chosen Friction} & 10  & 45  & 21.20 ± 5.12  & 11  & 42  & 21.08 ± 5.10 \\
        \textbf{Chosen Rationale} & 10  & 35  & 20.38 ± 3.44  & 12  & 32  & 19.67 ± 3.38 \\
        \textbf{Rejected Friction} & 6  & 32  & 15.88 ± 3.68  & 7   & 29  & 15.57 ± 3.75 \\
        \textbf{Rejected Rationale} & 8  & 41  & 20.10 ± 3.51  & 12  & 30  & 19.88 ± 3.47 \\
        \bottomrule
    \end{tabular}
     \caption{Token Length Statistics for the \textbf{WTD Simulated Friction} dataset using the \texttt{Meta-Llama-3-8B-Instruct} tokenizer.}
    \label{tab:wtd_data_token_stats}
\end{table*}
\begin{table*}[h!]
    \centering
    \small
    \renewcommand{\arraystretch}{1.2} 
    \begin{tabular}{lcccccc}
        \toprule
        \textbf{Field} & \multicolumn{3}{c}{\textbf{Train}} & \multicolumn{3}{c}{\textbf{Test}} \\
        \cmidrule(lr){2-4} \cmidrule(lr){5-7}
        & \textbf{Min} & \textbf{Max} & \textbf{Mean ± Std} & \textbf{Min} & \textbf{Max} & \textbf{Mean ± Std} \\
        \midrule
        \textbf{Dialogue History} & 16  & 555  & 309.88 ± 81.11 & 25  & 555  & 316.46 ± 79.16 \\
        \textbf{Belief State} & 41  & 140  & 84.94 ± 15.58  & 41  & 140  & 84.95 ± 16.27 \\
        \textbf{Chosen Friction} & 9  & 31  & 16.85 ± 3.47  & 9   & 27   & 16.87 ± 3.49 \\
        \textbf{Chosen Rationale} & 24  & 78  & 44.19 ± 8.46  & 26  & 78   & 44.43 ± 8.59 \\
        \textbf{Rejected Friction} & 9  & 31  & 17.12 ± 3.51  & 10  & 28   & 17.23 ± 3.41 \\
        \textbf{Rejected Rationale} & 24  & 73  & 40.00 ± 6.62  & 24  & 59   & 39.89 ± 6.43 \\
        \bottomrule
    \end{tabular}
    \caption{Token Length Statistics for the \textbf{WTD Original Friction} dataset using the \texttt{Meta-Llama-3-8B-Instruct} tokenizer.}
    \label{tab:wtd_original_token_stats}
\end{table*}
\begin{table*}[h!]
\centering
\resizebox{\textwidth}{!}{%
\small
\begin{tabular}{@{}p{4cm}p{4cm}p{8cm}@{}}
\toprule
\textbf{Personality Type} & \textbf{Facet} & \textbf{Description} \\ 
\midrule
Extraversion   & Assertiveness       & Tends to take charge and speak confidently. \\ 
               & Sociability         & Enjoys engaging with others and maintaining conversation. \\ 
               & Activity Level      & Shows high energy and enthusiasm. \\ 
               & Excitement Seeking  & Looks for novel and stimulating experiences. \\ 
               & Positive Emotions   & Expresses optimism and cheerfulness. \\ 
\midrule
Neuroticism    & Anxiety             & Shows worry and concern about potential mistakes. \\ 
               & Depression          & Tends to be pessimistic and doubtful. \\ 
               & Vulnerability       & Easily becomes overwhelmed or stressed. \\ 
               & Self-Consciousness  & Shows hesitation and uncertainty. \\ 
               & Anger               & Can become frustrated and irritated easily. \\ 
\midrule
Agreeableness  & Trust               & Readily trusts others and their suggestions. \\ 
               & Altruism            & Shows concern for others' success and well-being. \\ 
               & Compliance          & Tends to avoid conflicts and agree with others. \\ 
               & Modesty             & Downplays own contributions and abilities. \\ 
               & Sympathy            & Shows understanding and empathy towards others. \\ 
\bottomrule
\end{tabular}
}
\caption{Descriptions of our chosen 3 personality types and facet combinations from the Big Five framework that we use for simulated friction generation on the Weights Task. }
\label{tab:personality_facet_descriptions}
\end{table*}

\subsection{DeliData Friction Intervention Preference Dataset}
\label{app:deli_preference_data_generation}

In order to generate frictive state and friction interventions in the DeliData dataset, we use the prompt shown in \Cref{fig:deli_friction_generation_prompt}. In order to contextualize the extraction of frictive states, we only provide $h = 15$ previous utterances in each dialogue group (\texttt{group\_id}) assuming that frictive states are likely to be present within an "attentional-state"~\citep{grosz1986attention} window that describes the focused part in the discourse. This technique allows us to avoid unnecessary API calls while also providing a more focused dialogue context to GPT-4o. Additionally, since this dataset already contains manual human annotations of "probing" interventions (which, per our definitions in Sec.~\ref{sec:defs}, constitute a subset of friction interventions), we explicitly guide the data-generator to exclude probing interventions in extracting the frictive states. Note that each functionally-frictive state (denoted as $\phi$), as extracted by GPT-4o, resulted in two friction interventions, $f_w$ and $f_l$. In total, this generation process led to 6,238 ($x, \phi, f_w, f_l$) tuples after keeping 50 randomly sampled dialogue groups separate for the evaluation set, out of which 476 (33) were probing interventions in train (test) partitions. Additionally, we carry out another round of training pair augmentations since 6,238 samples is quite small compared to popular preference alignment datasets, such as Ultrafeedback's roughly 62k training preference pairs~\cite{Cui2024UltraFeedbackBL}.\footnote{We found 14 samples where GPT-4o did not return any strings for the frictive state description. We filtered out these samples from our training set.} The average rewards for the preferred and dispreferred interventions assinged by $\mu$ are 8.03 and 3.96 respectively (rated out of 10).

As such, for each training tuple \((x, \phi, f_w, f_l)\), we generate $N$ augmented versions~\((x', \phi', f_w', f_l')\) by applying a replacement mapping \(R: \Sigma \to \Sigma'\) $N$ times, where \(\Sigma\) represents the original set of card values (vowels\footnote{We did not replace instances of "I" to avoid noise from mistakenly replacing first-person references in the dialogues. Additionally, since vowels constitute the majority of prompted card solutions vs. consonants, applying our replacement function $R$ for vowels was enough to generate $\sim$62k additional samples, comparable to Ultrafeedback~\cite{Cui2024UltraFeedbackBL}}, odd numbers, and even numbers), and \(\Sigma'\) represents their replacements. The replacement function \(R\) is defined as follows:  Each vowel \(v \in \{A, E, O, U\}\) is replaced with another vowel \(v'\) such that \(v' \in \{A, E, O, U\} \setminus \{v\}\), where \(v'\) is sampled uniformly at random from the remaining vowels. Similarly, each odd number \(o \in \{1, 3, 5, 7, 9\}\) is replaced with another odd number \(o'\) such that \(o' \in \{1, 3, 5, 7, 9\} \setminus \{o\}\), where \(o'\) is sampled uniformly at random. Likewise, each even number \(e \in \{0, 2, 4, 6, 8\}\) is replaced with another even number \(e'\) such that \(e' \in \{0, 2, 4, 6, 8\} \setminus \{e\}\), with \(e'\) sampled uniformly at random. For example, if an utterance contains reference to card "A" and "6", the rules of the Wason Card task still applies equivalently for, say, "E" and "8"--while keeping the reasoning consistent with the original utterance and the utterance with replacement. We apply this replacement mapping across all fields in tuples ($x', \phi', f_w', f_l'$) in the training set. This led to training set of 68,618 preference pairs. Note that we only apply this augmentation for the training set to generate a reasonably large preference dataset for more robust training signals. Table~\ref{tab:deli_preference_token_stats} shows a detailed breakdown of the token-length statistics of the DeliData Friction preference dataset using the \texttt{Meta-Llama-3-8B-Instruct} tokenizer. 



 
\subsection{WTD Friction Intervention Preference Dataset}
\label{app:wtd_preference_data_generation}

\paragraph{WTD "Original" Friction dataset}

Unlike the DeliData dataset, which includes pre-annotated probing interventions as natural friction points, the Weights Task dataset~\citep{khebour2024text} consists of dense-paraphrased utterances transcribed manually~\cite{terpstra2023good} and with Whisper~\citep{radford2023robust}, making friction interventions sparse due to its multimodal nature. Manual inspection found only 3-4 frictive interventions per group, yielding $\approx$ 30-40 samples—insufficient for training an effective agent without overfitting, especially for LLMs with billions of parameters. As such, we carry out two phases of data-augmentations and preference annotations. In our first round, we generate the \textbf{WTD Original Friction} dataset which contains annotations of frictive-states and friction interventions. Similar to DeliData preference annotations~\Cref{app:deli_preference_data_generation}, we use a self-rewarding LLM set-up to first generate these states and interventions in an autoregressive manner and prompt $\mu$ to rate them in the same api-call, for each frictive state extraction. Since WTD dialogues can be substantially long ($>$ 200 utterances) for certain groups, we only consider a non-overlapping window of 10 previous utterances as context history  $h = 10$ for a more robust grounding for $\mu$; See Fig.~\ref{fig:deli_friction_generation_prompt} for details on the prompt used constructing the  \textbf{WTD Original Friction} dataset. This process led to 4299 (470) training (testing) preference pairs. Preferred interventions achieved mean scores (mean±std) of 8.36±1.12 (train) and 8.35±1.08 (test), while dispreferred interventions scored 6.35±1.13 (train) and 6.36±1.11 (test), demonstrating consistent preference margins across splits.

Note that we do not use \textbf{WTD Original Friction} for training any of our baselines---but use it for out-of-domain distribution (OOD) evaluation (see Sec.~\ref{ssec:datasets}). This allows us to more extensively evaluate \fricabbr~in checking test-time OOD generalization~\citep{rafailov2024direct, choi2024robust} against baselines---where OOD generalization is a major limitation in supervised preference alignment algorithms that depend crucially on the sampling distribution~\cite{yang2024regularizinghiddenstatesenables, fisch2024robustpreferenceoptimizationreward}.

\paragraph{WTD "Simulated" Friction dataset}

Additionally, for a more robust training and in order to evaluate multi-turn preference alignment in interventions, we use \citep{shani2024multiturnreinforcementlearningpreference}'s method to generate novel full collaborative conversations using the weight-definitions of the original WTD environment. This method is more akin to "West-of-N" sampling~\citep{pace2024westofnsyntheticpreferencesselfimproving} techniques that allow synthetic data generations with high-capacity LLMs---where highest and lowest rewarded candidates naturally form preference pairs. As shown in Fig.~\ref{fig:wtd_friction_generation_prompt_simulated}, we sample a full dialogue at once using $\mu$, while providing initial task-related guidelines and gold-truth labels of actual weights of the five blocks in the WTD dataset. For example, we explicitly prompt $\mu$ to role-play~\citep{li2023camel} the triad consisting of three participants in the weight-deduction process. Furthermore, to generate more realistic utterances, we utilize participant personality-facet combinations~ \citep{pan2023llms, mao2024editing} from Big 5\footnote{See Tab.~\ref{tab:personality_facet_descriptions} for our full set of personality-type and facet combinations. Similar to \citep{mao2024editing}, we choose three personality types from Big 5 framework for consistency.} personality classifications \cite{goldberg2013alternative} as additional attributes in the prompt. In other words, each sampled full-dialogue contains a unique combination of these personality-facet combinations for each participant (total 3,375 combinations).

Similarly, for each sampled frictive state within a conversation (dialogue), we generated $N=6$ friction interventions with corresponding effectiveness scores in resolving the frictive state. Since WTD data does not contain any probing intervention samples, in order to further ground these generations to the task, we also provide a one-shot example of a naturally occurring friction intervention (marked with $P1 (f)$ in Fig.~\ref{fig:wtd_friction_generation_prompt_simulated}). In total, out of the expected 3,375 personality-facet combinations (3*5 unique combinations for each participant), 3,362 were successfully generated using $\mu$ and parsed. Finally, to create the preference pairs, for each frictive state, we paired the lowest scoring response with all the higher scoring ones, akin to the West-of-N technique. This resulted in 56, 689 preference pairs after excluding 54 dialogues (amounting to 800 preference pairs) for the test set. This process finally resulted in the \textbf{WTD Simulated Friction} dataset. Preferred interventions achieved mean scores of 8.48±1.52 (train) and 8.51±1.50 (test), while dispreferred interventions scored 6.01±0.88 (train) and 6.08±0.87 (test), demonstrating consistent preference margins across splits.

\begin{figure*}[htbp]

\begin{tcolorbox}[width=\textwidth,title={\sc Friction Generation Prompt: Delidata Dataset}]

\textbf{System}: You are an expert in collaborative reasoning and dialogue analysis. Your task is to detect *frictive states* and generate *friction interventions* that resolve them in group dialogue. 
A frictive state occurs when a participant makes a claim that contradicts another 
participant's belief model (i.e., their assumed understanding of the rule or task
constraints), leading to misalignment in reasoning that could hinder progress. Friction interventions encourage self-reflection in participants and prompt them to reevaluate these contradicting beliefs and assumptions. 
\\
\\
\textbf{User}: Analyze this dialogue about the Wason card selection task. Participants see four cards showing numbers or letters and must test this rule: "All cards with vowels on one side have an even number on the other." Remember that the correct answer is to select a vowel and an odd number. Provide [N] frictive states with their resolutions in the following JSON 
format. For each state, include both a preferred and less preferred intervention
that could help resolve the conflict. Additionally, provide a one-sentence rationale for your intervention.
\\
\\
IMPORTANT: 
- Do not analyze utterances labeled as "probing" or statements immediately before them, as these frictive states have already been detected.\\
- For each frictive state detected, you should:\\
  * Identify the dialogue index where it occurs\\
  * Summarize the conflicting beliefs\\
  * Explain why the contradiction affects reasoning\\
\\
\\
Here is the provided dialogue:\\
$[$Dialogue$]$\\
\\
Message ID: $[$index\_where\_friction\_occurs$]$\\
Contradiction: $[$describe\_the\_conflicting\_beliefs$]$\\
Contradiction Reason: $[$explain\_why\_the\_contradiction\_affects\_reasoning$]$\\
\\
\textbf{Preferred Intervention}:\\
   Statement: $[$your\_friction\_intervention$]$\\
   Rationale: $[$your\_rationale$]$\\
   Score: $[$your\_score$]$\\
\\
\textbf{Less Preferred Intervention}:\\
   Statement: $[$your\_friction\_intervention$]$\\
   Rationale: $[$your\_rationale$]$\\
   Score: $[$your\_score$]$

\end{tcolorbox}

\centering
\caption{DeliData~\citep{karadzhov2023delidata} Friction Generation Prompt. We use GPT-4o as our sampling distribution $\mu$ and prompt it to simultaneously generate frictive states and friction interventions. For diversity, we use the default temperature of 1. This process implicitly provides us with preference rankings between intervention, via the reward scores. See Sec.~\ref{sec:defs} for definitions of frictive states and friction interventions. Note that we exclude already-present “probing” interventions in this generation process since are present in the original DeliData annotations.  }
\label{fig:deli_friction_generation_prompt}
\end{figure*}

\begin{figure*}[htbp]

\begin{tcolorbox}[width=\textwidth,title={\sc Friction Generation Prompt: Weights Task Dataset (WTD Original)}]

\textbf{System}: You are an expert in collaborative reasoning and dialogue analysis. Your task is to detect *frictive states* and generate *friction interventions* that resolve them in group dialogue. 
A frictive state occurs when a participant makes a claim that contradicts another 
participant's belief model (i.e., their assumed understanding of the rule or task
constraints), leading to misalignment in reasoning that could hinder progress. Friction interventions encourage self-reflection in participants and prompt them to reevaluate these contradicting beliefs and assumptions. 
\\
\\
\textbf{User}: Analyze this dialogue about the Weights Task dataset. Three participants (P1, P2, and P3) are collaborating to determine the weights of colored blocks using a scale. \\
\\
Block Weights (in grams): \\
- Red block: 10g \\
- Blue block: 10g \\
- Green block: 20g \\
- Purple block: 30g \\
- Yellow block: 50g \\
\\
Game Rules: \\
1. Participants can only weigh two blocks at a time \\
2. They are told the red block's weight at the start \\
3. All other block weights are initially unknown \\
4. Scale slider is not needed (blocks are in 10g increments) \\
\\
 Provide [N] frictive states with their resolutions in the following JSON 
format. For each state, include both a preferred and less preferred intervention
that could help resolve the conflict. Additionally, provide a one-sentence rationale for your intervention. 
\\
Here is the provided dialogue:
\\
$[$Dialogue$]$\\
\\
Message ID: $[$index\_where\_friction\_occurs$]$\\
Contradiction: $[$describe\_the\_conflicting\_beliefs$]$\\
Contradiction Reason: $[$explain\_why\_the\_contradiction\_affects\_reasoning$]$\\
\\
\textbf{Preferred Intervention}:\\
   Statement: $[$your\_friction\_intervention$]$\\
   Rationale: $[$your\_rationale$]$\\
   Score: $[$your\_score$]$\\
\\
\textbf{Less Preferred Intervention}:\\
   Statement: $[$your\_friction\_intervention$]$\\
   Rationale: $[$your\_rationale$]$\\
   Score: $[$your\_score$]$

\end{tcolorbox}

\centering
\caption{Weights Task dataset~\citep{khebour-etal-2024-common} Friction Generation Prompt. We use GPT-4o as our sampling distribution $\mu$ and prompt it to simultaneously generate frictive states and friction interventions. For diversity, we use the default temperature of 1.  }
\label{fig:wtd_friction_generation_prompt}
\end{figure*}

\begin{figure*}[htbp]

\begin{tcolorbox}[width=\textwidth,title={\sc Friction Generation Prompt: Weights Task Dataset (WTD Simulated)}]

\textbf{System}: You are an expert in collaborative reasoning and dialogue analysis. Your task is to *generate a complete dialogue* where participants (P1, P2, P3) discuss which block to measure next and how to measure them. The three participants have distinct personality types that influence their behavior and dialog must reflect
these personality traits in their communication style and behavior. The dialog is considered complete when all block weights are measured and agreed upon. Additionally, identify frictive states within the dialogue and provide N friction interventions at these points.\\
 
$[$Definition: Frictive State$]$\\
$[$Definition: Friction Intervention$]$\\

\textbf{User}: Three participants (P1, P2, P3) work together in the Weights Task to determine the weights of colored blocks (red=10g, blue=10g, green=20g, purple=30g, yellow=50g). They can only weigh two blocks at a time, start knowing only the red block's weight, and use a scale with 10g increments (no slider needed).  \\
 \textbf{Your tasks}:\\
Generate a full dialogue until all weights are correctly identified and agreed upon.\\
Identify frictive states where reasoning misalignment occurs.\\
Provide N friction interventions with their corresponding rationales at these points. Rank them by effectiveness in resolving the conflict. Assign each a quality score from 1 to 10.\\

P1 has \textbf{\{personality\_type\}} personality type with high \textbf{\{personality\_facet\}}.
 
Here is an example dialogue where friction statements are labeled as (f). Actions of participants are provided within “[]” blocks. \\ 
\\
P2: [pointing towards the purple block first and then towards the blue block] I think this one is purple and this one is blue. \\
P3: [reading from the laptop screen] Ok so blue is ten and purple is \\
P3: [looking at the blocks and asking a rhetorical question] Thirty \\
P1 (\textbf{f}): [putting green and red blocks on the left side of the scale and purple block on the right side] Yes verify real quick but I think it is \\
P2: [observing the balanced scale] Yes thirty \\
P1: [removing green, red, and purple blocks from the scale] Yeah we got them yeah \\

\textbf{Generated Dialogue:}\\
$[$Full\_generated\_dialogue\_until\_completion$]$\\
 
Message ID: $[$index\_where\_friction\_occurs$]$\\
Contradiction: $[$describe\_the\_conflicting\_beliefs$]$\\
Contradiction Reason: $[$explain\_why\_the\_contradiction\_affects\_reasoning$]$\\
\\
\textbf{Friction Interventions}:\\
   Statement: $[$your\_friction\_intervention$]$\\
   Rationale: $[$your\_rationale$]$\\
   Score: $[$your\_score$]$
\end{tcolorbox}

\centering
\caption{“Simulated” Weights Task dataset (WTD Simulated) Friction Generation Prompt.To ground these friction interventions with personality-traits of the participants, we use \cite{mao2024editing}'s prompting framework with personality-facet combinations. We use GPT-4o as our sampling distribution $\mu$ and prompt it to simultaneously generate frictive states and friction interventions. For diversity, we use the default temperature of 1.}
\label{fig:wtd_friction_generation_prompt_simulated}
\end{figure*}

\begin{table}[h!]
    \centering
    \small
    \begin{tabular}{@{}lccc@{}}
        \toprule
        \textbf{Personality Types}          & \textbf{P1} & \textbf{P2} & \textbf{P3} \\ 
        \cmidrule(lr){2-4}
        \hspace{0.5em} Extraversion         & 4740        & 4889        & 4741        \\
        \hspace{0.5em} Neuroticism          & 5928        & 5591        & 5921        \\
        \hspace{0.5em} Agreeableness        & 4573        & 4761        & 4579        \\ \bottomrule
    \end{tabular}
       \caption{Friction Count for Participants}
    \label{tab:friction_personality_counts}
\end{table}
\clearpage

\subsection{Distributional Analysis of Original and Simulated WTD}
\label{app:wtd_distributional_analysis}

We provide a detailed distributional analysis of the Original and Simulated versions of the Weights Task Dataset (WTD). This shows the out-of-distribution (OOD) quality of our evaluation on the Original WTD in \Cref{tab:full_results_gpt_evals_table} and \Cref{tab:opt_rm_performance_comparison_percentage}. This is to show that there are significant differences in the data distribution between the two sets---and thus evaluation on the Original WTD data reasonably satisfies the OOD setting when models are aligned using the Simulated data. Note that the original dialogues are speech-to-text transcripts of actual conversations~\citep{khebour-etal-2024-common} while simulated dialogues are generated using GPT-4o ($\mu$).

\begin{figure}[t]
    \centering
        \includegraphics[width=\linewidth]{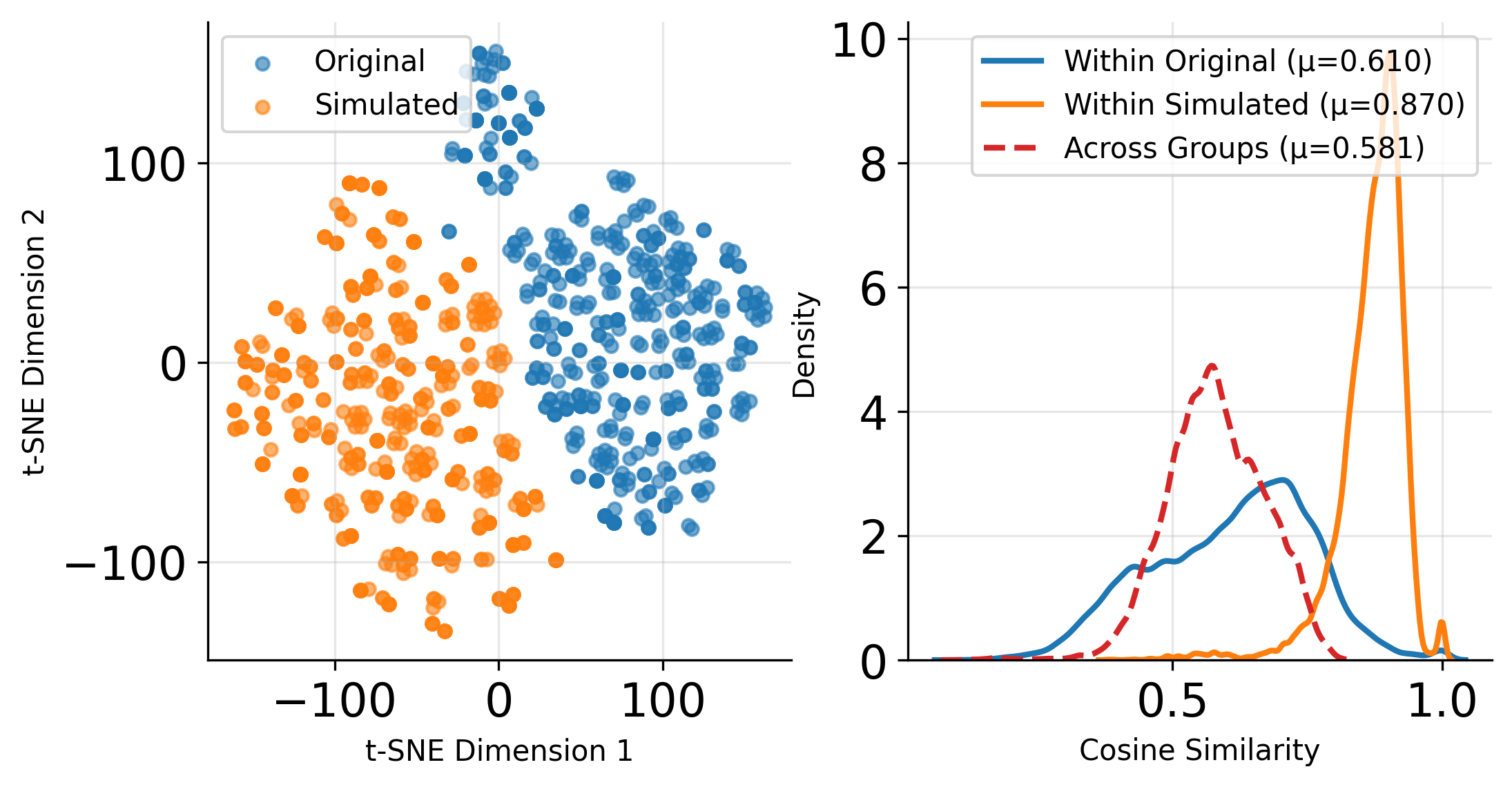}
        \includegraphics[width=\linewidth]{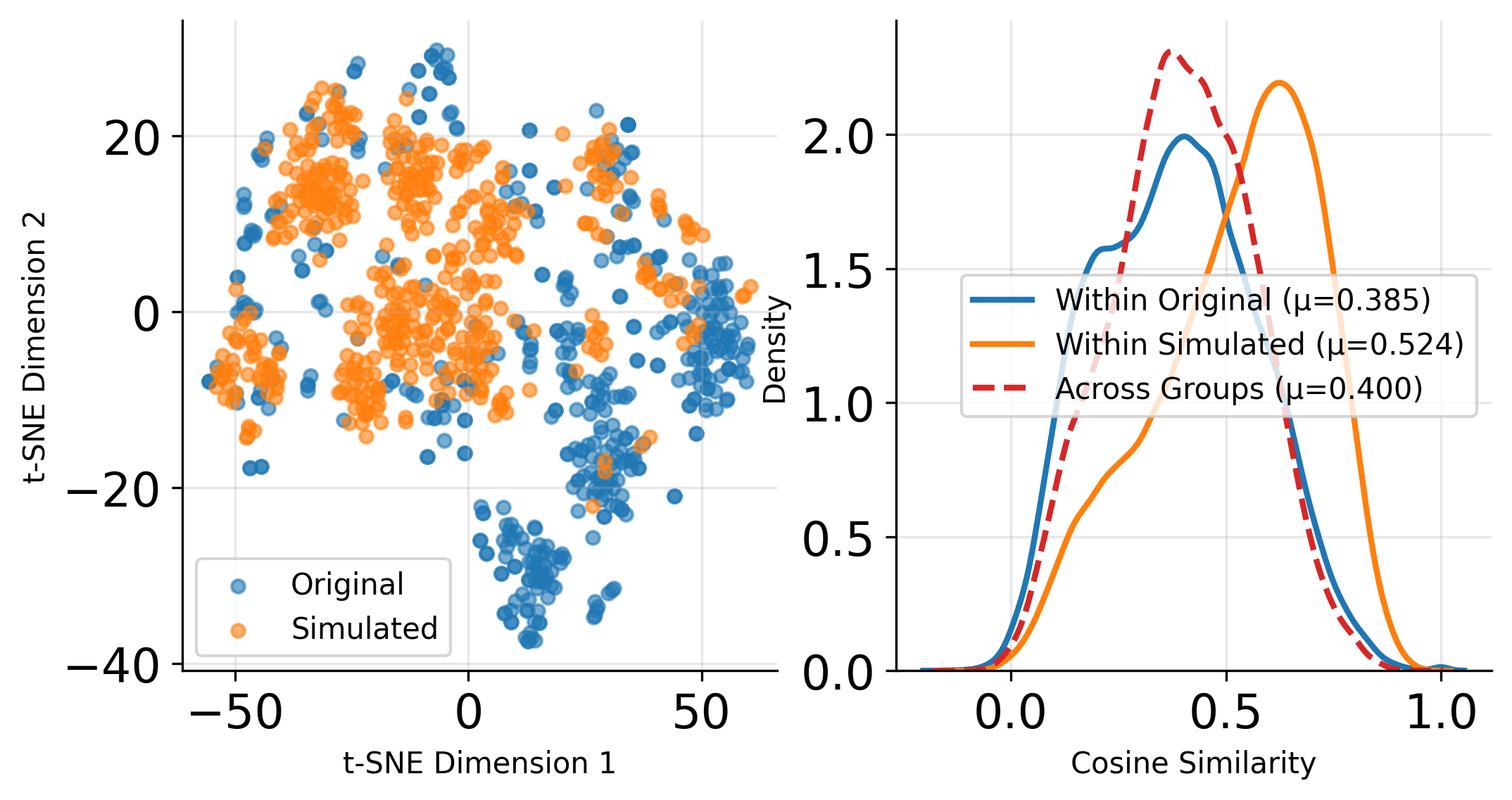}
     \caption{Plots showing distributional differences between WTD original and simulated data. Dialogue contexts (top) show clear separation with both within-group similarities exceeding across-group similarity ($\bar x=0.581$). In contrast, friction interventions (bottom) exhibit weaker separation with across-group similarity ($\bar x=0.400$) falling between within-group values.}
\label{fig:wtd_distribution_plot_both}
\end{figure}

\begin{figure*}[t]
\centering
\includegraphics[width=\textwidth]{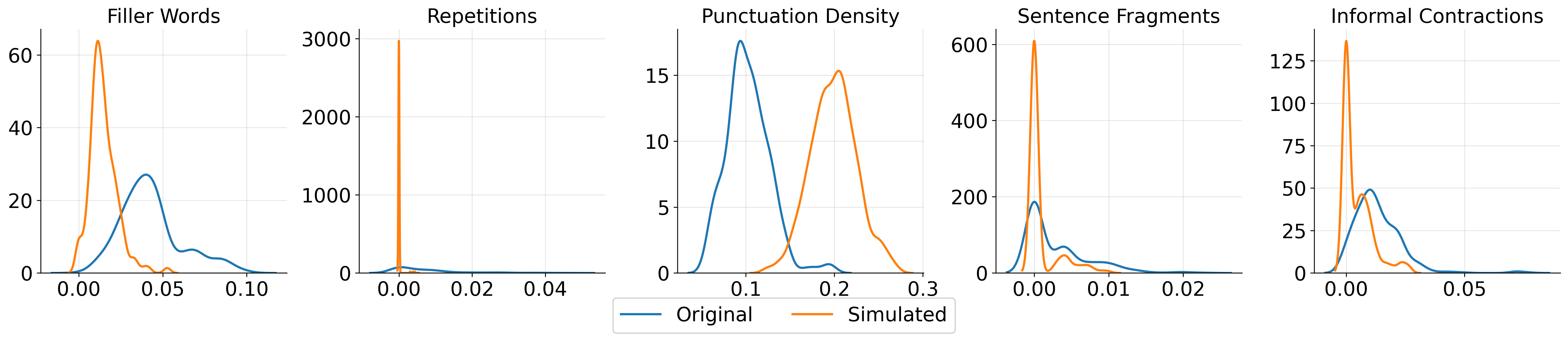}
\caption{Linguistic pattern differences between WTD original (speech-to-text transcripts) and simulated (GPT-4o-generated) collaborative dialogue data across five textual features. Plot shows results from 500 samples from the simulated and original evaluation sets.}
\label{fig:speech_patterns}
\end{figure*}

First, we sample 500 dialogue and friction intervention samples from both datasets. Next, we conduct three analyses: (1) t-SNE visualization and semantic similarity distributions using sentence embeddings\footnote{We use \url{https://huggingface.co/sentence-transformers/all-MiniLM-L6-v2} to get embeddings.} to examine clustering patterns and within/across-group similarities (see Fig.~\ref{fig:wtd_distribution_plot_both}, top plot), (2) the same embedding analysis applied to friction interventions to compare LLM-generated dialogues vs. human speech transcripts (see Fig.~\ref{fig:wtd_distribution_plot_both}, bottom plot), and (3) linguistic pattern analysis examining speech-to-text artifacts including filler words, repetitions, punctuation density, sentence fragments, and informal contractions to validate textual differences between human speech transcripts and AI-generated dialogues (shown in Fig.~\ref{fig:speech_patterns}.

The t-SNE embedding visualization shown in Fig.~\ref{fig:wtd_distribution_plot_both} (top) demonstrates clear separability between WTD Original and Simulated collaborative dialogues in semantic space, with minimal cluster overlap. Similarly, semantic similarity analysis reveals that simulated dialogues exhibit substantially higher internal coherence ($\bar x = 0.870$, $\sigma = 0.062$)\footnote{We use $\bar x$ in the text here to both denote the sample mean as these statistics are drawn from a representative subsample, and to avoid ambiguity in the use of $\mu$, which elsewhere refers to the data distribution of GPT-4o-generated data.} than original dialogues ($\bar x = 0.610$, $\sigma = 0.147$), while \textit{cross-group similarities remain consistently lower} ($\bar x = 0.581$)---signifying the differences in the two sets. All pairwise comparisons are statistically significant ($p < 0.001$). In contrast, friction intervention analysis in Fig.~\ref{fig:wtd_distribution_plot_both} (bottom) reveals significantly weaker distributional separation compared to dialogue data, with within-group similarities much closer ($\bar x = 0.385$ for original vs. $\bar x = 0.524$ for simulated) than the large gap observed in dialogue analysis ($\bar x= 0.610$ for original vs. $\bar x = 0.870$ for simulated). The cross-group similarity ($\bar x = 0.400$) falls between rather than below the within-group values, which is expected since interventions are LLM-generated for both splits.

To complement these results, we analyzed five linguistic features in the context data—filler words, repetitions, punctuation density, conjunction-led fragments, and informal contractions. These results are shown in Fig.~\ref{fig:speech_patterns}. Original data being naturally more realistic showed more disfluencies (e.g., filler words, fragments), while simulated data exhibited higher punctuation density, reflecting structured AI/LLM generation. These results indicate that the simulated dialogue data demonstrates more homogeneous content patterns unlike original dialogues that are more realistic and diverse. These results in addition to the differences in length distribution shown in \Cref{tab:wtd_data_token_stats} and \Cref{tab:wtd_original_token_stats} further validates the OOD characteristics of the original data---where \fricabbr~consistently outperforms baselines.

\subsection{Tie Counts: GPT-4o Evaluation}

\begin{figure*}[h!]
   \centering
   \includegraphics[width=0.8\textwidth]{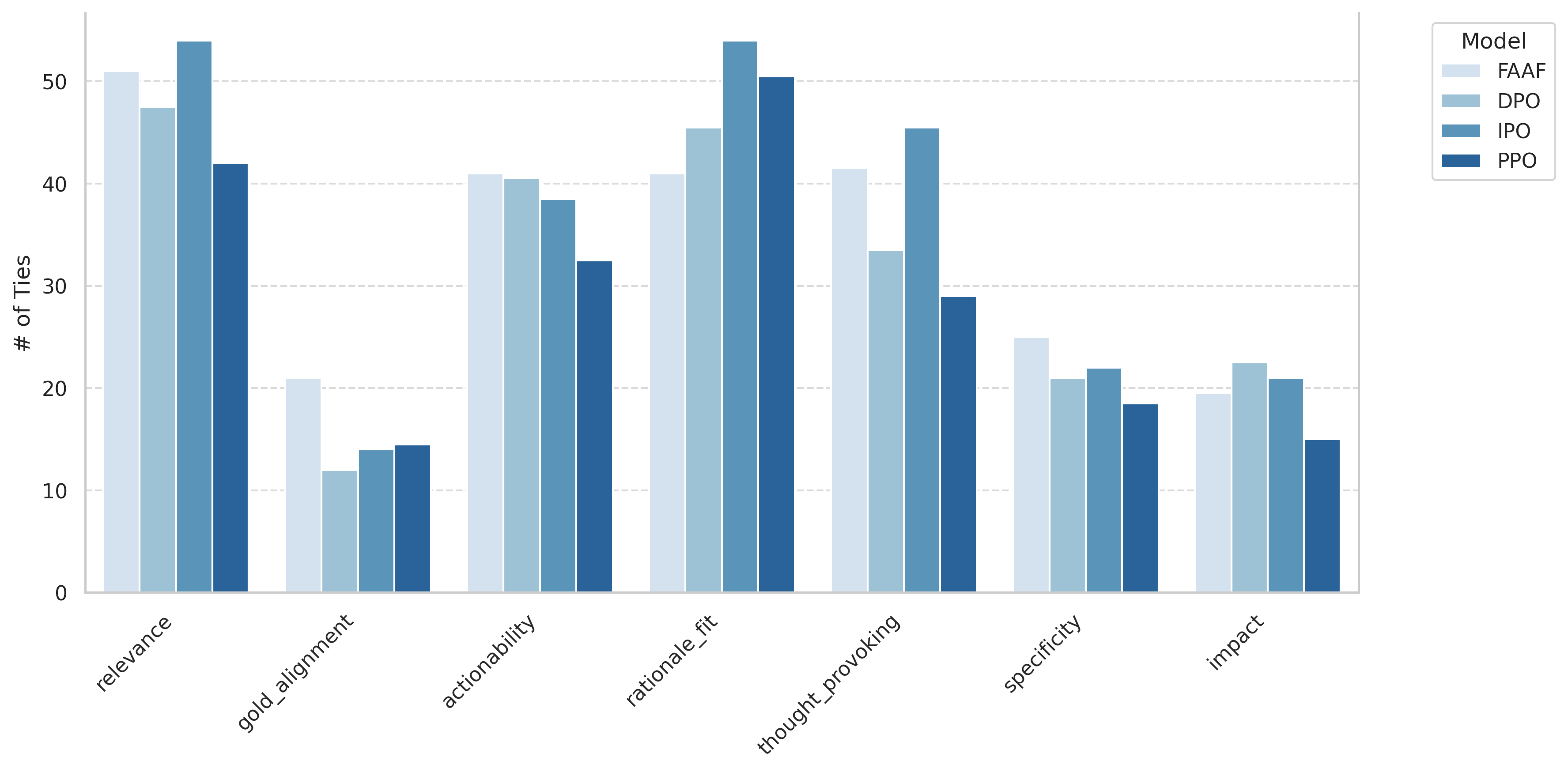}
   \vspace{2mm}
   \includegraphics[width=0.8\textwidth]{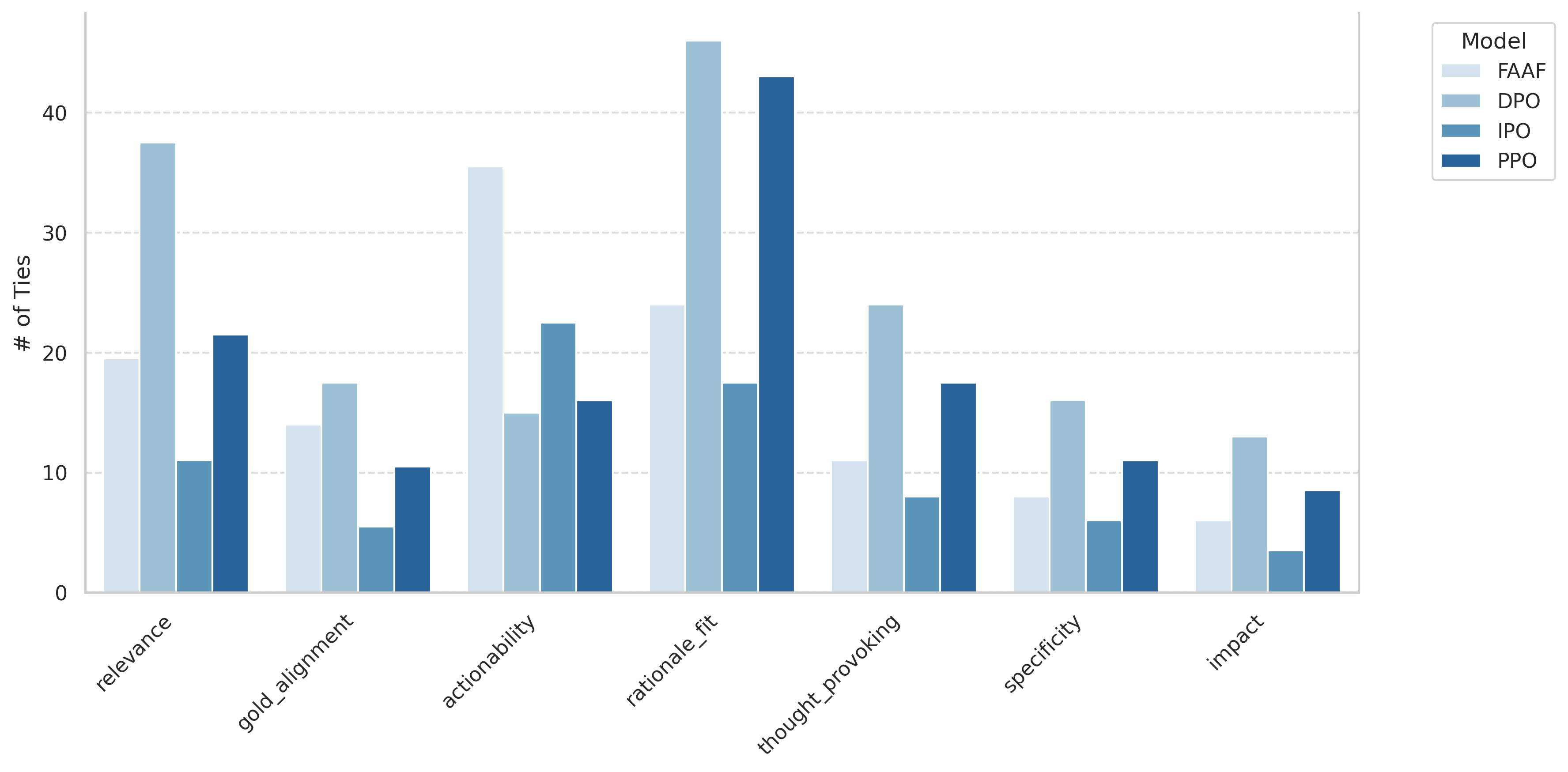}
   \vspace{2mm}
   \includegraphics[width=0.8\textwidth]{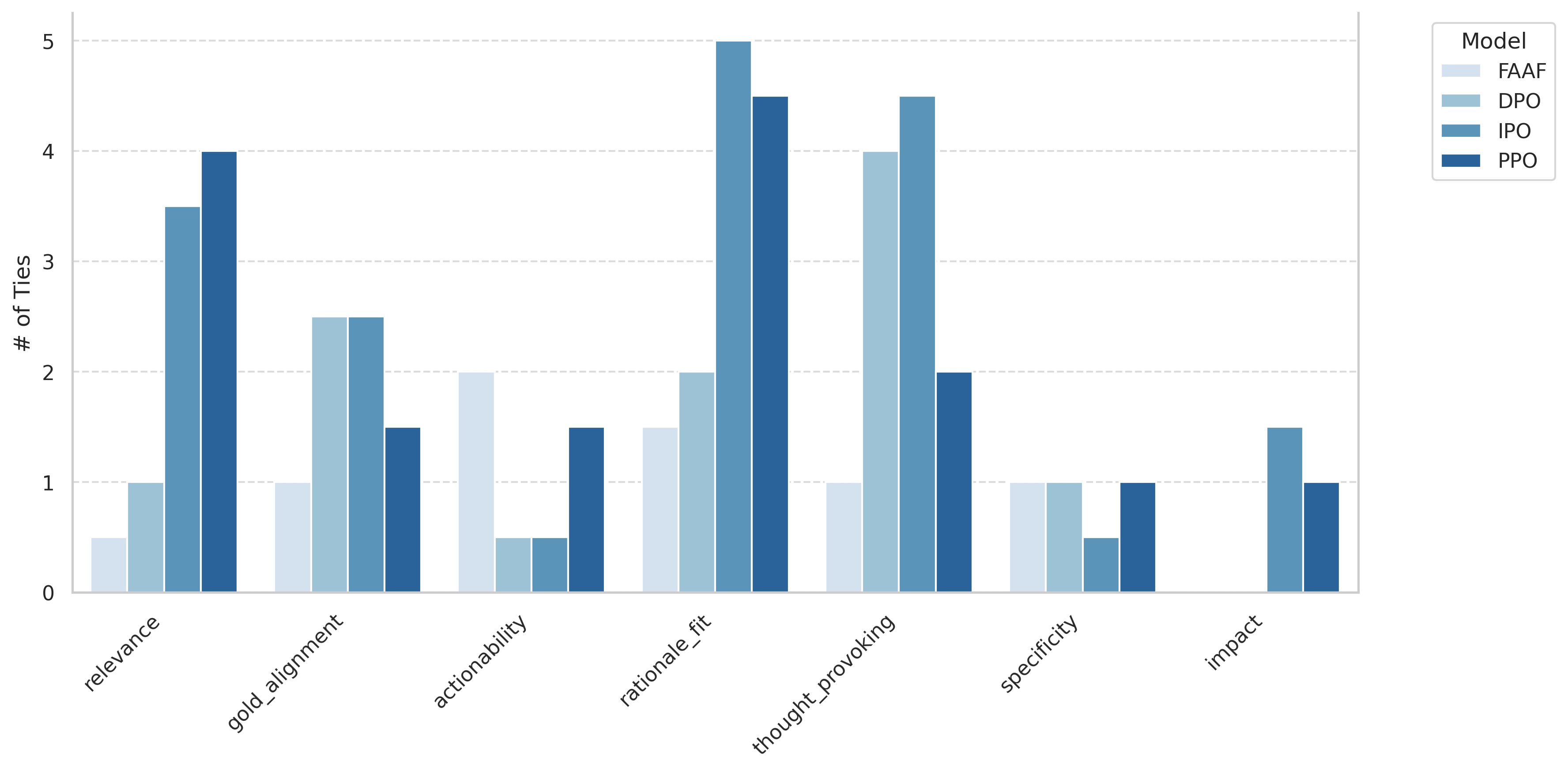}

       \caption{Comparison of average tie counts of baselines against SFT model over two runs across our 7 distinct dimensions (metrics) when evaluated using our GPT-4o-based LLM-as-a-judge evaluation in a "preference"-based setting (see Fig.~\ref{fig:friction_eval_prompt})---on DeliData (top), Simulated WTD (middle) and Original WTD (bottom). Note that there were no ties in GPT's "overall" preference between a baseline vs SFT model.}
    \label{fig:tie_counts_all_models}
\end{figure*}

Fig.~\ref{fig:tie_counts_all_models} shows the tie-count distribution (baselines vs. SFT model completions) over our 7 preference dimensions on DeliData (top), Simulated WTD (middle) and Original WTD (bottom) datasets,  when evaluated for win-rate computations using scores assigned by the LLM-judge (GPT-4o). To avoid positional bias in the placement of the sampled completions (friction interventions), we swap the positions of the two candidate samples in each run and then report the mean tie-count across each preference dimension. On average,  Fig.~\ref{fig:tie_counts_all_models}  reveals that the LLM-judge have lower raw-agreement on dimensions such as consistency of the friction intervention with its rationale (rationale\_fit), relevance and thought\_proving on all three datasets compared to aspects like gold-alignment, specificity and impact. This is expected since surface-level alignment with the golden samples are easier to assign a clear preference compared to metrics like rationale consistency especially when interventions from both the candidate and the opponent are well-justified. Consistent with our results from Table~\ref{tab:full_results_gpt_evals_table}, we find that \texttt{FAAF} model tends to tie less than other baselines on average. This trends is more pronounced in the WTD datasets consistent with \texttt{FAAF}'s overall performance as shown in our main results. 

\subsection{Human Validation of Generated Friction Interventions}
\label{appendices:human_val}

Following previous work that evaluates LLM-generated annotations and outputs \cite{wiegreffe2021measuring,wiegreffe2022reframing,nath2024okay,nath2024any}, in addition to choosing the winning intervention, we asked the human annotators\footnote{Our two human annotators have the following demographic breakdown: both male, college undergraduates, one Caucasian, one African, both fluent English speakers.} to evaluate the candidates in each sample across dimensions of {\it reasoning}, {\it specificity}, and {\it thought provoking}.  Annotators were asked to rate both candidate interventions on a 5-point Likert-type scale. For analysis, we bucketed the ratings together by valence---1 \& 2: negative valence (-1), 3: neutral valence (0), and 4 \& 5: positive valence (1), and calculated average valences and Krippendorff's $\alpha$ and Cohen's $\kappa$. We find that the average valence ratings of the various dimensions is low, very close to neutral, as are the $\alpha$ and $\kappa$ values ($\alpha = 0.276$, $\kappa = 0.205$ on DeliData samples, $\alpha = -0.265$, $\kappa = 0.004$ on WTD). There is little agreement on the qualities of the friction statement which suggests that although the annotators usually have strong agreement that there is a clear winner for each pair (see Sec.~\ref{ssec:datasets}), there is a lot of subjectivity on the qualities of these utterances. While the winning utterance was judged to be better at prompting reflection or redirecting the dialogue, it may nor be entirely clear to the annotators why. In addition, these qualities are loosely-derived from other human-LLM validation frameworks, which usually align somewhat with how LLMs themselves score things, which is often based on specific detail and level of informativity. These might not actually be the best qualities to emphasize in a collaborative dialogue, because they tend to violate Gricean principles \cite{grice1975logic} in a collaborative context, due to informativity and specific detail leading to redundancy, violating the maxim of quantity, etc.
\vfill
\newpage

\subsection{Training Settings and Hyperparameters}
\label{app:hyperparameters}

\begin{figure*}[h!]
    \centering
 
     {\includegraphics[width=\textwidth]{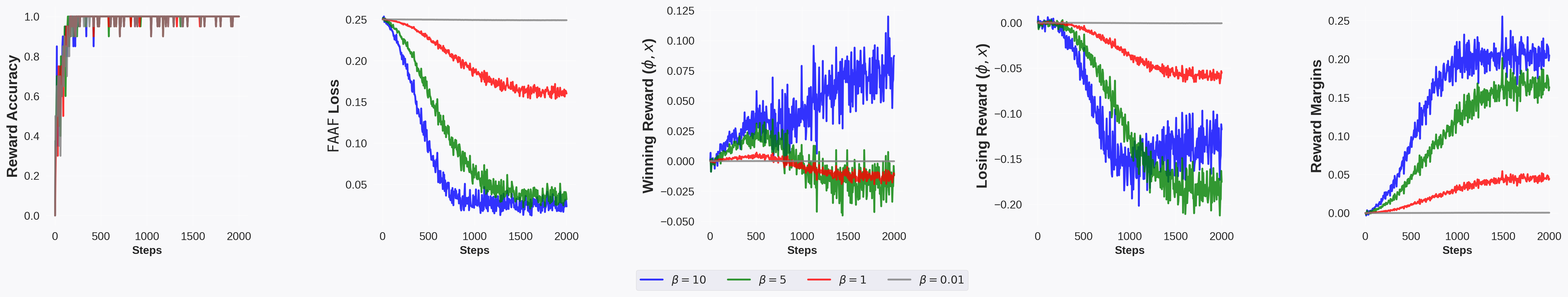}}
    \\
 {\includegraphics[width=\textwidth]{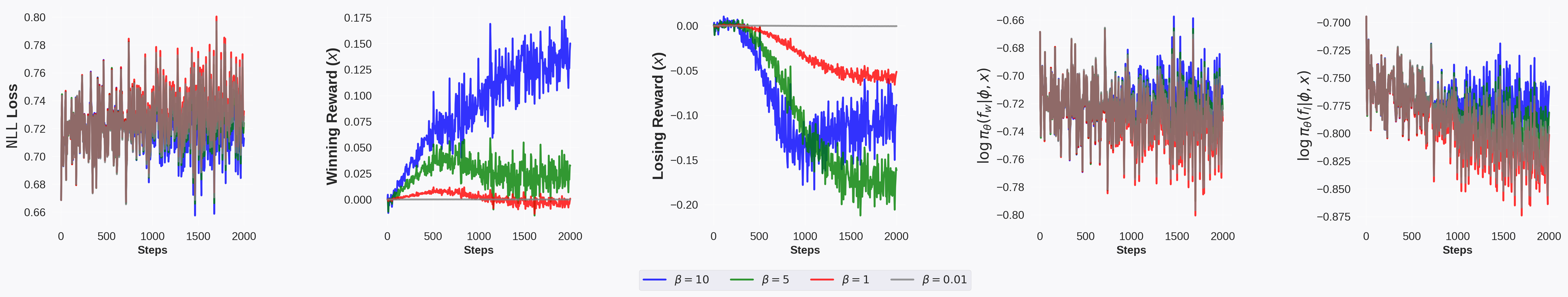}}
 {\includegraphics[width=\textwidth]{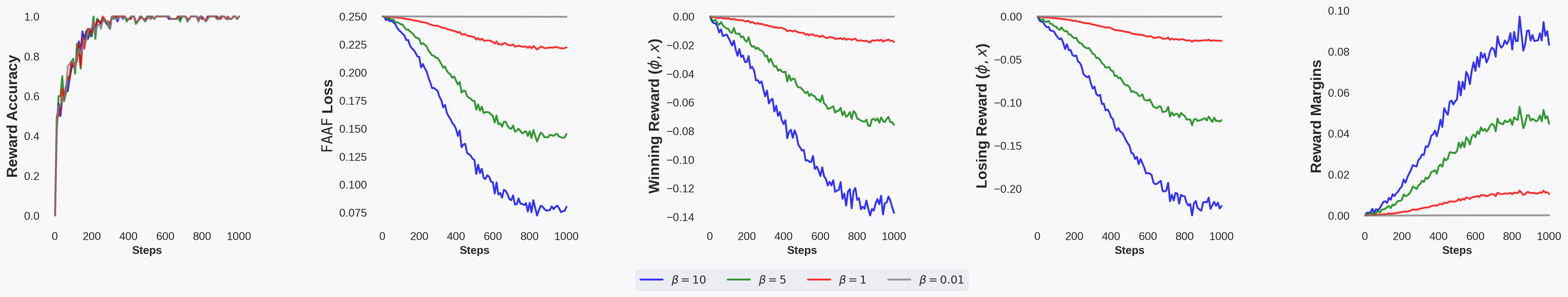}}
 {\includegraphics[width=\textwidth]{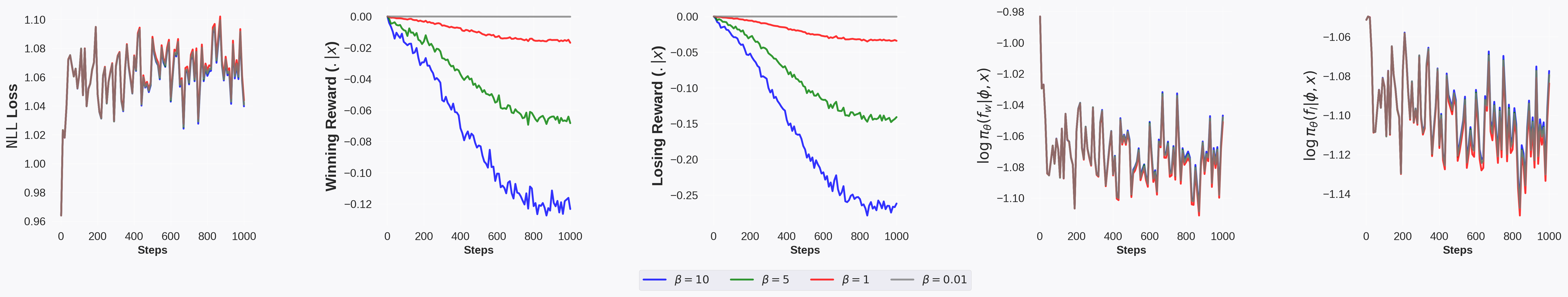}}
    \caption{Ablation study of \fricabbr's $\beta$ hyperparameter ($\beta \in \{10, 5, 1, 0.01\}$) during training on the Simulated WTD data (top half) and DeliData datasets (bottom half) across 2k and 1k training steps respectively. Higher $\beta$ values (e.g., $\beta=10$) show better implicit reward estimation as shown in \texttt{Reward Accuracy} plots and estimated preference-strengths (\texttt{Reward Margins}), while very small values ($\beta=0.01$) fail to distinguish preferences effectively. $\beta=10$ also minimizes NLL and \fricabbr~losses suggesting model stability and better convergence. As such, we report results with \fricabbr~models trained with $\beta=10$ in Table~\ref{tab:full_results_gpt_evals_table} and Table~\ref{tab:opt_rm_performance_comparison_percentage}.}
    \label{fig:wtd_training_beta_metrics}
\end{figure*}

As motivated in Sec.~\ref{sec:task-formulation}, \fricabbr-aligned $\pi_\theta$ learns to distinguish signals that determine why a particular intervention is more preferred by \textit{explicitly conditioning its implicit reward estimation on the frictive-state $\phi$}. This allows the model to estimate the true preference distribution $\mathcal{P}$ by balancing its load, from learning both \textit{with} and \textit{without} $\phi$-conditioning, given a context. This is empirically seen in Fig.~\ref{fig:wtd_training_beta_metrics} (top), where  $\pi_\theta$ displays a balanced\footnote{By balance, we mean that \textit{both} $\phi$-conditioned and $\phi$-unconditioned implicit rewards capture preference strengths from the data.} learning of "preference-strengths" between the winning and losing response (via the winning and losing response rewards as well as margins conditioned on $\phi$), subject to the KL-regularization strength parameter $\beta$.  We use the TRL Library's trainer classes for efficient multi-GPU training.

\paragraph{Hyperparameters for baselines}

All our preference alignment baselines:DPO~\citep{rafailov2024direct}, IPO~\citep{azar2024general} and PPO~\citep{schulman2017proximal} are initialized with the Supervised-finetuned (SFT) models that were trained on the winning responses ($f_w$) of DeliData and Simulated WTD training sets, following prior work to ensure the SFT model has reasonable support over the winning responses generated from $\mu$. 

For SFT models, we initialize them from the base \texttt{meta-llama/Meta-Llama-3-8B-Instruct} model in order to leverage its instruction following and general conversational abilities~\citep{llama3modelcard}. Due to compute constraints, we conducted all our training experiments with LoRA (Low-Rank Adaptation of Large Language Models), where LoRA $\alpha = 16$, LoRA dropout = $0.05$ and a LoRA R of 8 was used in training with the PEFT\footnote{\url{https://huggingface.co/docs/peft/index}} and SFT\footnote{\url{https://huggingface.co/docs/trl/en/sft_trainer}} trainers from the TRL library. We use the \texttt{bitsandbytes}\footnote{\url{https://huggingface.co/docs/transformers/main/en/quantization/bitsandbytes}} library to load our models in 4-bit quantization for more cost-efficient training. 

Additionally, as mentioned in Sec.~\ref{sec:exp}, we only compute the loss on completions (includes both frictive states $\phi$ and interventions $f_w$) using a \texttt{ConstantLengthDataset} format for more efficient training. We use a learning-rate (LR) of $1e-4$ with AdamW~\citep{loshchilov2017fixing, dettmers2024qlora} optimization with a cosine LR scheduler with a weight-decay of 0.05 and 100 warm-up steps. We train the SFT models for 6,000 steps ($\approx$ 1.5 epochs with approximately 58k samples) with an effective batch-size of 16 (gradient accumulation of 4) that reasonably achieves convergence on a 5\% validation set randomly sampled from the training sets of both datasets. For context-length, we use a maximum length of 4,096 tokens.

\paragraph{Offline baselines}

 For DPO and IPO, we use similar LoRA settings with a \texttt{max\_length} (including both prompts and responses) for 4,096 tokens with a \texttt{max\_prompt\_length} of 1,024 tokens that only minimally filters our preference pairs that exceed this length, and helps avoid out-of-memory (OOM) issues during training. We train for 2,000 steps with an effective batch size of 32 and an LR of $5e-6$, following default settings. Note that for IPO, we normalize the log-probabilities of the preferred and the dispreferred responses using their token-lengths. 

\paragraph{PPO baseline} 
 
For PPO, we additionally training an OPT 1.3B reward model (RM) following prior work~\citep{Hong2024ORPOMP} using a standard Bradley-Terry loss formulation using the TRL reward modeling library.\footnote{\url{https://github.com/huggingface/trl/blob/main/trl/trainer/reward_trainer.py}} Due to PPO's excessive compute requirements, for policy training, we use an effective batch size of 8 with a mini-batch size of 4 and gradient accumulation per 2 steps and train for 4,000 batches for two epochs. We constrain response tokens to be between 180 and 256 tokens using a \texttt{LengthSampler} while the queries are truncated to 1,024 tokens, with LR of $3e-6$ for DeliData and $1.41e-6$ for Simulated WTD. For sampling response tokens, we use a top-$p$ of 1.0 for diversity. We found that subtracting the baseline reward for the golden friction interventions ($f_w$) from the RM-assigned rewards stabilizes training. Therefore, we report results using this method in Table~\ref{tab:full_results_gpt_evals_table} and Table~\ref{tab:opt_rm_performance_comparison_percentage}.

\paragraph{\texttt{FAAF} Training Settings}

For training \texttt{FAAF}, we use a batch size of 8 with the same PEFT/LoRA settings mentioned above and train for 2,000 steps with a slightly smaller LR of $5e-7$, due to the smaller batch-sizes. For efficiency, we compute both the $\phi$-conditioned ($\pi_\theta(f|\phi,x)$) and unconditioned ($\pi_\theta(f|x)$) policy logits in parallel within each forward pass. The winning ($f_w$) and losing ($f_l$) intervention pairs for each conditioning type are batched together, requiring only two forward passes total per batch. We implement this using a modified version of the DPO Trainer\footnote{\url{https://huggingface.co/docs/trl/main/en/dpo_trainer}} from TRL, adapting it to handle the dual policy outputs. For data preprocessing, we filter pairs exceeding \texttt{max\_length} of 2,500 and 3,000 tokens in DeliData and Simulated WTD respectively, with \texttt{max\_prompt\_length} set to 1024 tokens. Following standard practice, we compute token-length normalized log-probabilities for more stable training. For the KL-regularization hyperparameter $\beta$, we conducted an ablation study over $\beta \in \{10, 5, 1, 0.01\}$. As shown in Fig.~\ref{fig:wtd_training_beta_metrics}, $\beta = 10$ achieves optimal performance across multiple metrics: (1) higher implicit reward accuracy in both $\phi$-conditioned and unconditioned policies, (2) better reward margins between winning and losing interventions, and (3) more stable convergence of the \texttt{FAAF} loss, while \texttt{NLL} loss or cross-entropy loss is relatively lower than lower $\beta$ values. Notably, while smaller $\beta$ values (e.g., $\beta = 0.01$) fail to distinguish preference margins effectively, $\beta = 10$ provides sufficient reward margins. We therefore use $\beta = 10$ for all \texttt{FAAF} experiments reported in our results. 

\paragraph{Training Hardware} We train all our models that require a reference model in memory on two Nvidia A100 GPUs, while the OPT 1.3B reward model (full-parameter training) and the SFT model were trained on a single A100 GPU. Training a single baseline for 2,000 steps roughly took 12 hours of GPU compute, but PPO models that were trained for 4,000 minibatches of size 8 took roughly 24 hours to train until convergence.


\section{\fricabbr's Robustness to Data Skew}

We do not claim that the policy trained with \fricabbr's empirical loss will have no dependence on the data distribution (whether sampling frequency or quality-wise) at all. In offline settings where the preference dataset is bound to a finite sample, there will of course be some data bias that affects the learned empirical model.

\fricabbr~is robust to possible data-skew because it combines two types of implicit rewards to create a more robust learning signal:

$\small\Delta R = \log\left(\frac{\pi_\theta(f_w|\phi,x)}{\pi_{\text{ref}}(f_w|\phi,x)}\right) - \log\left(\frac{\pi_\theta(f_l|\phi,x)}{\pi_{\text{ref}}(f_l|\phi,x)}\right)$

$\small\Delta R' = \log\left(\frac{\pi_\theta(f_w|x)}{\pi_{\text{ref}}(f_w|x)}\right) - \log\left(\frac{\pi_\theta(f_l|x)}{\pi_{\text{ref}}(f_l|x)}\right)$

The combined \fricabbr~loss function, which is more like IPO and does not have any RLHF-like BT-related sigmoid terms, is as follows:

\begin{equation}
\small
    \mathcal{L} = \mathbb{E}_{(x,\phi, f_w, f_l) \sim \mathcal{D}_\mu} \left[ \left(1 - \beta(\Delta R + \Delta R')\right)^2 \right]
\end{equation}

Empirically, \fricabbr~compares the combined effect of these implicit rewards ($\Delta R$ and $\Delta R'$) to the true preference probabilities, using terms that contain two types of conditioning: $\pi(f|\phi,x)$ and $\pi(f|x)$. This reduces the impact of specific examples appearing more frequently or with better quality in the training data.

For instance, consider training pairs where the frictive state $\phi$ is not articulated well in the training data but the intervention quality is high. These cases are definitely likely given the nature of the collaborative task where belief-states are inferred from observable utterances. In such cases, the implicit reward term $\pi_\theta(f_w|x)$ will continue to boost the likelihood of effective interventions, even though the direct contribution of $\phi$ is effectively ignored in the gradient from $\Delta R$ term due to high string similarity~\cite{pal2024smaug}. Intuitively, this provides a fallback option for the LLM to still boost learning the intervention (like token likelihood), similar to how implicit reward margins are enforced in certain works~\cite{pal2024smaug,meng2024simposimplepreferenceoptimization}.

Importantly, in our setting this is even more crucial since deploying aligned LLMs in collaborative tasks may require only exposing certain tokens (like preferred interventions without frictive states), so that the good intervention tokens continue to have high average likelihood. This is also empirically observed in the NLL (negative log likelihood loss) plot exclusively on winning/preferred intervention tokens in Fig.~\ref{fig:wtd_training_beta_metrics} (top left), where—for the right $\beta$ (10 in this case)---we can clearly see that the model tends to assign high likelihood as learning progresses. Also, note that, on average, the log-probs of winning tokens (conditioned on $\phi$) tend to decrease, as expected~\cite{rafailov2024r}, even though the reward margins tend to increase. This suggests that \fricabbr~loss is working consistently with its formulation in Eq.~\ref{eq:two_stage_main_objective_main_paper} and our core motivation from Sec.~\ref{sec:task-formulation}.

\section{Further Discussion of Evaluation Settings}
\label{app:eval-rationale}
Our evaluation follows evaluation settings from explicitly multiturn alignment benchmarks such as MTBench \cite{bai2024mt}. For example, MTBench assesses adaptability (\textit{indicating the model’s ability to respond effectively to user feedback}), or interactivity (to capture \textit{the capacity of models for proactive engagement with humans}). The aforementioned italics are verbatim from \citet{bai2024mt}, while MTBench as well as other frameworks assess helpfulness \cite{zheng2023judging,Cui2024UltraFeedbackBL}. These are likewise qualitative dimensions measured against a precollected set of responses wherein no user behavior after the LLM generation is tracked. The field has established sound reasons for conducting it this way:

\begin{enumerate}
    \item Dimensions are assessed relative to some gold-standard sample in the precollected task data. In a multiturn benchmark like MTBench, these are sample answers (from, say GPT-4) that showcase the desired dimensions. For our datasets we use, these are the gold-standard friction interventions that occur in the dialogues themselves. These samples occur mid-dialogue and thus represent both something that occurred given the preceding context and something that conditions the subsequent dialogue during data generation. Thus, multiturn evaluation gives as context the dialogue/interaction history up to the point at which a generation from the LLM is required, and then scores that generation along the different dimensions by implicitly comparing them to some gold sample. Our evaluation samples shown in Tables 9-12 follow this paradigm. \textbf{Thus, the generated interventions that receive the highest scores are those that are most aligned with the gold sample, and are therefore the most contextually appropriate in the dialogue given \textit{both} the preceding context (which both the model and the judge get to see) \textit{and} the following context (which is not visible to them).}
    
    \item Following standing practice, the interventions whose results are reported in Tables~\ref{tab:full_results_gpt_evals_table} and \ref{tab:opt_rm_performance_comparison_percentage} are sampled iteratively. For each evaluation, a dialogue history, starting at the beginning until the point where friction is required, is given to the Judge along with the candidate interventions. The evaluation is returned and the next segment of dialogue is appended to the history until the next point where friction is needed, this is given to the Judge, and the process repeats. At the end of the dialogue, the scores are aggregated and the final number is reported over all sampled interventions in all test dialogues. Since the data is precollected, the actual utterances in the dialogue history after the friction intervention do not change, but this maintains the identical conditions needed to make a consistent evaluation of two competing interventions. Since we get aggregate scores from each intervention inserted iteratively throughout a dialogue (and then average those scores over all dialogues), the model receiving the highest scores, particularly on gold-alignment and overall, in addition to the other dimensions, are implicitly the most successful at staying close to the trajectory of the actual task participant utterances.
    
    \item Now imagine an alternate condition where we want to continue the dialogue after the intervention from a source distribution $\mu$ other than the fixed dataset—this could be either actual humans performing the task or, having GPT-4 generate future utterances given a task description and context. Given a context $x$ and two candidate intervention, say $f_{\fricabbr}$ (from \fricabbr) and $f_{\text{DPO}}$ (from DPO), $\mu$ will generate different subsequent utterances when given $x$ and $f_{\fricabbr}$ and when given $x$ and $f_{\text{DPO}}$. \textbf{Thus, at the next intervention time, when evaluating \fricabbr~there would be a different trajectory to condition the generation of that next intervention than if one were evaluating DPO, and so not only do the combinatorics of the space explode, rendering evaluation intractable, but after the first intervention no fair comparison is possible since the dialogues have diverged.} This is one reason why MTBench \cite{bai2024mt} uses the "gold" context in their multiturn evaluation rather than a generated context, and also why we do. This is also one of reasons why MTBench uses a fixed set of user responses (or questions) but allows the researcher the flexibility to choose whichever gold model reference answer (GPT-4o in our case) to validate their own model responses with. We follow this same procedure.
\end{enumerate}

The above evaluation requirements also reinforce the importance of real-time human user or counterfactual evaluation studies in interactive alignment.

\section{Friction Intervention Evaluation Prompts and Sampled Representative Interventions}
\label{app:prompts-samples}
Fig.~\ref{fig:friction_eval_prompt} shows prompt used for friction intervention assessments in an LLM-as-a-judge format. We use a standard format~\cite{Cui2024UltraFeedbackBL} but adapt friction preference dimensions to collaborative task-specific settings. This prompt systematically scores friction interventions on 7 target dimensions of friction intervention quality such as correct reasoning, consistency with the agent's justification for friction, alignment with golden friction samples, clarity etc. For sampling from GPT-4o, we use standard settings with a nucleus sampling parameter (top-$p$)~\citep{holtzman2019curious} and temperature of 1.

Tables~\ref{tab:model_comparison_wtd_sim}--\ref{tab:model_comparison_deli2} show some representative interventions from each baseline and \fricabbr.

\begin{figure*}[htbp]
\begin{tcolorbox}[width=\textwidth,title={\sc Pairwise LLM-as-a-Judge Evaluation Prompt: Friction Interventions}]

System: You are an expert evaluating the quality of friction interventions in collaborative problem-solving. \\
\\
Game-definition: Participants (P1, P2, P3) are solving a block-weighing puzzle. They can only weigh two blocks at a time and know the red block is 10g. They must determine weights of all blocks (blue=10g, green=20g, purple=30g, yellow=50g) but don't know these values initially. A friction intervention is an indirect persuasion statement that prompts self-reflection and reevaluation of assumptions, like asking "Are we sure?" or suggesting to revisit steps. You must rate each intervention (between 1 to 5) along these **dimensions** given the json format below.  \\

$[$Dialogue$]$\\
$[$Gold intervention$]$ \\
$[$Intervention A$]$ \\
$[$Rationale A$]$ \\
$[$Intervention B$]$ \\
$[$Rationale B$]$ \\
\\
You must a choice between which of two interventions is more preferable and provide one sentence explanation at the end. 
 \\
 
1. Relevance: How well does the intervention address key issues or assumptions in the reasoning process? \\
2. Gold Alignment: How well does the friction intervention align with the golden friction sample? \\
3. Actionability: Does the friction intervention provide actionable guidance or suggest concrete steps for participants to improve their reasoning? \\
4. Rationale Fit: How well does the provided rationale align with the preference for the friction intervention? \\
5. Thought-Provoking: Encourages self-reflection \\
6. Specificity: Does the intervention pinpoint specific flaws, assumptions, or gaps? \\
7. Impact: To what extent does the friction intervention have the potential to change the course of the participants' reasoning? \\
\\
Format your response as follows: \\
\\ 
A: {relevance: $[1-5]$, gold\_alignment: $[1-5]$, actionability: $[1-5]$, rationale\_fit: $[1-5]$, thought\_provoking: $[1-5]$, specificity: $[1-5]$, impact: $[1-5]$} \\ B: {similar format} \\ Winner: $[$'A' or 'B'$]$ \\ Rationale: $[$One  sentence  explanation$]$

\end{tcolorbox}

\centering
\caption{Evaluation prompt used for friction intervention assessments in an LLM-as-a-judge format. }
\label{fig:friction_eval_prompt}
\end{figure*}

\begin{table*}[h!]
\small
\resizebox{\textwidth}{!}{
\begin{tabular}{p{0.15\textwidth}|p{0.85\textwidth}}
\toprule
\textbf{Prompt} & 
\begin{minipage}[t]{0.85\textwidth}
You are an expert in collaborative task analysis and reasoning. Your task is to analyze the dialogue history involving three participants (P1, P2, and P3) trying to deduce the weights of certain blocks.

For each dialogue: 

\texttt{\textless belief\_state\textgreater} Identify reasoning flaws or misunderstandings. \texttt{\textless /belief\_state\textgreater} 

\texttt{\textless rationale\textgreater} Justify the need for intervention and its impact. \texttt{\textless /rationale\textgreater} 

\texttt{\textless friction\textgreater} Generate an intervention to prompt reflection and alignment. \texttt{\textless /friction\textgreater}

\textbf{User Dialogue:} \\
P1: Alright, let’s get started! I say we measure the blue block against the red block first. Since we know the red is 10 grams, it’ll give us a good starting point! \\
P2: Great idea! I’m curious if the blue block is heavier or lighter. Let’s do it! \\
P3: I’m excited to see what we find! Measuring the blue block first sounds perfect. \\
P1: (Weighs blue block against red) Wow, the blue block is also balancing with the red! So, it must weigh 10 grams too! \\
P2: Awesome! Now we know two blocks weigh the same. What should we do next? \\
P3: Let’s measure the green block against one of the 10-gram blocks! I think it might be heavier. \\
P1: Sounds good! Let’s weigh the green block against the blue block. \\
P2: (Weighs green block against blue) Whoa! The green block is definitely heavier! \\
P3: That’s exciting! It must weigh more than 10 grams. I can feel it! \\
P1: But wait, how much heavier do you think it is? Should we weigh it against the red block next? \\
P2: Yeah, let’s do that! But should we assume it’s 20 grams just because it feels heavier? \\
P3: I think we should measure it first before jumping to conclusions! \\
P1: (Weighs green block against red) Yes! The green is heavier, but I think we should check against another 10-gram block to confirm. \\
P2: Good point! But what if we’re overthinking this? What if it’s not 20 grams? Maybe we should try weighing it against the purple block? \\
P3: That’s an interesting idea! We could find out if the green block is less than 30 grams. \\
P1: (Weighs green block against purple) Oh wow, it’s balancing! So, the green block is definitely less than 30 grams. \\
P2: So, it must be 20 grams then! What’s next? \\
P3: Let’s measure the purple block now. I’m really curious about how heavy it is! \\
P1: (Weighs purple block against green) It’s heavier! It must be the next heavier one, but I’m not sure about its exact weight. \\
P2: Maybe we should measure it against the yellow block next?

\textbf{\#\#\# Assistant:}
\end{minipage} \\
\midrule
\textbf{Gold (GPT-4o)} & \\
 
Friction & Let’s not assume anything about the yellow block’s weight until we weigh it against the purple one!\\
 
\midrule
\textbf{Model Outputs} & \\
\texttt{FAAF} & \textbf{Belief State:} P1 believes the green block weighs 20 grams, P2 is unsure about the exact weight of the green block, and P3 is cautious about making assumptions about the purple block's weight.
\\
 
& \textbf{Friction:} Wait, are we jumping to conclusions about the green block's weight? We've only confirmed it's less than 30 grams, but we haven't ruled out the possibility that it's actually 15 or 25 grams. Let's make sure we're not overestimating its weight before moving on to the next block. \\
\midrule
\texttt{DPO} & \textbf{Belief State:} P1 believes that the blue block weighs 10 grams since it balances with the red block. P2 is considering the possibility that the green block might weigh 20 grams since it feels heavier, but is uncertain and open to further measurement. P3 is considering the weight of the purple block, is curious, and wants to measure it to confirm its weight, but is not assuming a specific value yet.\\
 
& \textbf{Friction:} Hold on, maybe we're jumping to conclusions about the green block's weight. What if it's not 20 grams? Can we really confirm that by just feeling it? Let's consider other possibilities and make sure we're not relying on unverified assumptions. \\
\midrule
\texttt{PPO} & \textbf{Belief State:} P1 believes the green block is probably 20 grams based on its perceived "heft" (assumed connection between weight and feeling). P2 is cautious and thinks they shouldn't jump to conclusions. P3 is open-minded and willing to consider multiple options. \\
 
& \textbf{Friction:}  You know, we're getting ahead of ourselves by assuming the green block weighs 20 grams based on its size or feeling. Can we take a step back and really think about this? We've already found one block with the same weight as the red block. What if we weigh the green block against something else, like the yellow block? That would give us a better sense of where its weight really falls. \\
\midrule

\texttt{SFT} & \textbf{Belief State:}  P1 believes the yellow block is likely 50g. P2 is unsure about the yellow block's weight but thinks it might be 50g. P3 is unsure about the yellow block's weight and thinks it might be 50g or heavier.   \\
 
& \textbf{Friction:}    What if we measure the yellow block against the green block? That would give us some clarity about its weight.  \\
\bottomrule
\end{tabular}}
\caption{Comparison of model-generated friction interventions on the evaluation prompts of the Simulated WTD dataset.}
\label{tab:model_comparison_wtd_sim}
\end{table*}

\begin{table*}[h!]
\small
\resizebox{\textwidth}{!}{
\begin{tabular}{p{0.15\textwidth}|p{0.85\textwidth}}
\toprule
\textbf{Prompt} & 
\begin{minipage}[t]{0.85\textwidth}
You are an expert in collaborative task analysis and reasoning. Your task is to analyze the dialogue history involving three participants (P1, P2, and P3) trying to deduce the weights of certain blocks.

For each dialogue: 

\texttt{\textless belief\_state\textgreater} Identify reasoning flaws or misunderstandings. \texttt{\textless /belief\_state\textgreater} 

\texttt{\textless rationale\textgreater} Justify the need for intervention and its impact. \texttt{\textless /rationale\textgreater} 

\texttt{\textless friction\textgreater} Generate an intervention to prompt reflection and alignment. \texttt{\textless /friction\textgreater}

\textbf{User Dialogue:} \\
P1: this is nine rb so \\
P3: oh no they don't have any hints here \\
P3: Eighty gram sounds seems like a ninety gram to me but \\
P3: it's cool we have to i guess eighty alright \\
P2: Eighty let's let's go for eighty Yeah \\
P3: I would go for ninety but you guys are going for eighty Alright Cool Okay \\
P1: I would have gone for one thirty \\
P2: Eighty there you go Yeah \\
P4: For the final part of this test read the scenario and estimate the weight of the missing second mystery block \\
P2: what's this \\
P4: You will have two attempts to estimate and explain your answer you can now continue the survey \\
P1: can you read it out loud or something \\
P3: well the factory that creates the wooden blocks did not send us the second mystery block \\
P2: Okay \\
P3: you have two opportunities to estimate the weight of the second block explain in one two sentences how you determined your estimate indicate the weight of the second mystery block \\
P1: fibonacci like let's say that's the first one according to our translate thirteen one thirty \\
P2: eight mystery block's the first one so eight plus five thirteen yeah there is one thirty eight plus five \\
P3: is mystery block the first mystery block or the second mystery block \\
P2: just do it \\
P3: Wait wait wait a second \\
P3: so mystery block was the first mystery block \\
P2: Yeah \\
P3: and we already did that now this is yeah oh okay okay okay okay right \\
P2: yeah the second one means according to that statement is the second one mystery block two \\
P2: So eight plus five one thirty \\
P1: makes sense to me as much sense as i can make of it

\textbf{\#\#\# Assistant:}
\end{minipage} \\
\midrule
\textbf{Gold (GPT-4o)} & \\
 
Friction & Let’s not assume anything about the yellow block’s weight until we weigh it against the purple one!\\
 
\midrule
\textbf{Model Outputs} & \\
\texttt{FAAF} & \textbf{Belief State:} P1 believes that the second mystery block is 130 grams, based on the Fibonacci sequence and the weight of the first mystery block. P2 believes that the second mystery block is also 130 grams, following P1's reasoning. P3 is unsure and questioning the assumption that the second mystery block is the same as the first mystery block.
\\
 
& \textbf{Friction:} Hey guys, I think we're making a mistake here. We already estimated the weight of the first mystery block, and now we're assuming the second mystery block is the same? That doesn't seem right. Can we take a step back and think about this? Maybe there's another way to approach this?\\
\midrule
\texttt{DPO} & \textbf{Belief State:} P1 believes the pattern is linear and straightforward, assuming each block weighs 5 grams more than the previous one (e.g., 13 = 8 + 5, 18 = 13 + 5). P2 also assumes a linear pattern but thinks the mystery block is the second one. P3 is skeptical about the pattern, especially the value 80.
\\
 
& \textbf{Friction:} Wait a minute, are we too quick to assume the pattern is linear? What if there's another factor at play? For instance, what if the weights are related to the block's surface area or material composition? Could our assumptions be leading us down the wrong path? \\
\midrule
\texttt{PPO} & \textbf{Belief State:} P1 believes that the first mystery block has a weight of 130g, as they mentioned "fibonacci like, let's say that's the first one, according to our translate, thirteen, one thirty". P2 believes that the second mystery block has a weight of 135g, as they calculated "eight plus five, thirteen, yeah, there is one thirty eight plus five". P3 is unsure and hesitant, questioning the accuracy of the previous estimates. \\
 
& \textbf{Friction:}  Hey, let's take a step back and re-examine our assumptions. Before we jump to conclusions, can we re-read the instructions and see if there's any additional information we missed? Maybe we can approach this problem from a different angle, rather than relying on our initial guesses? \\
\midrule

\texttt{SFT} & \textbf{Belief State:}   P1 believes the second mystery block is likely 130 grams based on the Fibonacci sequence.  P2 believes the second mystery block is 80 grams based on the initial estimates.  P3 believes the second mystery block could be either 80 grams or 90 grams, expressing confusion about the weight.  \\
 
& \textbf{Friction:}    I think we might have misinterpreted the blocks again.  \\
\bottomrule
\end{tabular}}
\caption{Comparison of model-generated friction interventions on the evaluation prompts of the Original  WTD dataset.}
\label{tab:model_comparison_wtd_original}
\end{table*}


\begin{table*}[h!]
\small
\resizebox{\textwidth}{!}{
\begin{tabular}{p{0.15\textwidth}|p{0.85\textwidth}}
\toprule
\textbf{Prompt} & 
\begin{minipage}[t]{0.85\textwidth}
You are an expert in collaborative task analysis and reasoning. Participants must test the rule: 
\textit{"All cards with vowels have an even number on the other side."} 
A common mistake is verifying only one direction—ignoring the need to check whether odd-numbered cards might have vowels. This incomplete reasoning risks false validation. 

For each dialogue: 

\texttt{\textless belief\_state\textgreater} Identify reasoning flaws or misunderstandings. \texttt{\textless /belief\_state\textgreater} 

\texttt{\textless rationale\textgreater} Justify the need for intervention and its impact. \texttt{\textless /rationale\textgreater} 

\texttt{\textless friction\textgreater} Generate an intervention to prompt reflection and alignment. \texttt{\textless /friction\textgreater}

\textbf{User Dialogue:} \\
Alpaca: The task we just submitted \\
Narwhal: "the instructions aren't clear" \\
Falcon: Which card did you choose? I chose U and 4 \\
Narwhal: So which cards to turn? \\
Alpaca: So what did everyone submit? I did U and 5. \\
Duck: "I agree but we're supposed to flip the U and 4, right?" \\
Narwhal: U and B \\
Duck: Why would you flip B? \\
Narwhal: Because they are both letters. \\
Falcon: We should choose cards with vowels or an even number, I guess just U and 4. \\
Narwhal: U,4 \\
Alpaca: Well, at least we have a consensus on U. \\
Narwhal: U \\
Narwhal: U \\
Alpaca: Yes, we all said the letter U is one of the cards to flip.

\textbf{\#\#\# Assistant:}
\end{minipage} \\
\midrule
\textbf{Gold (GPT-4o)} & \\
 
Friction & How do you see consonants related to the even number requirement? Is it necessary in our current tests?\\
 
\midrule
\textbf{Model Outputs} & \\
\texttt{FAAF} & \textbf{Belief State:} Alpaca and Narwhal seem to be focusing on the "vowel" part of the rule, but not considering the "even number" part.
Duck is trying to correct Narwhal, but Narwhal is not listening to the correction.
Falcon is partially correct in suggesting that they should choose cards with vowels or even numbers, but doesn't specify which ones.
\\
 
& \textbf{Friction:} Hey, wait a minute! We're only considering one side of the rule. What about the other side? What if we flip a card with an odd number and it has a vowel? Wouldn't that mean we need to flip it too? \\
\midrule
\texttt{DPO} & \textbf{Belief State:} None\\
 
& \textbf{Friction:} Can we clarify the reasoning behind our choices? Let's try to think about both directions of the rule. What would happen if we choose a card with an odd number? Would it necessarily mean it doesn't have a vowel, or could it still have a vowel? How about the other direction? If we choose a card with a vowel, would it necessarily have an even number? \\
\midrule
\texttt{PPO} & \textbf{Belief State:} Narwhal states that `B' is not a vowel, which implies a misunderstanding of the rule regarding non-vowel cards.This affects reasoning as it leads to a potential oversight of the necessity to test odd-numbered cards, which could lead to incorrect conclusions about the rule. \\
 
& \textbf{Friction:}  What does that mean for the other cards we should check?  \\
\midrule

\texttt{SFT} & \textbf{Belief State:} Duck states `the other one is 5,' which implies they believe the 4 is already accounted for, while it is unclear whether 4 is being considered as part of the rule.The lack of clarity on whether the 4 is being considered could lead to confusion about the relevance of the odd-numbered card in the context of the rule.  \\
 
& \textbf{Friction:}   Could we clarify if the 4 is being considered as part of the rule or not?   \\
\bottomrule
\end{tabular}}
\caption{Comparison of model-generated friction interventions on the evaluation prompts of the DeliData dataset.}
\label{tab:model_comparison_deli1}
\end{table*}

\begin{table*}[h!]
\small
\resizebox{\textwidth}{!}{
\begin{tabular}{p{0.15\textwidth}|p{0.85\textwidth}}
\toprule
\textbf{Prompt} & 
\begin{minipage}[t]{0.85\textwidth}
You are an expert in collaborative task analysis and reasoning. Participants must test the rule: 
\textit{"All cards with vowels have an even number on the other side."} 
A common mistake is verifying only one direction—ignoring the need to check whether odd-numbered cards might have vowels. This incomplete reasoning risks false validation. 

For each dialogue: 

\texttt{\textless belief\_state\textgreater} Identify reasoning flaws or misunderstandings. \texttt{\textless /belief\_state\textgreater} 

\texttt{\textless rationale\textgreater} Justify the need for intervention and its impact. \texttt{\textless /rationale\textgreater} 

\texttt{\textless friction\textgreater} Generate an intervention to prompt reflection and alignment. \texttt{\textless /friction\textgreater}

\textbf{User Dialogue:} \\
Dolphin: "thinking about it, I guess it wouldn't matter what the C has on it"\\
Llama: would be E be one to turn?\\
Dolphin: "wouldn't you want to also check the 5, to test the theory? because if there is a vowel on the other side, that would disprove the theory"\\
Dolphin: E would definitely need to be turned\\
Kiwi: Oh true, the 5 might need to be turned as well\\
Dolphin: so, maybe E \& 5 is the answer?\\
Llama: I agree with the 5. unsure with the 6. but i dont think C should be turned\\
Llama: E and 5 seems like the best answer to me as well\\
Kiwi: Yes E and 5\\
Dolphin: ok, that is what i am going to chose, e \& 5\\
Kiwi: 5\\
Kiwi: E,5\\
Llama: same\\
Llama: E,5\\
Dolphin: it says we have to participate in the chat for at least 5 minutes before making our final selection. have we been chatting for 5 minutes?

\textbf{\#\#\# Assistant:}
\end{minipage} \\
\midrule
\textbf{Gold (GPT-4o)} & \\
 
Friction & What makes you think `just E' is sufficient? Can you clarify your reasoning? \\
 
\midrule
\textbf{Model Outputs} & \\
\texttt{FAAF} & \textbf{Belief State:} The participants seem to be engaging in a discussion about the rule "All cards with vowels have an even number on the other side." However, there appears to be a misunderstanding about the directionality of the rule. They are focusing on testing the rule by checking only one direction (vowels to even numbers) without considering the other direction (even numbers to vowels).
\\
 
& \textbf{Friction:} Can we flip both sides of the coin to test the rule? Let's not just focus on one direction. What if we assume the rule holds true for odd numbers as well? How would that affect our understanding of the situation? \\
\midrule
\texttt{DPO} & \textbf{Belief State:} It appears that the participants are making progress in their understanding of the rule, but there is still a potential contradiction in their reasoning.\\
 
& \textbf{Friction:} Hey team, let's reflect on our reasoning. When we say 'all cards with vowels have even numbers,' are we only considering one direction or both? Are we making assumptions about certain cards being irrelevant? Can you think of any scenarios where an odd-numbered card might have a vowel on the other side? Let's take a moment to consider all possible outcomes before making our final decision.  \\
\midrule
\texttt{PPO} & \textbf{Belief State:} Kiwi suggests turning the 5 and E, but there is a misunderstanding about the importance of the 5 in relation to the rule. The implication is that Kiwi believes the 5 is crucial to test the rule, which could lead to incorrect conclusions if the group doesn't fully understand why the 5 is necessary. 
\\
 
& \textbf{Friction:} What do you think is the significance of the 5 in relation to the rule?  \\
\midrule

\texttt{SFT} & \textbf{Belief State:} Kiwi's statement `E and 5' suggests a final conclusion without addressing potential errors, while it doesn't acknowledge the need for a more thorough discussion.This creates a misalignment in the reasoning process, as Kiwi may not fully consider the implications of their conclusion, potentially leading to incorrect submissions. 
  \\
 
& \textbf{Friction:}  Could you explain why you're confident in choosing `E and 5' without further discussion?  \\
\bottomrule
\end{tabular}}
\caption{Comparison of model-generated friction interventions on the evaluation prompts of the DeliData dataset.}
\label{tab:model_comparison_deli2}
\end{table*}

\end{document}